\documentclass[twoside,11pt]{article} 
\usepackage[margin=1.0in]{geometry}

\usepackage{amssymb,amsmath,amsthm}
\usepackage{mathtools}
\usepackage{bm,lscape}
\usepackage{bbm}
\usepackage{mathrsfs,dsfont}
\usepackage{wasysym}
\RequirePackage[
    colorlinks=true,
    linkcolor=blue,
    urlcolor=blue,
    citecolor=blue]{hyperref}
\RequirePackage{natbib}
\usepackage[utf8]{inputenc}
\usepackage[english]{babel}
\usepackage{algorithm}
\usepackage{algpseudocode}

\usepackage{graphicx}
\usepackage{color}
\usepackage{rotating}
\usepackage{authblk}
\usepackage{appendix}

\allowdisplaybreaks

\def\bSig\mathbf{\Sigma}

\newcommand{\E}{\mathbb{E}}

\newcommand{\ddr}{\mathrm{d}}

\usepackage{booktabs}

\newtheorem{theorem}{Theorem}
\newtheorem{proposition}[theorem]{Proposition}

\newtheorem{remark}[theorem]{Remark}

\providecommand{\keywords}[1]
{
  \small	
  \textbf{\textit{Keywords:}} #1
}

\begin{document}

\title{Learning-augmented count-min sketches via Bayesian nonparametrics}


\author[1]{Emanuele Dolera\thanks{emanuele.dolera@unipv.it}}
\author[2]{Stefano Favaro\thanks{stefano.favaro@unito.it}}
\author[3]{Stefano Peluchetti\thanks{speluchetti@cogent.co.jp}}
\affil[1]{\small{Department of Mathematics, University of Pavia, Italy}}
\affil[2]{\small{Department of Economics and Statistics, University of Torino and Collegio Carlo Alberto, Italy}}
\affil[3]{\small{Cogent Labs, Tokyo, Japan}}

\maketitle

\begin{abstract}
The count-min sketch (CMS) is a time and memory efficient randomized data structure that provides estimates of tokens' frequencies in a data stream of tokens, i.e. point queries, based on random hashed data. A learning-augmented version of the CMS, referred to as  CMS-DP, has been proposed by Cai, Mitzenmacher and Adams (\textit{NeurIPS} 2018), and it relies on Bayesian nonparametric (BNP) modeling of the data stream of tokens via a Dirichlet process (DP) prior, with estimates of a point query being obtained as suitable mean functionals of the posterior distribution of the point query, given the hashed data. While the CMS-DP has proved to improve on some aspects of CMS, it has the major drawback of arising from a ``constructive" proof that builds upon arguments tailored to the DP prior, namely arguments that are not usable for other nonparametric priors. In this paper, we present a ``Bayesian" proof of the CMS-DP that has the main advantage of building upon arguments that are usable, in principle, within a broad class of nonparametric priors arising from normalized completely random measures. This result leads to develop a novel learning-augmented CMS under power-law data streams, referred to as CMS-PYP, which relies on BNP modeling of the data stream of tokens via a Pitman-Yor process (PYP) prior. Under this more general framework, we apply the arguments of the ``Bayesian" proof of the CMS-DP, suitably adapted to the PYP prior, in order to compute the posterior distribution of a point query, given the hashed data. Applications to synthetic data and real textual data show that the CMS-PYP outperforms the CMS and the CMS-DP in estimating low-frequency tokens, which are known to be of critical interest in textual data, and it is competitive with respect to a variation of the CMS designed for low-frequency tokens. An extension of our BNP approach to more general queries, such as range queries, is also discussed. 
\end{abstract}



\section{Introduction}\label{sec1}

When processing large data streams, it is critical to represent data in compact structures that allow to efficiently extract information. Sketches form a broad class of compact randomized data structures that can be easily updated and queried to perform time and memory efficient estimation of some statistics of large data streams of tokens. They have found many applications in machine learning \citep{Agg(10)}, security analysis \citep{Dwo(10)}, natural language processing \citep{Goy(09)}, computational biology \citep{Zha(14),Leo(20)}, social networks \citep{Son(09)} and games \citep{Har(10)}. We refer to the monographs of \citet{Cor(12)} and \citet{Cor(20)}, and references therein, for a comprehensive and up-to-date review on sketches. A notable problem involving sketches is the estimation or recovery of the frequency of a token in the stream, typically referred to as a ``point query". The count-min sketch (CMS) of \citet{Cor(05)} is arguably the most popular approach to estimate point queries, and it relies on random hashing to obtain a sketched representation of the data. The CMS achieves the goal of using a compact data structure to save time and memory, while having provable theoretical guarantees on the estimated frequency through hashed data. Nevertheless, there are some aspects of the CMS that may be improved. First, the CMS provides point estimates, although the random hashing procedure may induce substantial uncertainty in the estimation, especially for low-frequency tokens. Second, the CMS relies on a finite universe of tokens, although it is common for large data streams to have an unbounded number of distinct tokens. Third, often there exists an a priori knowledge on the data, and therefore it may be desirable to incorporate such a knowledge into the CMS estimates.

Learning-augmented CMSs aim at improving the CMS through the use of statistical models for better exploiting the data \citep{Aam(19),Hsu(19)}. In such a context, \citet{Cai(18)} first considered a Bayesian nonparametric (BNP) approach that assumes tokens in the stream to be modeled as random samples from an unknown distribution, which is endowed with a Dirichlet process (DP) prior \citep{Fer(73)}. Then, the proposed learning-augmented CMS, referred to as CMS-DP, estimates a point query through suitable mean functionals of the posterior distribution of the point query, given the hashed data. The posterior distribution is at the core of the BNP approach, and it is derived through an intriguing ``constructive" proof that exploits a restriction property and a finite-dimensional projective property of the DP. The approach of \citet{Cai(18)} allows for an unknown (unbounded) number of distinct tokens in the universe and, most importantly: i) it allows to incorporate, through the DP prior, a priori knowledge on the data into the CMS estimates; ii) it leads, through the posterior distribution, to a natural assessment of the uncertainty of CMS estimates. \citet{Dol(21)} showed that the ``constructive" proof of \citet{Cai(18)} admits a non-trivial extension to the normalized inverse-Gaussian process (NIGP) prior, which features a distinguishing power-law tail behaviour \citep{Lij(05)}, in contrast with the geometric tail behaviour of the DP prior. This has led to the introduction of the CMS-NIGP, which is a learning-augmented CMS under power-law data streams, though with the critical limitation that it can not be tuned to the power-law degree of the data.

\subsection{Our contributions}

In this paper, we further investigate the BNP approach to develop learning-augmented CMSs. The peculiar interplay between the predictive distribution and the restriction property of the DP is the cornerstone of the ``constructive" proof of the CMS-DP. While providing an intuitive derivation of the CMS-DP, such a proof builds upon some heuristic arguments that are tailored to the DP prior. This is a critical limitation of the approach of \citet{Cai(18)}, especially with respect to the flexibility of incorporating a priori knowledge on the data into the CMS estimates. Here, we present a ``Bayesian" proof of the CMS-DP,  that is we compute rigorously the (regular) conditional distribution of a point query, given the hashed data, and we show that such a distribution coincides with the posterior distribution derived in \citet{Cai(18)}. Besides strengthening the BNP approach of \citet{Cai(18)}, our proof improves its flexibility by avoiding the use of peculiar properties of the DP, thus paving the way to go beyond the use of the DP prior. In this respect, nonparametric priors with power-law tail behaviour are of special interest, as power-law distributions occur in many situations of scientific interest, and they have significant consequences for the understanding of natural and social phenomena \citep{Cla(09)}. As well as being well known in natural language or textual data \citep{Zip(49),Can(03),Har(01)}, power-law phenomena have emerged for data arising from humans' electronic activities, e.g. patterns of website visits, emails, relations and interactions on social networks, password innovation, tags in annotation systems and editing of webpages \citep{Hub(99),Bar(05),Muc(13),Tri(14),Ryb(09),Mon(17)}. 

We extend the BNP approach of \citet{Cai(18)} to the Pitman-Yor process (PYP) prior \citep{Pit(97)}, which is arguably the most popular nonparametric prior with power-law tail behaviour. The PYP is indexed by a discount parameter that controls the tail behaviour of the prior, ranging from geometric tails to heavy power-law tails, and also captures the tail behaviour of the NIGP prior. The PYP has neither a restriction property nor a finite-dimensional projective property analogous to that of the DP, and hence: i) we apply the arguments of the ``Bayesian" proof of the CMS-DP, suitably adapted to the PYP, to compute the posterior distribution of a point query, given the hashed data; ii) we introduce a likelihood-free approach, which relies on the minimum Wasserstein distance method \citep{Ber(19)}, to estimate the prior's parameters. This leads to introduce a novel learning-augmented CMS under power-law data, referred to as CMS-PYP. Besides generalizing the CMS-DP to power-law data streams, the CMS-PYP improves remarkably the CMS-NIGP, as it can be tuned to the power-law degree of the data through the discount parameter. Applications to synthetic (Zipf) data and real textual data show that the CMS-PYP outperforms both the CMS and the CMS-DP in the estimation of low-frequency tokens, which are known to be of critical interest in textual data \citep{Pit(15)}; it turns out that the CMS-PYP also outperforms the CMS-NIGP for data with heavier power-law tails, whereas it is competitive with the CMS-NIGP for data with lighter power-law tails. In general, we show that CMS-PYP is competitive with respect to the count-mean-min (CMM) of \citet{Goy(12)}, which provides a variation of the CMS designed for the estimation of low-frequency tokens, and also with respect to the bootstrap-debiased-count-min (BDCM) of \citet{Tin(18)}.

\subsection{Organization of the paper}

The paper is structured as follows. In Section \ref{sec2} we briefly review the CMS-DP and its ``constructive" proof, and then we present our ``Bayesian" proof of the CMS-DP. In Section \ref{sec3} we develop the CMS-PYP through the computation of posterior distribution of a point query, given the hashed data, and the estimation of the prior's parameters. Section \ref{sec5} contains a numerical illustration of the CMS-PYP, both on synthetic data and real data. In Section \ref{sec6} we discuss our work, as well as its extension to the problem of estimating more general queries, and present some directions for future research. Proofs of our results, except for the ``Bayesian" proof of the CMS-DP, and additional technical results are deferred to appendices.


\section{A ``Bayesian" derivation of the CMS-DP}\label{sec2}

For any $m\geq1$ let $x_{1:m}=(x_{1},\ldots,x_{m})$ be a stream of $\mathcal{V}$-valued tokens, with $\mathcal{V}$ being a space of types (symbols). Assuming $x_{1:m}$ to be available through summaries obtained by its random hashing, the goal is to estimate, or recover, the frequency of a new token $x_{m+1}$ in $x_{1:m}$, i.e.
\begin{displaymath}
f_{x_{m+1}}=\sum_{i=1}^{m}\mathbbm{1}_{\{x_{i}\}}(x_{m+1}).
\end{displaymath}
The CMS \citep{Cor(05)} is the most popular approach to estimate the point query $f_{x_{m+1}}$. For positive integers $J$ and $N$ such that $[J]=\{1,\ldots,J\}$ and $[N]=\{1,\ldots,N\}$, let $h_{1},\ldots,h_{N}$, with $h_{n}:\mathcal{V}\rightarrow [J]$, be random hash functions that are i.i.d. according to a pairwise independent hash family $\mathcal{H}$. That is, $h\in\mathcal{H}$ is such that for all $v_{1},v_{2}\in\mathcal{V}$, with $v_{1}\neq v_{2}$, the probability that $v_{1}$ and $v_{2}$ hash to any $j_{1}$ and $j_{2}$, respectively, is $\text{Pr}[h(v_{1})=j_{1},\,h(v_{2})=j_{2}]=J^{-2}$. Pairwise independence is typically known as strong universality, and it implies uniformity, i.e. $\Pr[h(v)=j]=J^{-1}$ for any $j\in[J]$ \citep[Chapter 3]{Cor(20)}. Hashing $x_{1:m}$ through $h_{1},\ldots,h_{N}$ creates $N$ vectors of $J$ buckets, say $\{(C_{n,1},\ldots,C_{n,J})\}_{n\in[N]}$, as follows: $C_{n,j}$ is initialized at zero, and whenever a new token $x_i$ with $h_{n}(x_{i})=j$ is observed we set  $C_{n,j} \leftarrow 1+ C_{n,j}$ for every $n\in [N]$. The CMS estimates $f_{x_{m+1}}$ by
\begin{equation}\label{cms_main}
\hat{f}^{\text{\tiny{(CMS)}}}=\min\{C_{1,h_{1}(x_{m+1})},\ldots,C_{N,h_{N}(x_{m+1})}\}.
\end{equation}
We refer to \citet[Chapter 3]{Cor(20)} for a detailed account on the CMS and generalizations thereof dealing with general small summaries for big data. In this section, we consider the CMS-DP \citep{Cai(18)}, which is a learning-augmented version of the CMS that relies on BNP modeling of the stream $x_{1:m}$ through a DP prior. We briefly review the CMS-DP and its ``constructive" proof, and then we present our ``Bayesian" proof of the CMS-DP.

\subsection{The CMS-DP and its ``constructive" proof}

\subsubsection{The DP prior}

A simple and intuitive definition of the DP follows from its stick-breaking construction \citep{Fer(73),Set(94)}. For $\theta>0$ let: i) $(B_{i})_{i\geq1}$ be random variables i.i.d. as a Beta distribution with parameter $(1,\theta)$; ii) $(V_{i})_{i\geq1}$ be random variables independent of $(B_{i})_{i\geq1}$, and i.i.d. from a non-atomic distribution $\nu$ on $\mathcal{V}$. Then, define $P_{1}=B_{1}$ and $P_{j}=B_{j}\prod_{1\leq i\leq j-1}(1-B_{i})$ for $j\geq2$, in such a way that $\sum_{i\geq1}P_{i}=1$ almost surely. The (discrete) random probability measure $P=\sum_{j\geq1}P_{j}\delta_{V_{j}}$ is a DP on $\mathcal{V}$ with (base) distribution $\nu$ and mass parameter $\theta$. The law of $P$ thus provides a prior distribution on the space of discrete distributions on $\mathcal{V}$. For short, $P\sim\text{DP}(\theta;\nu)$. See \citet{Gho(17)} and references therein for a comprehensive account of the DP, including its definition in terms of the normalization of a Gamma completely random measure. For our work, it is useful to recall the restriction property and the finite-dimensional projective property of the DP \citep{Fer(73),Reg(01)}. The restriction property is stated as follows: if $A\subset\mathcal{V}$ and $P_{A}$ is the random probability measure on $A$ induced by $P\sim\text{DP}(\theta;\nu)$ on $\mathcal{V}$, i.e. the renormalized restriction of $P$ to $A$, then $P_{A}\sim\text{DP}(\theta\nu(A);\nu_{A}/\nu(A))$, where $\nu_{A}$ is the restriction of the measure $\nu$ to $A$. The finite-dimensional projective property is stated as follows: if $\{B_1,\ldots,B_{k}\}$ is a measurable $k$-partition of $\mathcal{V}$, for $k\geq1$, then $P\sim \text{DP}(\theta;\nu)$ is such that $(P(B_1), \ldots, P(B_k))$ is distributed as a Dirichlet distribution with parameter $(\theta\nu(B_{1}),\ldots,\theta\nu(B_{k}))$.

Because of the discreteness of $P\sim\text{DP}(\theta;\nu)$, a random sample $X_{1:m}=(X_{1},\ldots,X_{m})$ from $P$ induces a random partition of the set $\{1,\ldots,m\}$ into $1\leq K_{m}\leq m$ partition subsets, labelled by distinct types $\mathbf{v}=\{v_1,\ldots,v_{K_{m}}\}$, with corresponding frequencies $(N_{1,m},\ldots,N_{K_{m},m})$ such that $1\leq N_{i,m}\leq n$ and $\sum_{1\leq i\leq K_{m}}N_{i,m}=m$. For $1\leq l\leq m$ let $M_{l,m}$ be the number of distinct types with frequency $l$, i.e. $M_{l,m}=\sum_{1\leq i\leq K_{m}} \mathbbm{1}_{\{N_{i,m}\}}(l)$ such that $\sum_{1\leq l\leq m}M_{l,m}=K_{m}$ and $\sum_{1\leq l\leq m}lM_{l,m}=m$. The distribution of $\mathbf{M}_{m}=(M_{1,m},\ldots,M_{m,m})$ is defined on $\mathcal{M}_{m,k}=\{(m_{1},\ldots,m_{n})\text{ : }m_{l}\geq0,\,\sum_{1\leq l\leq m}m_{l}=k,\,\sum_{1\leq l\leq m}lm_{l}=m\}$, such that
\begin{equation}\label{eq_ewe}
\text{Pr}[\mathbf{M}_{m}=\mathbf{m}]=m!\frac{\theta^{k}}{(\theta)_{(m)}}\prod_{i=1}^{m}\frac{1}{i^{m_{i}}m_{i}!}\mathbbm{1}_{\mathcal{M}_{m,k}}(\mathbf{m}),
\end{equation}
where $(a)_{(n)}$ denotes the rising factorial of $a$ of order $n$, i.e. $(a)_{(n)}=\prod_{0\leq i\leq n-1}(a+i)$. See \cite[Chapter 3]{Pit(06)}, and references therein, for details on the sampling formula \eqref{eq_ewe}. Let $\mathbf{v}_{l}=\{v_{i}\in \mathbf{v}\text{ : } N_{i,m}=l\}$, i.e. the labels of types with frequency $l$, and let $\mathbf{v}_{0}=\mathcal{V}-\mathbf{v}$, i.e. the labels in of types not belonging to $\mathbf{v}$. The predictive distribution induced by $P\sim\text{DP}(\theta;\nu)$ is
\begin{align} \label{eq:pred_seen_dp}
\text{Pr}[X_{m+1} \in \mathbf{v}_{l}\,|\, X_{1:m}] =\text{Pr}[X_{m+1} \in \mathbf{v}_{l}\,|\, \mathbf{M}_{m}=\mathbf{m}]=
\begin{cases} 
 \frac{\theta}{\theta+m}&\mbox{ if } l=0\\[0.4cm] 
\frac{lm_{l}}{\theta+m}&\mbox{ if } l\geq1,
\end{cases}
\end{align}
for $m\geq1$. The DP prior is characterized as the sole (discrete) nonparametric prior for which: i) the conditional probability that $X_{m+1}$ belongs to $\mathbf{v}_{0}$, given $X_{1:m}$, depends on $X_{1:m}$ only through $m$; ii) the conditional probability that $X_{m+1}$ belongs to $\mathbf{v}_{l}$, given $X_{1:m}$, depends on $X_{1:m}$ only through $m$ and $M_{l,m}$. Such a characterization is typically referred to as the ``sufficientness" postulate of the DP \citep{Reg(78),Zab(97),Bac(17)}.

\subsubsection{The CMS-DP}

The CMS-DP of \citet{Cai(18)} assumes that the stream $x_{1:m}$ is modeled as a random sample $X_{1:m}$ from an unknown discrete distribution $P$, which is endowed with a DP prior. That is,
    \begin{align}\label{eq:bnp_dp}
    X_{1:m}\,|\, P  &\,\stackrel{\mbox{\scriptsize{iid}}}{\sim}\,P\\[0.2cm]
     \notag P &\, \sim\,\text{DP}(\theta;\nu)
    \end{align}
for $m\geq1$. Let $h_{1},\ldots,h_{N}$ be a collection of random hash functions that are i.i.d. from the strong universal family $\mathcal{H}$, and assume that $h_{1},\ldots,h_{N}$ are independent of $X_{1:m}$ for any $m\geq1$; in particular, by de Finetti's representation theorem, it holds that $h_{1},\ldots,h_{N}$ are independent of $P \sim\,\text{DP}(\theta;\nu)$. Under the CMS-DP the $X_{i}$'s are hashed through $h_{1},\ldots,h_{N}$, thus creating $\{(C_{n,1},\ldots,C_{n,J})\}_{n\in[N]}$, and estimates of the point query $f_{X_{m+1}}$, with $X_{m+1}$ being of an arbitrary type $v\in\mathcal{V}$, are obtained as functionals of the posterior distribution of $f_{X_{m+1}}$ given the hashed frequencies $\{C_{n, h_{n}(X_{m+1})}\}_{n\in[N]}$. \citet{Cai(18)} provided an intriguing ``constructive" derivation of such a posterior distribution, which relies on two main arguments: 
\begin{itemize}
\item[A1)] the restriction property of the DP, in combination with the ``sufficientness" postulate of the DP, implies that, because of the strong universality of $\mathcal{H}$ and the independence between $h_{n}$ and $X_{1:m}$, the tokens $X_{i}$'s hashed in the $j$-th bucket $C_{n,j}$ constitute random samples from a DP with scaled mass parameter $\theta/J$, for any $j\in[J]$ and $n\in[N]$;
\item[A2)] the finite-dimensional projective property of the DP implies that, because of the strong universality of $\mathcal{H}$, the vector of hashed frequencies $\mathbf{C}_{n}=(C_{n,1},\ldots,C_{n,J})$ is distributed according to a Dirichlet-Multinomial distribution with parameter $(\theta/J,\ldots,\theta/J)$, for any $n\in[N]$.
\end{itemize}

For a single hash function $h_{n}$, the main result of \citet{Cai(18)} may be summarized as follows. A random sample $X_{1:m}$ from $P\sim\,\text{DP}(\theta;\nu)$ induces a random partition of $\{1,\ldots,m\}$ into subsets labelled by $\mathbf{v}\in\mathcal{V}$, and \eqref{eq:pred_seen_dp} is the posterior distribution, given $X_{1:m}$, over which subset $X_{m+1}$ joins. The frequency of that subset is the point query $f_{X_{m+1}}$ we seek to estimate, i.e.
\begin{equation}\label{step1_cai}
\text{Pr}[f_{X_{m+1}}=l\,|\,X_{1:m}]=\text{Pr}[X_{m+1} \in \mathbf{v}_{l}\,|\, X_{1:m}]
\end{equation}
for $l=0,1,\ldots,m$. However, we are assuming that the sampling information $X_{1:m}$ is available only through $\{C_{n, h_{n}(X_{m+1})}\}_{n\in[N]}$, and hence the posterior distribution \eqref{step1_cai} is not of interest itself. Instead, it is of interest the distribution of $f_{X_{m+1}}$, which is obtained from \eqref{step1_cai} by marginalizing out $X_{1:m}$. By combining \eqref{step1_cai} with \eqref{eq_ewe} \citep[Section 3]{Cai(18)}, it holds that
\begin{align}\label{marg_dp_11}
p_{f_{X_{m+1}}}(l;m,\theta):=\text{Pr}[f_{X_{m+1}}=l]=\theta\frac{(m-l+1)_{(l)}}{(\theta+m-l)_{(l+1)}}.
\end{align}
For any $n\in[N]$, strong universality of $\mathcal{H}$ and independence between $h_{n}$ and $X_{1:m}$ imply that $h_{n}$ induces a (fixed) $J$-partition of $\mathcal{V}$, say $\{B_{h_{n},1},\ldots,B_{h_{n},J}\}$, and the measure with respect to $P\sim \text{DP}(\theta; \nu)$ of each $B_{h_{n},j}$ is $1/J$. Therefore, according to A1), $h_{n}$ turns $P\sim\text{DP}(\theta;\nu)$ into $J$ bucket-specific DPs, say $P_{j}\sim\text{DP}(\theta/J;J\nu_{B_{h_{n},j}})$ for $j=1,\ldots,J$, such that $P_{j}$ governs the distribution of the sole $X_{i}$'s hashed in $B_{h_{n},j}$. For any $l=0,1,\ldots,c_{n}$, \citet{Cai(18)} thus set
\begin{equation}\label{step2_cai}
\text{Pr}[f_{X_{m+1}} = l\,|\, C_{n, h_{n}(X_{m+1})}=c_{n}]=p_{f_{X_{m+1}}}\left(l;c_{n},\frac{\theta}{J}\right).
\end{equation}
This is an heuristic assignment, namely the left-hand side of \eqref{step2_cai} is not obtained through a rigorous computation of the (regular) conditional distribution of $f_{X_{m+1}}$ given $C_{n, h_{n}(X_{m+1})}$. We refer to such a derivation as the ``constructive" proof of the posterior distribution of $f_{X_{m+1}}$ given $C_{n, h_{n}(X_{m+1})}$.

For the collection of hash functions $h_{1},\ldots,h_{N}$, the posterior distribution of $f_{X_{m+1}}$, given $\{C_{n, h_{n}(X_{m+1})}\}_{n\in[N]}$, follows from Equation \eqref{step2_cai} by means of the assumption that the $h_{n}$'s are i.i.d. according to the strong universal family $\mathcal{H}$. In particular, by a direct application of Bayes theorem, \citet[Section 3]{Cai(18)}, showed that for $l=0,1,\ldots,\min\{c_{1},\ldots,c_{N}\}$ it holds that
\begin{align}\label{post_full_dp}
    &\text{Pr}[f_{X_{m+1}} = l\,|\, \{C_{n, h_{n}(X_{m+1})}\}_{n\in[N]}=\{c_{n}\}_{n\in[N]}]=\frac{\prod_{n\in[N]}p_{f_{X_{m+1}}}\left(l;c_{n},\frac{\theta}{J}\right)}{(p_{f_{X_{m+1}}}(l;m,\theta))^{N-1}}.
\end{align}
CMS-DP estimates of the point query $f_{X_{m+1}}$, with respect to a suitable choice of a loss function, are obtained as functionals of the posterior distribution \eqref{post_full_dp}, e.g. posterior mode, mean and median. We refer to  \citet{Cai(18)} for a detailed discussion on BNP estimators of $f_{X_{m+1}}$ and their interplay with the CMS. For a concrete application of \eqref{post_full_dp}, it remains to estimate the unknown prior's parameter $\theta>0$, and this follows from A2). In particular, $\mathbf{C}_{n}$ is distributed as a Dirichlet-Multinomial distribution with parameter $(\theta/J,\ldots,\theta/J)$, and the distribution of $\{\mathbf{C}_{n}\}_{n\in[N]}$ follows by the assumption that the $h_{n}$'s are i.i.d. from $\mathcal{H}$, that is
\begin{align}\label{marg}
&\text{Pr}[\{\mathbf{C}_{n}\}_{n\in[N]}=\{\mathbf{c}_{n}\}_{n\in[N]}]=\prod_{n\in[N]}\frac{m!}{(\theta)_{(m)}}\prod_{j=1}^{J}\frac{(\frac{\theta}{J})_{(c_{n,j})}}{c_{n,j}!}.
\end{align}
Equation \eqref{marg} provides the (marginal) likelihood function of $\{\mathbf{c}_{n}\}_{n\in[N]}$. The explicit form of such a function allows for an easy implementation of a Bayesian estimation of the prior's parameter $\theta$. \citet{Cai(18)} adopted an empirical Bayes approach, which consists in estimating $\theta$ by maximizing, with respect to $\theta$, the likelihood function \eqref{marg}. A fully Bayes, or hierarchical Bayes, approach can be also applied by setting a suitable prior distribution on $\theta$.

\subsection{A ``Bayesian" proof of the CMS-DP}

In \citet{Cai(18)}, the interplay between the predictive distribution and the restriction property of the DP is the cornerstone for the derivation of \eqref{step2_cai}, i.e. the posterior distribution of $f_{X_{m+1}}$ given $C_{n, h_{n}(X_{m+1})}$. The ``constructive" proof of the CMS-DP imposes two strong constraints with respect to the choice of the prior distribution: C1) the predictive distribution induced by the prior must have a simple analytical expression, i.e. the marginalization with respect to the sampling information $X_{1:m}$ must be doable explicitly, and it must satisfy a ``sufficientness" postulate analogous to that of the DP prior; C2) the prior distribution must have a restriction property analogous to that of the DP prior, which allows us to make use of the distribution of $f_{X_{m+1}}$ to assign the posterior distribution of $f_{X_{m+1}}$ given $C_{n, h_{n}(X_{m+1})}$. Nonparametric priors obtained by normalizing (homogeneous) completely random measures \citep{Jam(02), Pru(02), Pit(03), Reg(03),Jam(09)} form a broad class of priors that generalize the DP prior and satisfy C2); this follows from the Poisson process representation of completely random measures, for which the Poisson coloring theorem holds true \citep[Chapter 5]{Kin(93)}. However, the DP is the sole normalized (homogeneous) completely random measure that satisfies C1) \citep{Reg(78)}. Beyond normalized completely random measures, the PYP prior is a popular generalization of the DP prior that satisfies C1). However, the PYP does not satisfy C2); this is because the PYP is not a normalized completely random measure. To the best of our knowledge, the DP prior is the sole (discrete) nonparametric prior that satisfies both C1) and C2), and hence it is the sole prior for which the ``constructive" proof of \citet{Cai(18)} works. The ``constructive" proof thus determines a limitation for the BNP approach of  \citet{Cai(18)}, implying a lack of flexibility in the choice of the prior distribution for BNP modeling of $x_{1:m}$.

Here, we present a rigorous derivation of the posterior distribution of $f_{X_{m+1}}$ given $C_{n, h_{n}(X_{m+1})}$, which is referred to as the ``Bayesian" proof of the CMS-DP.  For any $n\in[N]$, we consider the problem of computing the (regular) conditional distribution of $f_{X_{m+1}}$ given $C_{n, h_{n}(X_{m+1})}$, i.e.
\begin{align}\label{eq:maintoprove}
&\text{Pr}[f_{X_{m+1}}=l\,|\,C_{h_{n}(X_{m+1})}=c_{n}]=\frac{\text{Pr}\left[f_{X_{m+1}}=l,\sum_{i=1}^{m}\mathbbm{1}_{\{h_{n}(X_{i})\}}(h_{n}(X_{m+1}))=c_{n}\right]}{\text{Pr}\left[\sum_{i=1}^{m}\mathbbm{1}_{\{h_{n}(X_{i})\}}(h_{n}(X_{m+1}))=c_{n}\right]},
\end{align}
for $l=0,1,\ldots,c_{n}$. In the next theorem, we show that the (regular) conditional distribution \eqref{eq:maintoprove} coincides with the posterior distribution \eqref{step2_cai} obtained by means of the ``constructive" proof. That is, the ``Bayesian" proof and the  ``constructive" proof lead to the same posterior distribution. As a critical feature, our ``Bayesian" proof stands out for not relying on the peculiar restriction property of the DP; instead, by exploiting the strong universality of $\mathcal{H}$, the ``Bayesian" proof relies on evaluating the numerator and the denominator of  \eqref{eq:maintoprove} through standard combinatorial arguments and well-known  distributional properties of a random sample $X_{1:m}$ from the DP \citep{Pit(03),Pit(06),San(06)}, i.e. marginal properties. It emerges that the ``Bayesian" proof has two main advantages with respect to the ``constructive" proof: i) it provides a rigorous proof of the CMS-DP, which avoids any heuristic assignment of the posterior distribution, thus strengthening the BNP approach of \citet{Cai(18)}; ii) it avoids the use of the peculiar restriction property of the DP, thus paving the way to the use of more general classes of prior distributions than the sole DP prior.

\begin{theorem}\label{teo_direct}
For $m\geq1$, let $x_{1:m}$ be a stream of tokens that are modeled as a random sample $X_{1:m}$ from $P\sim\text{DP}(\theta;\nu)$, and let $X_{m+1}$ be an additional random sample from $P$. Moreover, let $h_{n}$ be a random hash function distributed as the strong universal family $\mathcal{H}$, and let $h_{n}$ be independent of $X_{1:m}$ for any $m\geq1$, that is $h_{n}$ is independent of $P$. Then, for $l=0,1,\ldots,c_{n}$
\begin{align}\label{eq_direct}
&\text{Pr}[f_{X_{m+1}} = l\,|\, C_{n, h_{n}(X_{m+1})}=c_{n}]=\frac{\theta}{J}\frac{(c_{n}-l+1)_{(l)}}{(\frac{\theta}{J}+c_{n}-l)_{(l+1)}}.
\end{align}
\end{theorem}

\begin{proof}
The proof consists of three steps: i) evaluate the numerator of \eqref{eq:maintoprove}; ii) evaluate the denominator of \eqref{eq:maintoprove}; iii) evaluate \eqref{eq:maintoprove} with respect to what obtained in step i) and step ii). First, we observe that the independence between $h_{n}$ and $X_{1:m}$ allows us to invoke the ``freezing lemma'' \citep[Lemma 4.1]{Bal(17)}, according to which we can treat $h_{n}$ as it was fixed, i.e. non-random. To simplify the notation, we remove the subscript $n$ from $h_{n}$ and $c_{n}$. We start with the denominator of \eqref{eq:maintoprove}. Uniformity of $h$ implies that $h$ induces a (fixed) $J$-partition $\{B_{1},\ldots,B_{J}\}$ of $\mathcal{V}$ such that $B_{j}=\{v\in\mathcal{V}\text{ : }h(v)=j\}$ and $\nu(B_{j})=J^{-1}$ for $j=1,\ldots,J$. Then, the finite-dimensional projective property of the DP implies that $P(B_{j})$ is distributed as a Beta distribution with parameter $(\theta/J,\theta(1-1/J))$ for $j=1,\ldots,J$. Hence, we write
\begin{align}\label{eq:denom1}
&\text{Pr}\left[\sum_{i=1}^{m}\mathbbm{1}_{\{h(X_{i})\}}(h(X_{m+1}))=c\right]\\
&\notag\quad=J{m\choose c}\E[(P(B_{j}))^{c+1}(1-P(B_{j}))^{m-c}]\\
&\notag\quad=J{m\choose c}\int_{0}^{1}p^{c+1}(1-p)^{m-c}\frac{\Gamma(\theta)}{\Gamma(\frac{\theta}{J})\Gamma(\theta-\frac{\theta}{J})}p^{\frac{\theta}{J}-1}(1-p)^{\theta-\frac{\theta}{J}-1}\ddr p\\
&\notag\quad=J{m\choose c}\frac{\Gamma(\theta)}{\Gamma(\frac{\theta}{J})\Gamma(\theta-\frac{\theta}{J})}\frac{\Gamma(\frac{\theta}{J}+c+1)\Gamma(\theta-\frac{\theta}{J}+m-c)}{\Gamma(\theta+m+1)}.
\end{align}
This completes the study of the denominator of \eqref{eq:maintoprove}. Now, we consider the numerator of \eqref{eq:maintoprove}. Let us define the event $B(m,l)=\{X_{1}=\cdots=X_{l}=X_{m+1},\{X_{l+1},\ldots,X_{m}\}\cap\{X_{m+1}\}=\emptyset\}$. Then,
\begin{align}\label{eq:num1_1}
&\text{Pr}\left[f_{X_{m+1}}=l,\sum_{i=1}^{m}\mathbbm{1}_{\{h(X_{i})\}}(h(X_{m+1}))=c\right]\\
&\notag={m\choose l}\text{Pr}\Bigg[B(m,l),\,\sum_{i=1}^{m}\mathbbm{1}_{\{h(X_{i})\}}(h(X_{m+1}))=c\Bigg]\\
&\notag={m\choose l}\text{Pr}\Bigg[B(m,l),\,\sum_{i=l+1}^{m}\mathbbm{1}_{\{h(X_{i})\}}(h(X_{m+1}))=c-l\Bigg].
\end{align}
That is, the distribution of $(f_{X_{m+1}},C_{j})$ is completely determined by the distribution of the random variables $(X_{1},\ldots,X_{m+1})$. Let $\Pi(s,k)$ denote the set of all possible partitions of the set $\{1,\ldots,s\}$ into $k$ disjoints subsets $\pi_{1},\ldots,\pi_{k}$ such that $n_{i}$ is the cardinality of $\pi_{i}$. In particular, from \citet[Equation 3.5]{San(06)}, for any measurable $A_{1},\ldots,A_{m+1}$ we have that
\begin{displaymath}
\text{Pr}[X_{1}\in A_{1},\ldots,X_{m+1}\in A_{m+1}]=\sum_{k=1}^{m+1}\frac{\theta^{k}}{(\theta)_{(m+1)}}\sum_{(\pi_{1},\ldots,\pi_{k})\in\Pi(n+1,k)}\prod_{i=1}^{k}(n_{i}-1)!\nu(\cap_{m\in\pi_{i}}A_{m})
\end{displaymath}
for $m\geq1$. Let $\mathscr{V}$ be the Borel $\sigma$-algebra of $\mathcal{V}$. Let $\nu_{\pi_{1},\ldots,\pi_{k}}$ be a probability measure on $(\mathcal{V}^{m+1},\mathscr{V}^{m+1})$ defined as 
\begin{displaymath}
\nu_{\pi_{1},\ldots,\pi_{k}}(A_{1}\times\cdots\times A_{m+1})=\prod_{1\leq i\leq k}\nu(\cap_{m\in\pi_{i}}A_{m}),
\end{displaymath}
and attaching to $B(m,l)$ a value that is either $0$ or $1$. In particular, $\nu_{\pi_{1},\ldots,\pi_{k}}(B(m,l))=1$ if and only if one of the $\pi_{i}$'s is equal to the set $\{1,\ldots,l,m+1\}$. Hence, based on the measure $\nu_{\pi_{1},\ldots,\pi_{k}}$, we write
\begin{align*}
&\text{Pr}\left[B(m,l),\sum_{i=l+1}^{m}\mathbbm{1}_{\{h(X_{i})\}}(h(X_{m+1}))=c-l\right]\\
&\quad=\sum_{k=2}^{m-l+1}\frac{\theta^{k}}{(\theta)_{(m+1)}}\sum_{(\pi_{1},\ldots,\pi_{k-1})\in\Pi(m-l,k-1)}l!\prod_{i=1}^{k-1}(n_{i}-1)!\nu_{\pi_{1},\ldots,\pi_{k}}\left(\sum_{i=l+1}^{m}\mathbbm{1}_{\{h(X_{i})\}}(h(X_{m+1}))=c-l\right)\\
&\quad=\theta\frac{(\theta)_{(m-l)}}{(\theta)_{(m+1)}}l!\\
&\quad\quad\times\sum_{r=1}^{m-l}\frac{\theta^{r}}{(\theta)_{(m-l)}}\sum_{(\pi_{1},\ldots,\pi_{r})\in\Pi(m-l,r)}\prod_{i=1}^{r}(n_{i}-1)!\nu_{\pi_{1},\ldots,\pi_{r}}\left(\sum_{i=1}^{m-l}\mathbbm{1}_{\{h(X_{i})\}}(h(X_{m+1}))=c-l\right).
\end{align*}
Now, 
\begin{displaymath}
\sum_{r=1}^{m-l}\frac{\theta^{r}}{(\theta)_{(m-l)}}\sum_{(\pi_{1},\ldots,\pi_{r})\in\Pi(m-l,r)}\prod_{i=1}^{r}(n_{i}-1)!\nu_{\pi_{1},\ldots,\pi_{r}}\left(\cdot\right)
\end{displaymath}
is the distribution of a random sample $(X_{1},\ldots,X_{m-l})$ under $P\sim\text{DP}(\theta;\nu)$. Again, the distribution of $(X_{1},\ldots,X_{m-l})$ is given in \citet[Equation 3.5]{San(06)}. Using the fact that $P(B_{j})$ is distributed as a Beta distribution with parameter $(\theta/J,\theta(1-1/J))$, for $j=1,\ldots,J$, we write
\begin{align*}
&\text{Pr}\left[B(m,l),\sum_{i=l+1}^{m}\mathbbm{1}_{\{h(X_{i})\}}(h(X_{m+1}))=c-l\right]\\
&\quad=\theta\frac{(\theta)_{(m-l)}}{(\theta)_{(m+1)}}l!\\
&\quad\quad\times\sum_{r=1}^{m-l}\frac{\theta^{r}}{(\theta)_{(m-l)}}\sum_{(\pi_{1},\ldots,\pi_{r})\in\Pi(m-l,r)}\prod_{i=1}^{r}(n_{i}-1)!\nu_{\pi_{1},\ldots,\pi_{r}}\left(\sum_{i=1}^{m-l}\mathbbm{1}_{\{h(X_{i})\}}(h(X_{m+1}))=c-l\right)\\
&\quad=\theta\frac{(\theta)_{(m-l)}}{(\theta)_{(m+1)}}l!{m-l\choose c-l}\E[(P(B_{j}))^{c-l}(1-P(B_{j}))^{m-c}]\\
&\quad=\theta\frac{(\theta)_{(m-l)}}{(\theta)_{(m+1)}}l!{m-l\choose c-l}\int_{0}^{1}p^{c-l}(1-p)^{m-c}\frac{\Gamma(\theta)}{\Gamma(\frac{\theta}{J})\Gamma(\theta-\frac{\theta}{J})}p^{\frac{\theta}{J}-1}(1-p)^{\theta-\frac{\theta}{J}-1}\ddr p\\
&\quad=\theta\frac{(\theta)_{(m-l)}}{(\theta)_{(m+1)}}l!{m-l\choose c-l}\frac{\Gamma(\theta)}{\Gamma(\frac{\theta}{J})\Gamma(\theta-\frac{\theta}{J})}\frac{\Gamma(\frac{\theta}{J}+c-l)\Gamma(\theta-\frac{\theta}{J}+m-c)}{\Gamma(\theta+m-l)}, 
\end{align*}
where the second identity follows from an application of \citet[Proposition 3.1]{San(06)} under the DP prior; see also the formule displayed at page 469 of \citet{San(06)}. From \eqref{eq:num1_1} we write
\begin{align}\label{eq:num1_2}
&\text{Pr}\left[f_{X_{m+1}}=l,\sum_{i=1}^{m}\mathbbm{1}_{\{h(X_{i})\}}(h(X_{m+1}))=c\right]\\
&\notag\quad={m\choose l}\theta\frac{(\theta)_{(m-l)}}{(\theta)_{(m+1)}}l!{m-l\choose c-l}\frac{\Gamma(\theta)}{\Gamma(\frac{\theta}{J})\Gamma(\theta-\frac{\theta}{J})}\frac{\Gamma(\frac{\theta}{J}+c-l)\Gamma(\theta-\frac{\theta}{J}+m-c)}{\Gamma(\theta+m-l)}\\
&\notag\quad=\theta\frac{m!}{(c-l)!(m-c)!}\frac{\Gamma(\theta)}{\Gamma(\frac{\theta}{J})\Gamma(\theta-\frac{\theta}{J})}\frac{\Gamma(\frac{\theta}{J}+c-l)\Gamma(\theta-\frac{\theta}{J}+m-c)}{\Gamma(\theta+m+1)}.
\end{align}
This completes the study of the numerator of \eqref{eq:maintoprove}. Then, by combining \eqref{eq:num1_2} and \eqref{eq:denom1}, for $l=0,1,\ldots,m$
\begin{align}\label{eq:final_1}
&\text{Pr}[f_{X_{m+1}}=l\,|\,C_{h(X_{m+1})}=c]\\
&\notag\quad=\frac{\theta\frac{m!}{(c-l)!(m-c)!}\frac{\Gamma(\theta)}{\Gamma(\frac{\theta}{J})\Gamma(\theta-\frac{\theta}{J})}\frac{\Gamma(\frac{\theta}{J}+c-l)\Gamma(\theta-\frac{\theta}{J}+m-c)}{\Gamma(\theta+m+1)}}{J{m\choose c}\frac{\Gamma(\theta)}{\Gamma(\frac{\theta}{J})\Gamma(\theta-\frac{\theta}{J})}\frac{\Gamma(\frac{\theta}{J}+c+1)\Gamma(\theta-\frac{\theta}{J}+m-c)}{\Gamma(\theta+m+1)}}\\
&\notag\quad=\frac{\theta}{J}\frac{(c-l+1)_{(l)}}{(\frac{\theta}{J}+c-l)_{(l+1)}},
\end{align}
which follows directly from the (regular) conditional distribution \eqref{eq:maintoprove}, whose denominator and numerator are replaced by Equation \eqref{eq:denom1} and Equation \eqref{eq:num1_2}, respectively. The proof is completed.
\end{proof}

The next proposition is an interesting complement to Theorem \ref{teo_direct}: i) it provides a novel (constructive) representation of the posterior distribution \eqref{eq_direct} in terms of a mixture of Binomial distributions; ii) it characterizes the large $c_{m}$ asymptotic behaviour of the posterior distribution \eqref{eq_direct}. Hereafter, we denote by $\stackrel{\text{w}}{\longrightarrow}$ the convergence in distribution or weak convergence.

\begin{proposition}\label{more_cmsdp}
Let $B_{a,b}$ be a Beta random variable, and denote by $f_{B_{a,b}}$ its density function. Under the setting of Theorem \ref{teo_direct}, if $F_{X_{m+1}}$ is a random variable distributed as \eqref{eq_direct}, then as $c_{n}\rightarrow+\infty$
\begin{equation}\label{distr_asym}
\frac{F_{X_{m+1}}}{c_{n}}\stackrel{\text{w}}{\longrightarrow}B_{1,\frac{\theta}{J}}.
\end{equation}
Moreover, if $\text{Binomial}\,(n,p)$ denotes the Binomial distribution with parameter $(n,p)$, then for $l=0,1,\ldots,c_{n}$
\begin{equation}\label{distr_ident}
\text{Pr}[f_{X_{m+1}} = l\,|\, C_{n, h_{n}(X_{m+1})}=c_{n}]=\int_{0}^{1}\text{Binomial}\,(l;c_{n},p)f_{B_{1,\frac{\theta}{J}}}(p)\ddr p.
\end{equation}
\end{proposition}

See Appendix \ref{proof_binomial} for the proof of Proposition \ref{more_cmsdp}. Let  $F_{X_{m+1}}$ be a random variable whose distribution coincides with the posterior distribution \eqref{eq_direct}. Then, Equation \eqref{distr_ident} shows that the distribution of $F_{X_{m+1}}$ is a mixture of Binomial distributions, with the mixing distribution on the success probability being a Beta distribution with parameter $(1,\theta/J)$. That is, the posterior distribution \eqref{eq_direct} admits a straightforward representation in terms of a Beta-Binomial distribution with parameter $(c_{n},1,\theta/J)$ \citep[Chapter 6]{Joh(05)}. Moreover, Equation \eqref{distr_asym} shows that the mixing distribution is precisely the limiting distribution of the proportion $c_{n}^{-1}F_{X_{m+1}}$ as $c_{n}\rightarrow+\infty$. Then, according to Proposition \ref{more_cmsdp}, we write $F_{X_{m+1}}=\sum_{1\leq i\leq c_{n}}Z_{i}$, where, by means of de Finetti's representation theorem, $(Z_{i})_{i\geq1}$ is an exchangeable sequence of Bernoulli random variables with de Finetti's measure being the Beta distribution with parameter $(1,\theta/J)$. Concerning large $m$ behavior, Berry-Esseen estimates for the vicinity of the (rescaled) law of $F_{X_{m+1}}$ to the aforesaid de Finetti's measure are contained in \citet{DF(20b)}, while other similar estimates can be found in \citet{Do(13)}.  
Then, for a collection of hash functions $h_{1},\ldots,h_{N}$, one may combine Proposition \ref{more_cmsdp} with Equation \eqref{post_full_dp} to obtain an alternative representation, in terms of product of Beta-Binomial distributions, of the posterior distribution of $f_{X_{m+1}}$, given $\{C_{n, h_{n}(X_{m+1})}\}_{n\in[N]}$. Despite Proposition \ref{more_cmsdp} has not a direct impact with respect to the implementation of the CMS-DP, in the sense of improving its computation, we believe it is of interest in shedding light on distributional properties of the posterior distribution of $f_{X_{m+1}}$, given $C_{h_{n}(X_{m+1})}$.  


\section{A learning-augmented CMS under power-law streams}\label{sec3}

The ``Bayesian" proof of Section \ref{sec2} paves the way to extend the BNP approach of \citet{Cai(18)} to more general classes of (discrete) nonparametric priors than the DP prior, thus leading to introduce novel learning-augmented CMSs. In principle, any prior arising from the normalization of completely random measures \citep[Chapter 4]{Pit(06)} can be applied within the general setting of the ``Bayesian" proof. Here, we consider the problem of developing a learning-augmented CMS under power-law streams of tokens, and therefore it is natural to focus on priors featuring a power-law tail behaviour. In particular, we assume the stream $x_{1:m}$ to be modeled as a random sample from an unknown discrete distribution $P$, which is endowed with a PYP prior $\mathcal{Q}$. Within (discrete) nonparametric priors with power-law tail behaviour, the PYP prior stands out for both its mathematical tractability, flexibility and interpretability, and hence it is the natural candidate for applications \citep{Deb(15),Bac(17)}. We recall that the PYP has neither a restriction property nor a ``sufficientness" postulate analogous to those featured by the DP, and therefore the ``constructive" proof of \citet{Cai(18)} cannot be applied to obtain the posterior distribution of a point query. Moreover, we recall that the PYP does not feature a finite-dimensional projective property analogous to that of the DP, and therefore prior's parameters cannot be estimated through an empirical Bayes procedure, as discussed in \citet{Cai(18)}, or through a hierarchical (fully) Bayes procedure. In this section, we adapt the "Bayesian" proof of Section \ref{sec2} in order to compute the posterior distribution of the point query $f_{X_{m+1}}$, given the hashed frequencies $\{C_{n, h_{n}(X_{m+1})}\}_{n\in[N]}$, under a PYP prior. Then, we exploit the predictive distribution of the PYP prior to implement a likelihood-free approach to estimate the PYP prior's parameters. Our work leads to a generalization of the CMS-DP, referred to as the CMS-PYP, which is a novel learning-augmented CMS under power-law streams.

\subsection{The CMS-PYP}

\subsubsection{The PYP prior}

A simple and intuitive definition of the PYP follows from its stick-breaking construction \citep{Per(92),Pit(95),Pit(97)}. For $\alpha\in[0,1)$ and $\theta>-\alpha$ let: i) $(B_{i})_{i\geq1}$ be independent random variables such that $B_{i}$ is distributed as a Beta distribution with parameter $(1-\alpha,\theta+i\alpha)$; ii) $(V_{i})_{i\geq1}$ random variables, independent of $(B_{i})_{i\geq1}$, and i.i.d. from a non-atomic distribution $\nu$ on $\mathcal{V}$. Then, define $P_{1}=B_{1}$ and $P_{j}=B_{j}\prod_{1\leq i\leq j-1}(1-B_{i})$ for $j\geq2$, in such a way that $\sum_{i\geq1}P_{i}=1$ almost surely. The (discrete) random probability measure $P=\sum_{j\geq1}P_{j}\delta_{V_{j}}$ is a PYP on $\mathcal{V}$ with (base) distribution $\nu$, discount parameter $\alpha$ and mass parameter $\theta$. For short, we write $P\sim\text{PYP}(\alpha,\theta;\nu)$. We refer to \citet{Per(92)} and \citet{Pit(97)} for an alternative definition of the PYP through a suitable transformation of the $\alpha$-stable completely random measure \citet{Kin(93)}. See also \citet[Chapter 4]{Pit(06)} and references therein. The DP arises as a special case of the PYP by setting $\alpha=0$. For the purposes of the present paper, it is useful to recall the power-law tail behaviour featured by the PYP prior. In particular, let $P\sim\text{PYP}(\alpha,\theta;\nu)$ with $\alpha\in(0,1)$, and  let $(P_{(j)})_{j\geq1}$ be the decreasing ordered random probabilities $P_{j}$'s of $P$ \citep[Chapter 3]{Pit(06)}. Then, as $j\rightarrow+\infty$ the $P_{(j)}$'s follow a power-law distribution of exponent $c=\alpha^{-1}$ \citep{Pit(97)}. That is, $\alpha\in(0,1)$ controls the power-law tail behaviour of the PYP through small probabilities $P_{(j)}$'s: the larger $\alpha$ the heavier the tail of $P$. See also \citet[Section 10]{Gne(07)} for a detailed account on the tail behaviour of the PYP prior.

As for the DP, the discreteness of $P\sim\text{PYP}(\alpha,\theta;\nu)$ implies that a random sample $X_{1:m}=(X_{1},\ldots,X_{m})$ from $P$ induces a random partition of the set $\{1,\ldots,m\}$ into $1\leq K_{m}\leq m$ partition subsets, labelled by distinct types $\mathbf{v}=\{v_1,\ldots,v_{K_{m}}\}$, with corresponding frequencies $(N_{1,m},\ldots,N_{K_{m},m})$ such that $1\leq N_{i,m}\leq n$ and $\sum_{1\leq i\leq K_{m}}N_{i,m}=m$. For $1\leq l\leq m$ let $M_{l,m}$ be the number of distinct types with frequency $l$, i.e. $M_{l,m}=\sum_{1\leq i\leq K_{m}} \mathbbm{1}_{\{N_{i,m}\}}(l)$ such that $\sum_{1\leq l\leq m}M_{l,m}=K_{m}$ and $\sum_{1\leq l\leq m}lM_{l,m}=m$. The distribution of $\mathbf{M}_{m}$ is
\begin{equation}\label{eq_ewe_py}
\text{Pr}[\mathbf{M}_{m}=\mathbf{m}]=m!\frac{\left(\frac{\theta}{\alpha}\right)_{(k)}}{(\theta)_{(m)}}\prod_{i=1}^{m}\left(\frac{\alpha(1-\alpha)_{(i-1)}}{i!}\right)^{m_{i}}\frac{1}{m_{i}!}\mathbbm{1}_{\mathcal{M}_{m,k}}(\mathbf{m}),
\end{equation}
such that
\begin{equation}\label{eq:distpy}
\text{Pr}[K_{m}=k]=\frac{\left(\frac{\theta}{\alpha}\right)_{(k)}}{(\theta)_{(m)}}\mathscr{C}(m,k;\alpha)
\end{equation}
for $k=1,\ldots,m$, where $\mathscr{C}(m,k;\alpha)=(k!)^{-1}\sum_{0\leq i\leq k}{k\choose i}(-1)^{i}(-i\alpha)_{(m)}$ denotes the generalized factorial coefficient \citep{Cha(05)}, with the proviso that $\mathscr{C}(0,0;\alpha)=1$ and $\mathscr{C}(m,0;\alpha)=0$. See \citet[Chapter 3]{Pit(06)} for details on \eqref{eq_ewe_py} and on \eqref{eq:distpy}. Now, let $\mathbf{v}_{l}=\{v_{i}\in \mathbf{v}\text{ : } N_{i,m}=l\}$, i.e. the labels of types with frequency $l$ and let $\mathbf{v}_{0}=\mathcal{V}-\mathbf{v}$, i.e. the labels of types not belonging to $\mathbf{v}$. The predictive distribution induced by $P\sim\text{PYP}(\alpha,\theta;\nu)$ is
\begin{align} \label{eq:pred_seen_py}
\text{Pr}[X_{m+1} \in \mathbf{v}_{l}\,|\, X_{1:m}] =\text{Pr}[X_{m+1} \in \mathbf{v}_{l}\,|\, \mathbf{M}_{m}=\mathbf{m}]=
\begin{cases} 
 \frac{\theta+k\alpha}{\theta+m}&\mbox{ if } l=0\\[0.4cm] 
\frac{m_{l}(l-\alpha)}{\theta+m}&\mbox{ if } l\geq1,
\end{cases}
\end{align}
for $m\geq1$. In particular, the PYP prior is characterized as the sole (discrete) nonparametric prior for which: i) the conditional probability that $X_{m+1}$ belongs to $\mathbf{v}_{0}$, given $X_{1:m}$m depends on $X_{1:m}$ only through $m$ and $K_{m}$; ii) the conditional probability that $X_{m+1}$ belongs to $\mathbf{v}_{l}$, given $X_{1:m}$, depends on $X_{1:m}$ only through $m$ and $M_{l,m}$  \citep[Proposition 1]{Bac(17)}.

\begin{figure}[!htb]
    \centering
    \includegraphics[width=1\linewidth]{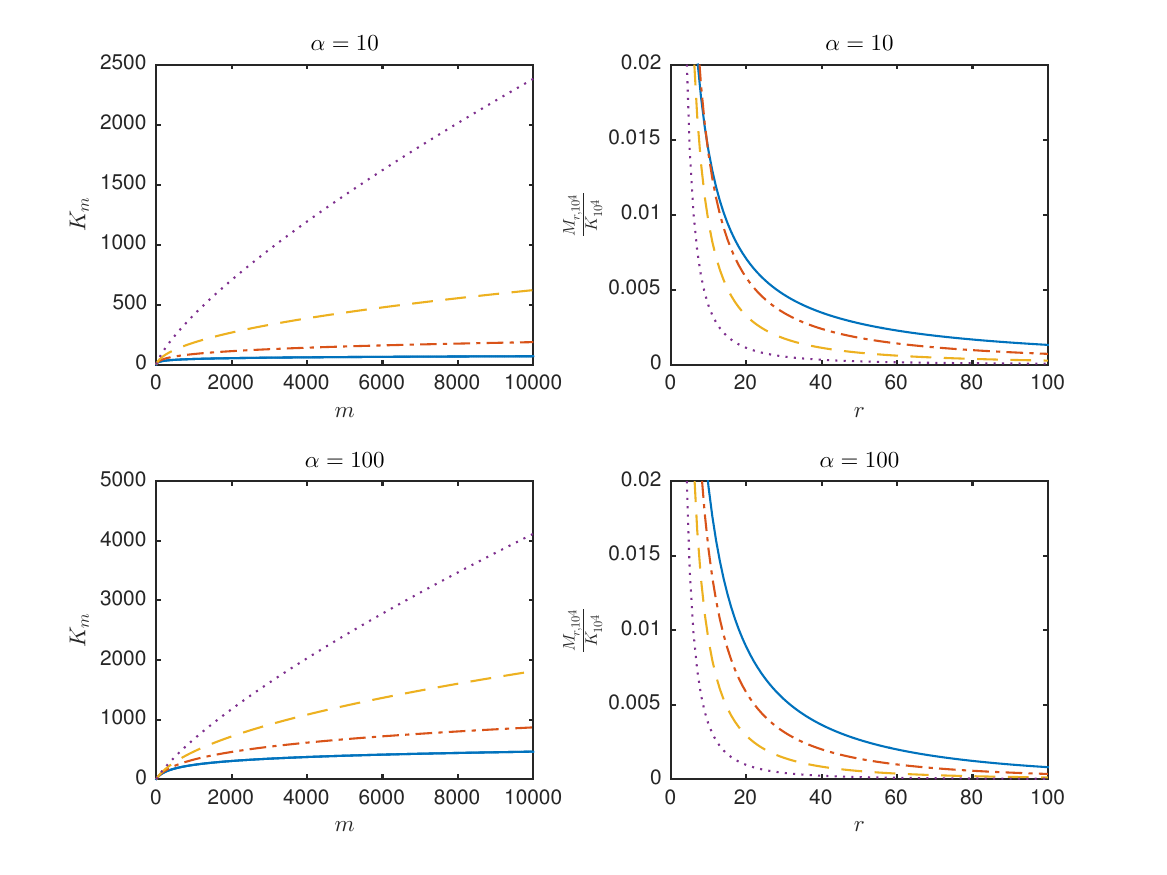}
    \caption{\small{Behaviours in $m\geq1$ of the statistics $K_{m}$ and $M_{r,m}/K_{m}$, for $1\leq m\leq 10^{4}$ under $P\sim\text{PYP}(\alpha,\theta;\nu)$: $\alpha=0$ (blue -), $\alpha=.25$ (red -.), $\alpha=.5$ (yellow --) and $\alpha=.75$ (purple :).}}
    \label{fig:draws}
\end{figure}

At the sampling level, the power-law tail behaviour of $P\sim\text{PYP}(\alpha,\theta;\nu)$ emerges from the analysis of the large $m$ asymptotic behaviour of $K_{m}$ and $M_{r,m}/K_{m}$ \citep[Chapter 3]{Pit(06)}. Let $X_{1:m}$ be a random sample from $P$.  \citet[Theorem 3.8]{Pit(06)} shows that, as $m\rightarrow+\infty$,
\begin{align} \label{eq:sigma_diversity}
\frac{K_{m}}{m^{\alpha}}\stackrel{\text{a.s}}{\longrightarrow} S^{-\alpha}_{\alpha,\theta},
\end{align}
where $S_{\alpha,\theta}$ is a polynomially tilted $\alpha$-stable random variable, that is the distribution of $S_{\alpha,\theta}$ has density function $f_{S_{\alpha,\theta}}(x)\propto x^{-\theta}g_{\alpha}(x)\mathbbm{1}_{\mathbb{R}^{+}}(x)$ for $g_{\alpha}$ being the positive $\alpha$-stable density function. According to \eqref{eq:sigma_diversity}, it holds $K_{n}\approx m^{\alpha}S^{-\alpha}_{\alpha,\theta}$ for large $m$, or equivalently $K_{n}\approx [(\theta+m)^{\alpha}-\theta^{\alpha}]S^{-\alpha}_{\alpha,\theta}$ for large $m$ \citep{Fav(09)}. It follows from \eqref{eq:sigma_diversity} that, as $m\rightarrow+\infty$, 
\begin{align} \label{eq:sigma_diversity_l}
    \frac{M_{l,m}}{K_{m}}\stackrel{\text{a.s}}{\longrightarrow}\frac{\alpha(1-\alpha)_{(l-1)}}{l!}.
\end{align}
Equation \eqref{eq:sigma_diversity} shows that the number $K_{m}$ of distinct types in $X_{1:m}$, for large $m$, grows as $m^{\alpha}$. This is precisely the growth of the number of distinct types in $m\geq1$ random samples from a power-law distribution of exponent $c=\alpha^{-1}$. Moreover, Equation \eqref{eq:sigma_diversity_l} shows that $p_{\alpha,l}=\alpha(1-\alpha)_{(l-1)}/l!$ is the large $m$ asymptotic proportion of the number of distinct types with frequency $l$ in $X_{1:m}$. Then, it holds $ p_{\alpha,l}\approx c_{\alpha}l^{-\alpha -1}$ for large $l$, for a constant $c_{\alpha}$. This is precisely the distribution of the number of distinct types with frequency $l$  in $m\geq1$ random samples from a power-law distribution of exponent $c=\alpha^{-1}$. See Figure \ref{fig:draws} for an illustration of the large $m$ behaviour of $K_{m}$ and $M_{l,n}$ under the PYP prior, for some choices of the parameter $(\alpha,\theta)$.

\subsubsection{The CMS-PYP}\label{cms_pyp}

To introduce the CMS-PYP, we assume that the stream $x_{1:m}$ is modeled as random samples $X_{1:m}$ from an unknown discrete distribution $P$, which is endowed with a PYP prior. That is,
    \begin{align}\label{eq:bnp_py}
    X_{1:m}\,|\, P  &\,\stackrel{\mbox{\scriptsize{iid}}}{\sim}\,P\\[0.2cm]
     \notag P &\, \sim\,\text{PYP}(\alpha,\theta;\nu)
    \end{align}
for $m\geq1$. Let $h_{1},\ldots,h_{N}$ be a collection of random hash functions that are i.i.d. from the strong universal family $\mathcal{H}$, and assume that $h_{1},\ldots,h_{N}$ are independent of $X_{1:m}$ for any $m\geq1$; in particular, by de Finetti's representation theorem, $h_{1},\ldots,h_{N}$ are independent of $P \sim\,\text{PYP}(\alpha,\theta;\nu)$. Under the CMS-PYP the $X_{i}$'s are hashed through $h_{1},\ldots,h_{N}$, thus creating $\{(C_{n,1},\ldots,C_{n,J})\}_{n\in[N]}$, and estimates of the point query $f_{X_{m+1}}$, with $X_{m+1}$ being of an arbitrary type $v\in\mathcal{V}$, are obtained as functionals of the posterior distribution of $f_{X_{m+1}}$ given the hashed frequencies $\{C_{n, h_{n}(X_{m+1})}\}_{n\in[N]}$. As for the derivation of the CMS-DP in Section \ref{sec2}, the assumption of independence between the $h_{n}$'s and $X_{1:m}$ plays a critical role to obtain the posterior distribution of $f_{X_{m+1}}$ given $\{C_{n, h_{n}(X_{m+1})}\}_{n\in[N]}$; that is, it allows us to treat the $h_{n}$'s as they were fixed, i.e. non-random hash functions. For a single hash function $h_{n}$, in the next theorem we provide a rigorous derivation of the posterior distribution of $f_{X_{m+1}}$, given $C_{n, h_{n}(X_{m+1})}$.

\begin{theorem}\label{teo_pyp}
For $m\geq1$, let $x_{1:m}$ be a stream of tokens that are modeled as a random sample $X_{1:m}$ from $P\sim\text{PYP}(\alpha,\theta;\nu)$, and let $X_{m+1}$ be an additional random sample from $P$. Moreover, let $h_{n}$ be a random hash function distributed as the strong universal family $\mathcal{H}$, and let $h_{n}$ be independent of $X_{1:m}$ for any $m\geq1$, that is $h_{n}$ is independent of $P$. Then, for $l=0,1,\ldots,c_{n}$
\begin{align}\label{eq:posterior_py}
&p_{f_{X_{m+1}}}(l;m,c_{n},\alpha,\theta)\\
    &\notag\quad:=\text{Pr}[f_{X_{m+1}} = l\,|\, C_{n, h_{n}(X_{m+1})}=c_{n}]\\
    &\notag\quad=\frac{\theta}{J}{c_{n}\choose l}(1-\alpha)_{(l)}\frac{\sum_{i=0}^{c_{n}-l}\sum_{j=0}^{m-c_{n}}\left(\frac{\theta+\alpha}{\alpha}\right)_{(i+j)}\left(\frac{1}{J}\right)^{i}\left(1-\frac{1}{J}\right)^{j}\mathscr{C}(c_{n}-l,i;\alpha)\mathscr{C}(m-c_{n},j;\alpha)}{\sum_{i=0}^{c_{n}+1}\sum_{j=0}^{m-c_{n}}\left(\frac{\theta}{\alpha}\right)_{(i+j)}\left(\frac{1}{J}\right)^{i}\left(1-\frac{1}{J}\right)^{j}\mathscr{C}(c_{n}+1,i;\alpha)\mathscr{C}(m-c_{n},j;\alpha)}.
\end{align}
\end{theorem}

\begin{figure}[!htb]
\vspace{0.3cm}
    \centering
    \includegraphics[width=1\linewidth]{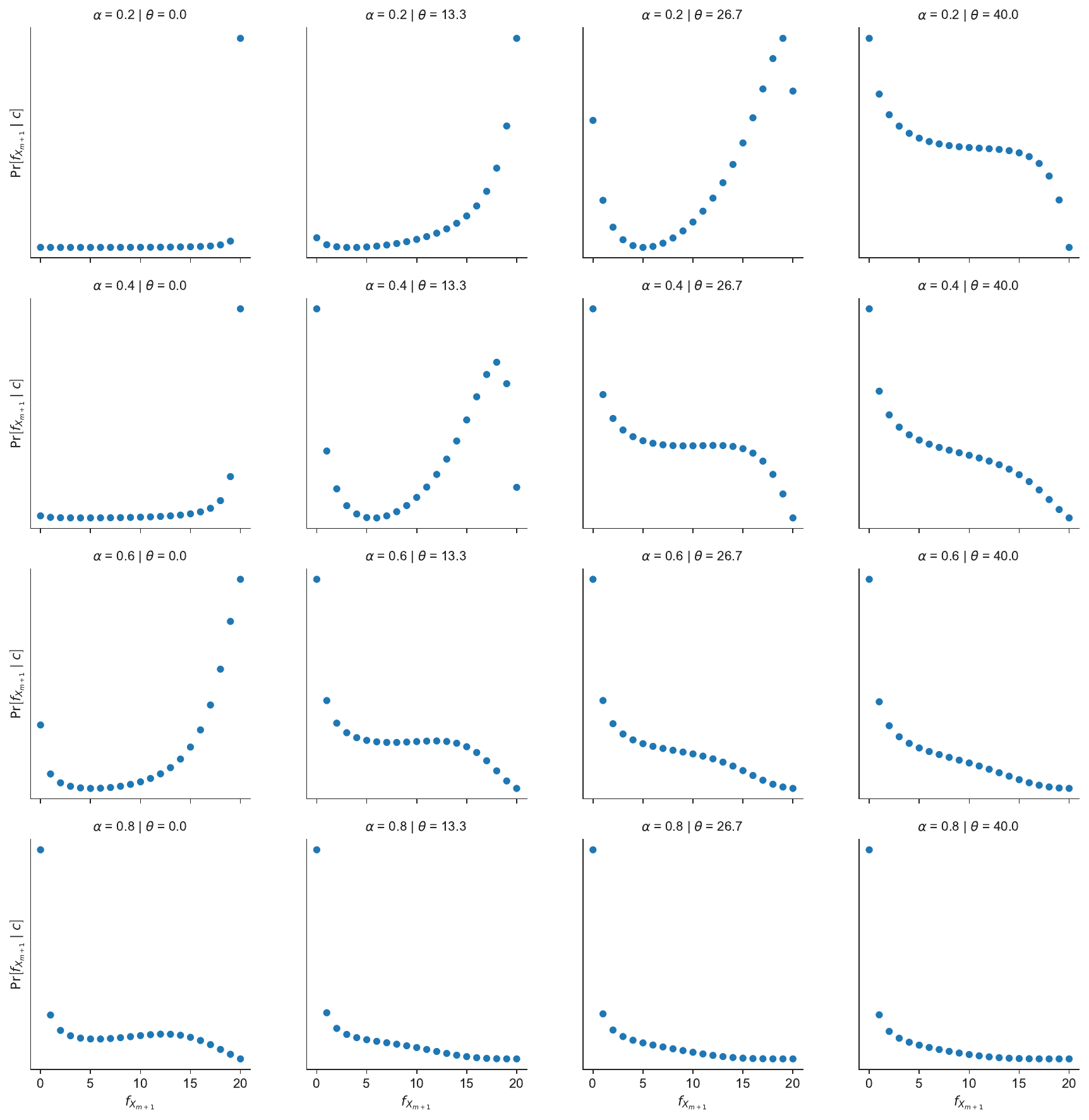}
    \caption{\small{Posterior distribution of $f_{X_{m+1}}$ given $C_{n, h_{n}(X_{m+1})}=c_{n}$ under $P\sim\text{PYP}(\alpha,\theta;\nu)$: $m = 1000$, $J = 50$ and $c_{n} = 20$.}}
    \label{fig:pmf}
\end{figure}

See Appendix \ref{proof_pyp} for the proof of Theorem \ref{teo_pyp}; note that the proof is along lines similar to the ``Bayesian" proof presented in Section \ref{sec2} under the DP prior. Theorem \ref{teo_pyp} provides an extension of Theorem \ref{teo_direct} to the more general BNP model \eqref{eq:bnp_py}; in particular, Theorem \ref{teo_direct} can be recovered from Theorem \ref{teo_pyp} by setting $\alpha=0$. See Appendix \ref{sec4app} for details. For $\alpha\in[0,1)$, an alternative expression for \eqref{eq:posterior_py} may be given in terms of the distribution \eqref{eq:distpy} of the number $K_{m}$ of distinct types in a random sample from the PYP. If $c_{n}>0$, then for $l=0,1,\ldots,c_{n}$
\begin{align}\label{eq:posterior_py_alternative}
&p_{f_{X_{m+1}}}(l;m,c_{n},\alpha,\theta)\\
&\notag\quad=\frac{\theta}{J}{c_{n}\choose l}(1-\alpha)_{(l)}\frac{(\theta)_{(c_{n}-l)}\E\left[\frac{\left(\frac{\theta+\alpha}{\alpha}\right)_{(K_{c_{n}-l}+K_{m-c_{n}})}}{\left(\frac{\theta}{\alpha}\right)_{(K_{c_{n}-l})}\left(\frac{\theta}{\alpha}\right)_{(K_{m-c_{n}})}}\left(\frac{1}{J}\right)^{K_{c_{n}-l}}\left(1-\frac{1}{J}\right)^{K_{m-c_{n}}}\right]}{(\theta)_{(c_{n}+1)}\E\left[\frac{\left(\frac{\theta}{\alpha}\right)_{(K_{c_{n}+1}+K_{m-c_{n}})}}{\left(\frac{\theta}{\alpha}\right)_{(K_{c_{n}+1})}\left(\frac{\theta}{\alpha}\right)_{(K_{m-c_{n}})}}\left(\frac{1}{J}\right)^{K_{c_{n}+1}}\left(1-\frac{1}{J}\right)^{K_{m-c_{n}}}\right]}
\end{align}
with the proviso that $K_{0}=0$, where $K_{c_{n}-l}$ and $K_{m-c_{n}}$ in the numerator of  \eqref{eq:posterior_py_alternative} are independent random variables for any $l=0,1,\ldots,c_{n}-1$, and $K_{c_{n}+1}$ and $K_{m-c_{n}}$ in the denominator of  \eqref{eq:posterior_py_alternative} are independent random variables. See Appendix \ref{sec3app} for the proof of Equation \eqref{eq:posterior_py_alternative}. Equation \eqref{eq:posterior_py_alternative} gives a probabilistic representation of the posterior distribution  \eqref{eq:posterior_py}, whose critical terms are the expected value of a suitable functional of $(K_{c_{n}-l},K_{m-c_{n}})$, i.e. the numerator of \eqref{eq:posterior_py_alternative}, and the expected value of a suitable functional of $(K_{c_{n}+1},K_{m-c_{n}})$, i.e. the denominator of \eqref{eq:posterior_py_alternative}. See Appendix \ref{sec3app1} for a further alternative expression of \eqref{eq:posterior_py}, which is in terms of  exponentially tilted $\alpha$-stable random variables \citep{Zol(86)}. Figure \ref{fig:pmf} shows the shape behaviour of the posterior distribution \eqref{eq:posterior_py_alternative} for different values of $(\alpha,\theta)$, keeping $m$ $J$ and $c_{n}$ fixed. For $\alpha=0$, i.e. under the DP prior, \citet{Cai(18)} showed that the posterior distribution of $f_{X_{m+1}}$, given $C_{n, h_{n}(X_{m+1})}$ is  monotonically decreasing or increasing. Under the PYP,  the additional parameter $\alpha\in(0,1)$ allows for a more flexible shape behaviour.

\begin{remark}
Equation \eqref{eq:posterior_py_alternative} is useful for the numerical evaluation of the posterior distribution \eqref{eq:posterior_py}, since it avoids numerical issues that arise in evaluating the generalized factorial coefficients. In particular, \eqref{eq:posterior_py_alternative} allows for a Monte Carlo (MC) evaluation of \eqref{eq:posterior_py}, which requires to sample from the random variable $K_{m}$, for suitable choices of $m$. Sampling $K_{m}$ is straightforward, and it exploits the predictive probabilities of the PYP. In particular, from \eqref{eq:pred_seen_py}, $\text{Bernoulli}(p)$ is the Bernoulli distribution with parameter $p$, for $p\in(0,1)$, then sampling $K_{m}$ reduces to sample $(m-1)$ Bernoulli random variables. See Algorithm 1 in Section \ref{sec5}. 
\end{remark}

Under the PYP prior, Theorem \ref{teo_pyp} shows that the posterior distribution of $f_{X_{m+1}}$, given $C_{n, h_{n}(X_{m+1})}$, depends on the sampling information through $c_{n}$ and $m$. This is a critical difference with respect to the DP prior, where Theorem \ref{teo_pyp} shows that the posterior distribution of $f_{X_{m+1}}$, given $C_{n, h_{n}(X_{m+1})}$, depends on the sampling information only through $m$. Therefore, under the PYP prior, one may consider different large $m$ asymptotic behaviours for the posterior distribution \eqref{eq:posterior_py}. Here, we start by considering a local limit theorem of \eqref{eq:posterior_py} for $m\rightarrow+\infty$, while $c_{n}$ is fixed. Under the setting of Theorem \ref{teo_pyp}, for  any $l=0,1,\ldots,c_{n}$ it holds
\begin{align}\label{local_limit}
p(l;c_{n},\alpha,\theta):=\lim_{m\rightarrow+\infty}p_{f_{X_{m+1}}}(l;m,c_{n},\alpha,\theta)={c_{n}\choose l}(1-\alpha)_{(l)}\frac{(\theta+2\alpha)_{(c_{n}-l)}}{(\theta+\alpha+1)_{(c_{n})}}
\end{align}
and
\begin{align}\label{local_bino}
p(l;c_{n},\alpha,\theta)=\int_{0}^{1}\text{Binomial}(l;c_{n},p)f_{B_{1-\alpha,\theta+2\alpha}}(p)\ddr p,
\end{align}
where $f_{B_{a,b}}$ is the density function of the distribution of a Beta random variable $B_{a,b}$. See Appendix \ref{app_local} for the proof of Equation \eqref{local_limit} and Equation \eqref{local_bino}. The next proposition is an interesting complement to Theorem \ref{teo_pyp}, providing the large $m$ asymptotic behaviour of the posterior distribution \eqref{eq:posterior_py}. In particular, we consider $m\rightarrow+\infty$ and $c_{n}\rightarrow+\infty$ with the assumption that $c_{n}=\lambda m$ for some choice of $\lambda\in(0,1)$. Such a large $m$ asymptotic behaviour is in line with the large $c_{n}$ asymptotic behaviour presented in Proposition \ref{more_cmsdp} under the DP prior.

\begin{proposition}\label{asymp_py}
For $\alpha\in(0,1)$ and $c>0$ let $S_{\alpha,c}$ be a polynomially tilted $\alpha$-stable random variable, i.e. the distribution of $S_{\alpha,c}$ has density function $f_{S_{\alpha,c}}(x)\propto x^{-c}g_{\alpha}(x)\mathbbm{1}_{\mathbb{R}^{+}}(x)$ for $g_{\alpha}$ being the positive $\alpha$-stable density function; moreover, set $Z_{\alpha,\theta+\alpha}=(J-1)^{1/\alpha}S_{\alpha,0}/S_{\alpha,\theta+\alpha}$
and $W_{\alpha,\theta}=(J-1)^{1/\alpha}S_{\alpha,0}/S_{\alpha,\theta}$, with $S_{\alpha,0}$ being independent of $S_{\alpha,\theta+\alpha}$ and of $S_{\alpha,\theta}$, and denote by $f_{Z_{\alpha,\theta+\alpha}}$ and $f_{W_{\alpha,\theta}}$ the density functions of the distributions of $Z_{\alpha,\theta+\alpha}$ and $W_{\alpha,\theta}$, respectively. Under the setting of Theorem \ref{teo_pyp}, let $F_{X_{m+1}}$ be a random variable with distribution \eqref{eq:posterior_py}. As $m\rightarrow+\infty$ and under the large $m$ asymptotic regime $c_{n}=\lambda m$, for some choice of $\lambda\in(0,1)$,
\begin{equation}\label{distr_asym_pyp}
\frac{F_{X_{m+1}}}{c_{n}}\stackrel{\text{w}}{\longrightarrow}B^{(\lambda)}_{1-\alpha,\theta+\alpha},
\end{equation}
where $B^{(\lambda)}_{1-\alpha,\theta+\alpha}$ is a random variable whose distribution has density function of the following form
\begin{displaymath}
f_{B^{(\lambda)}_{1-\alpha,\theta+\alpha}}(x)=\frac{\frac{\Gamma(\theta+1)}{\Gamma(\theta+\alpha)\Gamma(1-\alpha)}}{f_{W_{\alpha,\theta}}(\lambda^{-1}-1)}\frac{f_{Z_{\alpha,\theta+\alpha}}\left(\frac{\lambda^{-1}-1}{1-x}\right)}{1-x}x^{1-\alpha-1}(1-x)^{\theta+\alpha-1}\mathbbm{1}_{(0,1)}(x).
\end{displaymath}
\end{proposition}

See Appendix \ref{proof_asymp_py} for the proof of Proposition \ref{asymp_py}. As in the context of the DP prior discussed in Section \ref{sec2}, Proposition \ref{asymp_py} shows that the posterior distribution of $f_{X_{m+1}}$ given $C_{h_{n}(X_{m+1})}$ admits a representation in terms of a mixture of Binomial distributions. In particular, Proposition \ref{asymp_py} may be viewed as the natural counterpart of Proposition \ref{more_cmsdp}, though the resulting mixing distribution is not as simple as the Beta distribution of Proposition \ref{more_cmsdp}. For the collection of hash functions $h_{1},\ldots,h_{N}$, the posterior distribution of $f_{X_{m+1}}$, given $\{C_{n, h_{n}(X_{m+1})}\}_{n\in[N]}$, follows from  Theorem \ref{teo_pyp} by means of the assumption that the $h_{n}$'s are i.i.d. according to the strong universal family $\mathcal{H}$. In particular, by a direct application of Bayes theorem, straightforward calculations show that for $l=0,1,\ldots,\min\{c_{1},\ldots,c_{N}\}$ it holds that
\begin{align}\label{eq:posterior_py_compl}
    &\text{Pr}[f_{X_{m+1}} = l\,|\, \{C_{n, h_{n}(X_{m+1})}\}_{n\in[N]}=\{c_{n}\}_{n\in[N]}]=\frac{\prod_{n\in[N]}p_{f_{X_{m+1}}}(l;m,c_{n},\alpha,\theta)}{(p_{f_{X_{m+1}}}(l;m,\alpha,\theta))^{N-1}},
\end{align}
where
\begin{displaymath}
p_{f_{X_{m+1}}}(l;m,\alpha,\theta):=\text{Pr}[f_{X_{m+1}}=l]={m\choose l}(1-\alpha)_{(l)}\frac{(\theta+\alpha)_{(m-l)}}{(\theta+1)_{(m)}}
\end{displaymath}
for $l=0,1,\ldots,m$, where $f_{B_{1-\alpha,\theta+\alpha}}$ denotes the density function of the distribution of a Beta random variable with parameter $(1-\alpha,\theta+\alpha)$. See Appendix \ref{sec5app} for the proof of Equation \eqref{eq:posterior_py_compl}. CMS-PYP estimates of the point query $f_{X_{m+1}}$, with respect to a suitable choice of a loss function, are obtained as functionals of the posterior distribution \eqref{eq:posterior_py_compl}, e.g. posterior mode, mean and median. In general, the numerical evaluation of the posterior distribution \eqref{eq:posterior_py_compl}, as well as the evaluation of its alternative expression in terms of  \eqref{eq:posterior_py_alternative}, requires care in order to achieve numerical stability and efficiency, as it is discussed in the last part of this section.

To apply \eqref{eq:posterior_py_compl}, it remains to estimate the prior's parameter $(\alpha,\theta)$. For ease of exposition, we denote by $\mathbf{C}$ the $N \times J$ matrix with entries $C_{n,j}$ for $n\in[N]$ and $j\in[J]$. Assuming that the matrix $\mathbf{C}$ has been computed from $m$ tokens, the sum of the entries of each row of $\mathbf{C}$ is equal to the sample size $m$. Since the PYP does not have a restriction property analogous to that of the DP, under the BNP model \eqref{eq:bnp_py} the distribution of $\mathbf{C}$ is not available in closed-form. Hence,  the prior's parameter $(\alpha,\theta)$ cannot be estimated following the empirical Bayes approach adopted by \citet{Cai(18)} in the context of the DP prior. Instead, here we estimate $(\alpha,\theta)$ by relying on the minimum Wasserstein distance method \citep{Ber(19)}. This method estimates $(\alpha,\theta)$ by selecting the value of $(\alpha,\theta)$ that minimizes the expected Wasserstein distance between a summary statistic of the data and the corresponding summary statistic of synthetic data generated under the BNP model \eqref{eq:bnp_py}. In our context, a natural choice for the summary statistic is the matrix $\mathbf{C}$. By construction, the rows of $\mathbf{C}$ are i.i.d.; moreover, since $\mathcal{H}$ is assumed to be a perfectly random hash family, each column of $\mathbf{C}$ is exchangeable. Then, we can define the reference summary statistic $\overline{\mathbf{C}}$ as a vector of length  $NJ$ containing the (unordered) entries of the matrix $\mathbf{C}$. For any fixed $m^{\prime}\geq1$ and a any fixed prior's parameter $(\alpha,\theta)$, let $\widetilde{X}_{1:m^{\prime}}=(\widetilde{X}_{1},\ldots,\widetilde{X}_{m^{\prime}})$ be a random sample from $P\sim\text{PYP}(\alpha,\theta;\nu)$, i.e. $\widetilde{X}_{1:m^{\prime}}$ is modeled as \eqref{eq:bnp_py}. For a moderate sample size $m^{\prime}$, generating random variates from $\widetilde{X}_{1:m^{\prime}}$ is straightforward by means of the predictive distribution \eqref{eq:pred_seen_py} of the PYP. These random variates, by a direct transformation through the hash functions $h_{1},\ldots,h_{N}$ drawn at random from $\mathcal{H}$, lead to random variates from the hashed frequencies and to random variates from reference summary statistic, denoted by $\widetilde{\mathbf{C}}(\alpha,\theta,m^{\prime})$.

In practice, $m$ is such that $m\gg m^{\prime}$ and the computational cost of sampling from \eqref{eq:pred_seen_py} scales super-linearly in $m^{\prime}$. To account for this mis-match we scale the entries of $\widetilde{\mathbf{C}}(\alpha,\theta,m^{\prime})$ by $m/m^{\prime}$, so that each row of $\widetilde{\mathbf{C}}(\alpha,\theta,m^{\prime})m/s$ sum to $m$. Now, we are interested in finding $(\hat{\alpha},\hat{\theta})$ such that 
\begin{equation}\label{eq:wass_ana}
(\hat{\alpha},\hat{\theta})=\arg\min_{(\alpha,\theta)}\E\left[\mathcal{W}_1\left(\overline{\mathbf{C}},\widetilde{\mathbf{C}}(\alpha,\theta,m^{\prime})\frac{m}{m^{\prime}}\right)\right],
\end{equation}
where $\mathcal{W}_1$ is the Wasserstein distance of order $1$, and the expectation is taken with respect to $\widetilde{\mathbf{C}}$. To fully specify the optimization problem we choose $\rho(x, y)=|x-y|$ as distance underlying $\mathcal{W}_1$ \citep{Ber(19)}. We make use an MC approximation of the expectation in \eqref{eq:wass_ana}, i.e., 
\begin{equation}\label{eq:wass_emp}
\frac{1}{R}\sum_{r=1}^R\mathcal{W}_p\left(\overline{\mathbf{C}},\widetilde{\mathbf{C}}_r(\alpha,\theta,m^{\prime})\frac{m}{m^{\prime}}\right)
\end{equation}
for $R\geq1$, where $(\widetilde{\mathbf{C}}_{1}(\alpha,\theta,m^{\prime}),\ldots,\widetilde{\mathbf{C}}_{R}(\alpha,\theta,m^{\prime}))$ are i.i.d. according to $\widetilde{\mathbf{C}}(\alpha,\theta,m^{\prime})$. We refer to \citet{Ber(19)} for a theoretical and empirical analysis of the minimum distance Wasserstein method. To improve the MC approximation displayed in \eqref{eq:wass_emp}, which might be detrimental for the minimization problem in \eqref{eq:wass_ana}, we fix the same random numbers underlying the routines used for generating random variates from the predictive distribution \eqref{eq:pred_seen_py} of the PYP over all values of $(\alpha,\theta)$. Moreover the optimization is carried out via noise-robust Gaussian optimization \citep{Let(19)}. We report experimental results in Section \ref{sec5}.

\subsection{Computational aspects of the CMS-PYP}

\begin{figure}[!htb]
\vspace{0.3cm}
\centering
\includegraphics[width=1\linewidth]{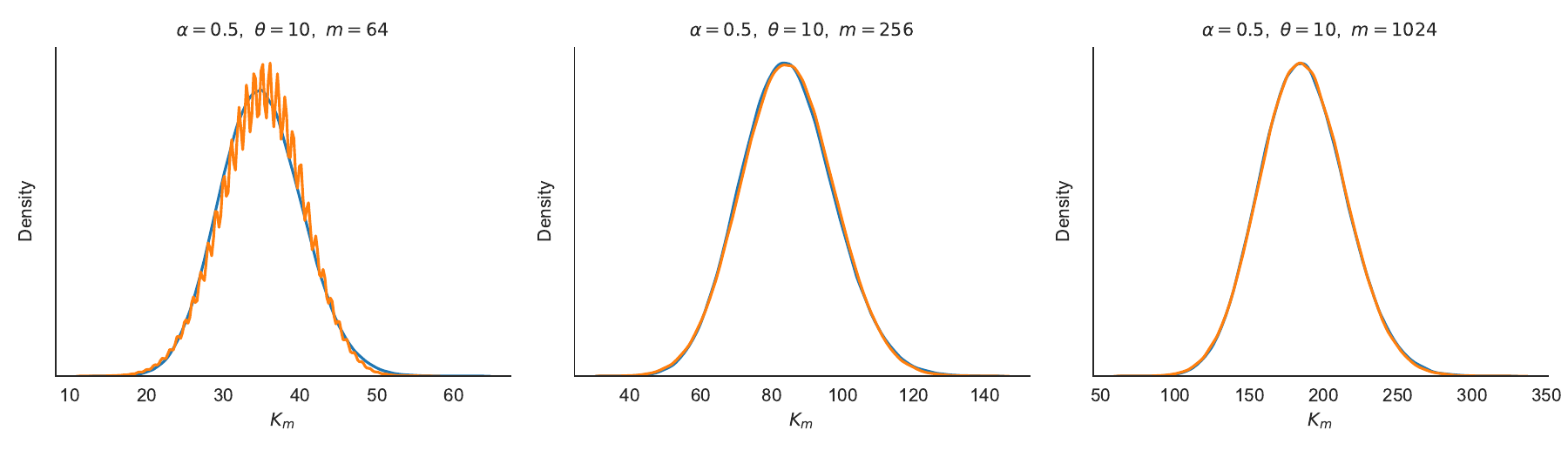}
\caption{\small{Exact sampling of $K_m$ via Algorithm 1 and (asymptotic) approximate sampling of $K_m$ via $K_m \sim [(\theta+m)^{\alpha}-\theta^{\alpha}]S^{-\alpha}_{\alpha,\theta}$ for different values of $m$, $\alpha=0.5$, $\theta=10$; densities estimated by kernel density estimation.}}
\label{fig:k_conv}
\end{figure}

For the CMS-PYP estimator of $f_{x_{m+1}}$ we consider the posterior mean $\hat{f}^{\text{\tiny{(PYP)}}}$, that is the expected value of the posterior distribution \eqref{eq:posterior_py_compl}. The evaluation of $\hat{f}^{\text{\tiny{(PYP)}}}$ follows from two steps:
\begin{itemize}
\item[i)] the estimation of the prior's parameter $(\alpha,\theta)$ by means of the minimum Wasserstein distance method;
\item[ii)] the evaluation, with respect to the estimated prior's parameter, of the posterior distribution \eqref{eq:posterior_py_compl}.
\end{itemize}
Step i) has been described above. Step ii) can be implemented either via the exact representation in \eqref{eq:posterior_py_alternative} or via its limiting behaviour in \eqref{local_limit}, which is accurate provided that the total number of observed tokens $m$ is large relative to the considered $c_n$. This is often the case, especially for real world large datasets where applying CMS in any of its variants is most warranted. In our numerical experiments we consider datasets whose total observed tokens range from 2 millions to almost 1 billion. The evaluation \eqref{eq:posterior_py_alternative} requires the computation of multiple expectations, one for each $l=0,\dots,c_n$, which we approximate via MC integration. For each MC estimator to be valid it is necessary to sample each $K_{c_n-l}$ independently from $K_{m-c_n}$ in each expectation term. However the MC estimators themselves, one for each $l=0,\dots,c_n$, can be correlated. One sample for all MC estimators can be thus obtained as follows: i) Algorithm 1 is used to sample the vector $[K_{c_n-l}\mid l=0,\dots,c_n]$ in one pass with $\mathcal{O}(c_n)$ cost ii) $K_{m-c_n}$ is sampled from the distribution of $[(\theta+(m-c_n))^{\alpha}-\theta^{\alpha}]S^{-\alpha}_{\alpha,\theta}$ where $S^{-\alpha}_{\alpha,\theta}$ is a polynomially tilted $\alpha$-stable random variable. Sampling from $S^{-\alpha}_{\alpha,\theta}$ can be achieved efficiently by using rejection sampling as described in \citet{Dev(09)}. The convergence of $K_{m-c_n}$ to its limiting distribution is fast in $m-c_n$, as illustrated in Figure \ref{fig:k_conv}. To ensure numerical stability with both \eqref{eq:posterior_py_alternative} and \eqref{local_limit} we work in log-space, i.e. compute the (natural) logarithm of each multiplicative term of \eqref{eq:posterior_py_compl}, and exponentiate back only as final computation.  Similarly, to avoid underflow/overflow issues, we apply the "log-sum-exp" trick to sums arising from the MC estimators. The denominator of \eqref{eq:posterior_py_alternative} does not need to be evaluated, as it suffices to compute $p_{f_{X_{m+1}}}(l;c_{n},\alpha,\theta)$ up to a constant of proportionality and then normalize the masses to sum to up to $1$. In doing this, the MC variance is additionally reduced.

\begin{algorithm}
\caption{Sampling $K_{c-l}$ for $l=0,\dots,c$}
\begin{algorithmic}
\State $K[0] \gets 0$;
\State $K[1] \gets 1$;
\State $i\gets 1$;
\While{$i \leq c$}
    \State $Ber\gets\text{random sample from Bernoulli}\left(\frac{\theta+\alpha K[i-1]}{\theta+i}\right)$;
    \State $K[i]\gets K[i-1] + Ber$;
    \State $i\gets i+1$;
\EndWhile
\State \Return reverse of $K$
\end{algorithmic}
\end{algorithm}


\section{Experiments}\label{sec5}

We present numerical experiments for the CMS-PYP. We apply the CMS-PYP to synthetic and real textual data, and we compare its performance with respect to some Bayesian and non-Bayesian approaches. With regards to the Bayesian approaches, we consider the CMS-DP of \citet{Cai(18)}, here denoted by $\hat{f}^{\text{\tiny{(DP)}}}$, and the CMS-NIGP of \citet{Dol(21)}, here denoted by $\hat{f}^{\text{\tiny{(NIGP)}}}$. In particular, $\hat{f}^{\text{\tiny{(DP)}}}$ is the mean of the posterior distribution of $f_{X_{m+1}}$, given $\{C_{n,h_{n}(X_{m+1})}\}_{n\in[N]}$, under the DP prior, whereas $\hat{f}^{\text{\tiny{(NIGP)}}}$ is the mean of the posterior distribution of $f_{X_{m+1}}$, given $\{C_{n,h_{n}(X_{m+1})}\}_{n\in[N]}$, under the NIGP prior. With regards to non-Bayesian approaches, we consider the CMS of \citet{Cor(05)}, the CMM of \citet{Goy(12)} and the BDCM of \citet{Tin(18)}. The CMS, here denoted by $\hat{f}^{\text{\tiny{(CMS)}}}$, has beed specified in \eqref{cms_main}. Both the CMM, here denoted by $\hat{f}^{\text{\tiny{(CMM)}}}$, and the BDCM, here denoted by $\hat{f}^{\text{\tiny{(BDCM)}}}$, rely on the same summary statistics used in the CMS, i.e. the hashed frequencies $\{\mathbf{C}_{n}\}_{n\in[N]}$. This facilitates the implementation of a fair comparison among estimators, since the storage requirement and sketch update complexity are unchanged. The CMM estimator subtracts the value of the estimated noise from each of the $N$ counters, and returns minimum between the median of the $N$ residues and the CMS estimator. For a token $x$ the noise corresponding to each counter is given by $(m - C_{n,h_n(x)})/(J - 1)$. Similarly to the CMM, the BDCM estimator aims at de-biasing the CMS estimator. Specifically, we consider the Algorithm 1 of \citet{Tin(18)}. An (almost) unbiased estimator is obtained by removing the mean of the CMS estimators computed over the vector counters $\{\mathbf{C}_{n,j}\}_{n\in[N]}$ for $j=1,\dots,J$. The BDCM estimator is obtained by restricting this unbiased estimator to be non-negative. In all the experiments random hash functions $h_n(x)$ as 
$$
h_n(x) = ((a_n x + b_n) \bmod LP) \bmod J,
$$
where $x$ is the non-negative integer index corresponding to the token of interest, $LP$ denotes a large prime number (here chosen to be $2^{32}-1$), and $a_n$ and $b_n$ are two random integer numbers that are i.i.d. as a Uniform distribution on $[1,LP]$. The required hash functions are generated once for each numerical experiment and kept fixed while comparing different estimators.

\subsection{Estimation of the prior's parameter $(\alpha,\theta)$}\label{sec52}

We present an empirical study of the likelihood-free estimation approach detailed in Section \ref{sec3}. We start with a scenario where the data generating process (PYP-DGP) is \eqref{eq:bnp_py}. In particular, we generate 10 synthetic datasets of $m=300000$ tokens each, for different prior's parameter $(\alpha,\theta)$. See Table \ref{tab:simu_pyp}. For each dataset, the estimation of the prior's parameter $(\alpha,\theta)$ is performed by means of \eqref{eq:wass_ana} and \eqref{eq:wass_emp} with $m^{\prime}=100000$ for $R=25$. The optimization procedure is based on \citet{Let(19)}, as implemented by the \texttt{AX} library. See \href{https://ax.dev/}{https://ax.dev/} for details. The stochastic objective function \eqref{eq:wass_emp} is evaluated a total of 50 times for each dataset. Results from Table \ref{tab:simu_pyp} support our inferential procedure for $(\alpha,\theta)$. It is also apparent that, for the datasets under consideration, $\alpha$ is more easily identified than $\theta$. We also consider synthetic datasets generated from Zipf's distributions with (exponent) parameter $c>1$, i.e. a Zipf's data generating process with parameter $c$ ($\mathcal{Z}_{c}$-DGP). In particular, we recall that the parameter $c$ controls the tail behaviour of the Zipf's distribution: the smaller $c$ the heavier is the tail of the distribution, i.e., the smaller $c$ the larger the fraction of types with low-frequency tokens. We generate $7$ synthetic datasets of $m=500000$ tokens each, for different parameter $c$. See Table \ref{tab:simu_zip}. For each dataset, the estimation of the prior's parameter $(\alpha,\theta)$ is performed by means of \eqref{eq:wass_ana} and \eqref{eq:wass_emp} with $m^{\prime}=100000$ for $R=25$. The optimization procedure is still based on the work of \citet{Let(19)}. The stochastic objective function \eqref{eq:wass_emp} is evaluated a total of 50 times for each dataset. The results from Table \ref{tab:simu_zip} shows that the PYP prior is able to adapt to different power-law tails behaviours. In particular, we observe that the larger $c$ the smaller $\hat{\alpha}$, which is in agreement with the interpretation of $\alpha$ as the parameter controlling the tail behaviour of the PYP prior.

\begin{table}[!htb]
    \centering
    \begin{tabular}{ccccc}
        \toprule
        \multicolumn{2}{c}{PYP-DGP} & &\multicolumn{2}{c}{Estimates} \\
        \cmidrule(r){1-2} \cmidrule(r){4-5}
        $\alpha$ &  $\theta$ & & $\hat{\alpha}$ &  $\hat{\theta}$\\
         \midrule
        0.00 & 25.00 & &0.02 & 36.31\\
        0.10 & 25.00 & &0.11 & 21.86\\
        0.20 & 25.00 & &0.18 & 16.78\\
        0.30 & 25.00 && 0.26 & 22.83\\
        0.40 & 25.00 & &0.41 & 17.32\\
        0.50 & 25.00 & &0.56 & 10.69\\
        0.60 & 25.00 & &0.56 & 13.42\\
        0.70 & 25.00 && 0.63 & 24.89\\
        0.80 & 25.00 & &0.77 & 10.21\\
        0.90 & 25.00 & &0.88 & 11.26\\
        \bottomrule
    \end{tabular}
    \caption{\small{Prior's parameter $(\alpha,\theta)$ estimates, under PYP-DGP.}}
    \label{tab:simu_pyp}
\end{table}

\begin{table}[!htb]
    \centering
    \begin{tabular}{cccc}
        \toprule
        \multicolumn{1}{c}{$\mathcal{Z}_{c}$-DGP} & &\multicolumn{2}{c}{Estimates} \\
        \cmidrule(r){1-1} \cmidrule(r){3-4}
        $c$  & & $\hat{\alpha}$ &  $\hat{\theta}$\\
        \midrule
        1.05 & &0.92 & 25.37\\
        1.18 & &0.80 & 5.56\\
        1.33 & &0.71 & 1.53\\
        1.54 & &0.67 & 0.61\\
        1.82 & &0.38 & 0.49\\
        2.22 & &0.17 & 0.11\\
        2.86 & &0.01 & 0.23\\
        \bottomrule
    \end{tabular}
    \caption{\small{Prior's parameter $(\alpha,\theta)$ estimates, under $\mathcal{Z}_{c}$-DGP.}}
    \label{tab:simu_zip}
\end{table}

\subsection{Applications to synthetic and real data}\label{sec53}

With regards to synthetic data, we consider datasets generated from Zipf's distributions with exponent $c=1.3,\,1.6,\,1.9,\,2.2$. Each dataset consists of  $m=500000$ tokens. We make use of a $2$-universal hash family, and then assume the following pairs of hashing parameters: i) $J=320$ and $N=2$; ii) $J=160$ and $N=4$. Bayesian and non-Bayesian approaches are compared in terms of the MAE (mean absolute error) between true frequencies and their estimates. Table \ref{tab:experiment_sin} and Table \ref{tab:experiment_sin_2} report the MAE for the Bayesian approaches, with respect to the case $J=320$ and $N=2$ and the case $J=160$ and $N=4$, respectively. From Table \ref{tab:experiment_sin} and Table \ref{tab:experiment_sin_2}, it is clear that $\hat{f}^{\text{\tiny{(PYP)}}}$ has a remarkable better performance than $\hat{f}^{\text{\tiny{(DP)}}}$ in the estimation of low-frequency tokens. In particular, for both Table \ref{tab:experiment_sin} and Table \ref{tab:experiment_sin_2}, if we consider the bin of low-frequencies $(0,256]$ the MAE of $\hat{f}^{\text{\tiny{(PYP)}}}$ is alway smaller than the MAE of $\hat{f}^{\text{\tiny{(DP)}}}$, i.e. $\hat{f}^{\text{\tiny{(PYP)}}}$ outperforms $\hat{f}^{\text{\tiny{(DP)}}}$. This behaviour becomes more and more evident as the parameter $c$ decreases, that is the heavier is the tail of the distribution the more the estimator $\hat{f}^{\text{\tiny{(PYP)}}}$ outperforms the estimator $\hat{f}^{\text{\tiny{(DP)}}}$. For any fixed exponent $c$, the gap between the MAEs of $\hat{f}^{\text{\tiny{(PYP)}}}$  and $\hat{f}^{\text{\tiny{(DP)}}}$ reduces as $v$ increases, and this reduction is much more evident as $c$ becomes large. For any exponent $c$ we expect a frequency threshold, say $v^{\ast}(c)$, such that $\hat{f}^{\text{\tiny{(PYP)}}}$ underestimates $f_{x_{m+1}}$ for $v>v^{\ast}(c)$. From Table \ref{tab:experiment_sin} and Table \ref{tab:experiment_sin_2}, for any two exponents $c_{1}$ and $c_{2}$ such that $c_{1}<c_{2}$ it will be $v^{\ast}(c_{1})>v^{\ast}(c_{2})$. From Table \ref{tab:experiment_sin}, i.e. $J=320$ and $N=2$, it emerges that $\hat{f}^{\text{\tiny{(PYP)}}}$ has a remarkable better performance than $\hat{f}^{\text{\tiny{(NIGP)}}}$ for data with heavier power-law tails ($c=1.3,\,1.6$), whereas from Table \ref{tab:experiment_sin_2}, i.e. $J=160$ and $N=4$,  $\hat{f}^{\text{\tiny{(PYP)}}}$ is competitive with $\hat{f}^{\text{\tiny{(NIGP)}}}$ for data with heavier power-law tails ($c=1.3,\,1.6$). In general, the better performance of $\hat{f}^{\text{\tiny{(PYP)}}}$ with respect to $\hat{f}^{\text{\tiny{(NIGP)}}}$ is not surprising, as $\hat{f}^{\text{\tiny{(NIGP)}}}$ has the critical limitation that it can not be tuned to the power-law degree of the data.

Table \ref{tab:experiment_sin_all1} and Table \ref{tab:experiment_sin_all2} report the MAE for the non-Bayesian approaches for the case $J=320$ and $N=2$ and the case $J=160$ and $N=4$, respectively. From Table \ref{tab:experiment_sin_all1} and Table \ref{tab:experiment_sin_all2}, it is clear that $\hat{f}^{\text{\tiny{(PYP)}}}$ outperforms the $\hat{f}^{\text{\tiny{(CMS)}}}$ in the estimation of low-frequency tokens for both the choices of hashing parameters, i.e. the case $J=320$ and $N=2$ and the case $J=160$ and $N=4$. Moreover, $\hat{f}^{\text{\tiny{(PYP)}}}$ outperforms both $\hat{f}^{\text{\tiny{(CMM)}}}$ and $\hat{f}^{\text{\tiny{(BDCM)}}}$ in the estimation of low-frequency token. The better performance of $\hat{f}^{\text{\tiny{(PYP)}}}$ with respect to $\hat{f}^{\text{\tiny{(CMM)}}}$ is a remarkable results, as the CMM is known to stand out in the estimation of low-frequency tokens \citet[Figure 1]{Goy(12)}. In general, a good performance in the estimation of low-frequency tokens is a desirable feature in natural language or textual data, where it is common the power-law behaviour of the data stream of tokens. In such a data, highest frequency events are often of low interest: frequent words are often grammatical, highly polysemous or without any interesting semantics, while low-frequency words are more relevant.

We conclude by presenting an application of the CMS-PYP to some textual datasets, for which the distribution of words is typically a power-law distribution \citep{Cla(09)}. In particular, we consider 4 textual datasets of increasing corpora size: the 20 Newsgroups dataset\footnote{\scriptsize\url{http://qwone.com/~jason/20Newsgroups/}}, the Enron dataset\footnote{\scriptsize\url{https://archive.ics.uci.edu/ml/machine-learning-databases/bag-of-words/}}, the WikiText-103 dataset\footnote{\scriptsize\url{https://blog.salesforceairesearch.com/the-wikitext-long-term-dependency-language-modeling-dataset/}} and the 1 Billion Word Language Model Benchmark (1BWLMB) dataset\footnote{\scriptsize\url{https://www.statmt.org/lm-benchmark/}}. The 20 Newsgroups dataset consists of $m=2765300$ tokens with $53975$ distinct tokens, whereas the Enron dataset consists of $m=6412175$ tokens with $28102$ distinct tokens. Following the experiments in \citet{Cai(18)}, we make use of a $2$-universal hash family, with the following hashing parameters: i) $J=12000$ and $N=2$; ii) $J=8000$ and $N=4$. By means the goodness of fit test proposed in \citet{Cla(09)}, we found that the 20 Newsgroups and Enron datasets fit with a power-law distribution with exponent $\nu=2.3$ and $\nu=2.1$, respectively. The CMS-PYP estimators $\hat{f}^{\text{\tiny{(PYP)}}}$ for the 20 Newsgroups and Enron datasets are obtained through the implementation of \eqref{eq:posterior_py_alternative}. Table \ref{tab:experiment_1_full} and Table \ref{tab:experiment_2_full} report the MAEs of $\hat{f}^{\text{\tiny{(CMS)}}}$, $\hat{f}^{\text{\tiny{(CMM)}}}$ and $\hat{f}^{\text{\tiny{(BDCM)}}}$, whereas Table \ref{tab:experiment_rea} reports the MAEs of the estimators $\hat{f}^{\text{\tiny{(DP)}}}$ and $\hat{f}^{\text{\tiny{(PYP)}}}$. Results of Table \ref{tab:experiment_1_full} and Table \ref{tab:experiment_2_full} and Table \ref{tab:experiment_rea} confirm the behaviour observed in Zipf' synthetic data. That is, $\hat{f}^{\text{\tiny{(PYP)}}}$ outperforms $\hat{f}^{\text{\tiny{(DP)}}}$ for low-frequency tokens. Moreover, a comparison with respect to $\hat{f}^{\text{\tiny{(CMM)}}}$ reveals that $\hat{f}^{\text{\tiny{(PYP)}}}$ is competitive with $\hat{f}^{\text{\tiny{(CMM)}}}$ in the context of the estimation of low-frequency tokens.

Finally, we consider the WikiText-103 and 1BWLMB datasets. The former consists of $m=82810656$ tokens with $606753$ distinct tokens, whereas the latter consists of $m=658195953$ tokens with $1256524$ distinct tokens. The fit test of \citet{Cla(09)} results in power-law distributions with exponent $\nu=2.15$ and $\nu=1.5$ respectively. Taking into account the increased corpora sizes we consider the following hashing parameters: i) $J=50000$ and $N=2$; ii) $J=35000$ and $N=4$ for WikiText-103; i) $J=140000$ and $N=2$; ii) $J=100000$ and $N=4$ for 1BWLMB. The CMS-PYP estimators $\hat{f}^{\text{\tiny{(PYP)}}}$ are obtained through the implementation of \eqref{local_limit}. Table \ref{tab:experiment_rea_2} reports the MAEs of the estimators $\hat{f}^{\text{\tiny{(DP)}}}$ and $\hat{f}^{\text{\tiny{(PYP)}}}$ applied to the WikiText-103 dataset and to the 1BWLMB dataset. The CMS-PYP estimator offers a competitive performance with respect to both the DP and the CMM estimators. The use of \eqref{local_limit} reduces the computational significantly, in which case the time required to compute the CMS-PYP estimators is similar to the time required for DP estimators.

\begin{table}[!htb]
    \centering
    \resizebox{0.9\textwidth}{!}{\begin{tabular}{lrrrrrrrrrrrr}
        \toprule
        \multicolumn{1}{l}{} & \multicolumn{3}{c}{$\mathcal{Z}_{1.3}$} & \multicolumn{3}{c}{$\mathcal{Z}_{1.6}$} & \multicolumn{3}{c}{$\mathcal{Z}_{1.9}$} & \multicolumn{3}{c}{$\mathcal{Z}_{2.2}$} \\
        \cmidrule(r){2-4} \cmidrule(r){5-7} \cmidrule(r) {8-10} \cmidrule(r){11-13}
        Bins of $x_{m+1}$  & $\hat{f}^{\text{\tiny{(DP)}}}$ & $\hat{f}^{\text{\tiny{(NIGP)}}}$ & $\hat{f}^{\text{\tiny{(PYP)}}}$ & $\hat{f}^{\text{\tiny{(DP)}}}$ & $\hat{f}^{\text{\tiny{(NIGP)}}}$ & $\hat{f}^{\text{\tiny{(PYP)}}}$ & $\hat{f}^{\text{\tiny{(DP)}}}$ & $\hat{f}^{\text{\tiny{(NIGP)}}}$ & $\hat{f}^{\text{\tiny{(PYP)}}}$ & $\hat{f}^{\text{\tiny{(DP)}}}$ & $\hat{f}^{\text{\tiny{(NIGP)}}}$ & $\hat{f}^{\text{\tiny{(PYP)}}}$ \\[0.05cm]
        \midrule
            (0,1] & 1,057.61 & 231.31 & 1.12 &   626.85 &134.75 &  3.36 &   306.70 & 65.71& 115.15  &   51.38 &  12.91& 3.80 \\
            (1,2] & 1,194.67 & 287.43&  2.08 &   512.43 & 119.22 & 2.29 &   153.57 & 37.03 & 31.16  &  288.27 & 61.87 & 93.99  \\
            (2,4] & 1,105.16 & 262.18&  3.63 &   472.59 & 95.78 & 1.85 & 2,406.00 & 353.73& 1,237.41 &  133.31 &26.90 & 17.57  \\
            (4,8] & 1,272.02 & 302.89&  7.40 &   783.88 & 175.10 & 8.89 &   457.57 & 83.30& 136.16  &  117.76 & 21.58 & 8.26  \\
           (8,16] & 1,231.63 &257.08 & 11.83 &   716.52 &136.66 & 10.00 &   377.99 &66.44 &  90.41  &  411.21 & 77.39& 127.69  \\
          (16,32] & 1,252.18 & 248.41& 22.58 &   829.17 & 190.05& 14.81 &   286.98 &41.99 &  65.47  &  501.00 & 90.29&178.07  \\
          (32,64] & 1,309.14 &284.12 & 39.23 &   780.70 &139.52 & 36.47 &   413.95 & 67.30& 181.84  &  216.84 &  48.00 &92.07  \\
         (64,128] & 1,716.76 &312.59 & 104.03 &   946.20 &125.07 & 79.94 & 1,869.23 & 353.10& 1,678.82 &   63.05 & 65.91& 85.70  \\
        (128,256] & 1,102.96 & 97.9& 168.34 & 1,720.49 & 273.50& 342.18 &   199.87 &110.32 &  98.20  &   45.98 & 130.94&136.25  \\
        \bottomrule
    \end{tabular}}
    \caption{\small{Synthetic data: MAE for $\hat{f}^{\text{\tiny{(DP)}}}$, $\hat{f}^{\text{\tiny{(NIGP)}}}$ and $\hat{f}^{\text{\tiny{(PYP)}}}$, case $J=320,N=2$.}}
    \label{tab:experiment_sin}
\end{table}

\begin{table}[!htb]
    \centering
    \resizebox{\textwidth}{!}{
    \begin{tabular}{lrrrrrrrrrrrrr}
        \toprule
        \multicolumn{1}{l}{} & \multicolumn{3}{c}{$\mathcal{Z}_{1.3}$} & \multicolumn{3}{c}{$\mathcal{Z}_{1.6}$} & \multicolumn{3}{c}{$\mathcal{Z}_{1.9}$} & \multicolumn{3}{c}{$\mathcal{Z}_{2.2}$} \\
        \cmidrule(r){2-4} \cmidrule(r){5-7} \cmidrule(r) {8-10} \cmidrule(r){11-13}
        Bins of $x_{m+1}$  & $\hat{f}^{\text{\tiny{(CMS)}}}$ & $\hat{f}^{\text{\tiny{(CMM)}}}$ & $\hat{f}^{\text{\tiny{(BDCM)}}}$ &  $\hat{f}^{\text{\tiny{(CMS)}}}$ & $\hat{f}^{\text{\tiny{(CMM)}}}$ & $\hat{f}^{\text{\tiny{(BDCM)}}}$  & $\hat{f}^{\text{\tiny{(CMS)}}}$ & $\hat{f}^{\text{\tiny{(CMM)}}}$ & $\hat{f}^{\text{\tiny{(BDCM)}}}$  & $\hat{f}^{\text{\tiny{(CMS)}}}$ & $\hat{f}^{\text{\tiny{(CMM)}}}$ & $\hat{f}^{\text{\tiny{(BDCM)}}}$  \\[0.05cm]
        \midrule
            (0,1] &  1,061.3 & 161.72 & 99.02  &      629.40 &    62.19 & 57.08  &      308.11 &    81.10 &  77.37   &     51.65 &   1.04 & 27.80  \\
            (1,2] &  1,197.9 & 169.74 & 35.67  &      514.31 &   102.42 & 45.38  &      154.20 &     2.00 &  55.14   &     289.50 &   2.04 & 21.16  \\
            (2,4] &  1,108.3 & 116.37 & 62.74  &      474.82 &    52.10 & 26.18  &    2,419.51 & 2,215.85 & 1156.54  &  134.05 &   3.40 & 24.63  \\
            (4,8] &  1,275.9 & 378.04 & 145.04 &      786.73 &   214.46 & 40.30  &      460.13 &   258.90 &  68.67   &    118.40 &   6.44 & 62.20  \\
           (8,16] &  1,236.1 & 230.32 & 90.54  &     719.84 &   232.24 & 66.08  &     380.05 &   139.50 &  39.40   &     413.13 & 129.03 & 51.54 \\
          (16,32] &  1,256.8 & 221.98 & 172.14 &    831.70 &    79.73 & 62.98  &    288.59 &    23.90 & 263.34   &    503.60 & 364.30 & 32.60  \\
          (32,64] &  1,312.8 & 235.87 & 197.59 &     783.90 &   184.99 & 73.98  &     415.58 &    54.82 & 116.04   &    217.81 &  82.92 & 28.73  \\
         (64,128] &  1,721.7 & 766.29 & 119.64 &    950.31 &   304.36 & 56.60  &   1,875.50 & 1,762.20 & 1,120.00 &   64.01 &  64.01 & 25.79  \\
        (128,256] &  1,107.7 & 334.57 & 121.17 &  1,727.19 & 1,488.38 & 109.26 &    202.09 &   163.61 & 239.54   &     46.80 &  46.80 & 25.39  \\
        \bottomrule
    \end{tabular}}
       \caption{\small{Synthetic data: MAE for $\hat{f}^{\text{\tiny{(CMS)}}}$, $\hat{f}^{\text{\tiny{(CMM)}}}$ and $\hat{f}^{\text{\tiny{(BDCM)}}}$, case $J=320,N=2$.}}
           \label{tab:experiment_sin_all1}
\end{table}

\begin{table}[!htb]
    \centering
    \resizebox{0.9\textwidth}{!}{\begin{tabular}{lrrrrrrrrrrrr}
        \toprule
        \multicolumn{1}{l}{} & \multicolumn{3}{c}{$\mathcal{Z}_{1.3}$} & \multicolumn{3}{c}{$\mathcal{Z}_{1.6}$} & \multicolumn{3}{c}{$\mathcal{Z}_{1.9}$} & \multicolumn{3}{c}{$\mathcal{Z}_{2.2}$} \\
        \cmidrule(r){2-4} \cmidrule(r){5-7} \cmidrule(r) {8-10} \cmidrule(r){11-13}
        Bins of $x_{m+1}$  & $\hat{f}^{\text{\tiny{(DP)}}}$ & $\hat{f}^{\text{\tiny{(NIGP)}}}$ & $\hat{f}^{\text{\tiny{(PYP)}}}$ & $\hat{f}^{\text{\tiny{(DP)}}}$ & $\hat{f}^{\text{\tiny{(NIGP)}}}$ & $\hat{f}^{\text{\tiny{(PYP)}}}$ & $\hat{f}^{\text{\tiny{(DP)}}}$ & $\hat{f}^{\text{\tiny{(NIGP)}}}$ & $\hat{f}^{\text{\tiny{(PYP)}}}$ & $\hat{f}^{\text{\tiny{(DP)}}}$ & $\hat{f}^{\text{\tiny{(NIGP)}}}$ & $\hat{f}^{\text{\tiny{(PYP)}}}$ \\[0.05cm]
        \midrule
            (0,1] & 2,206.09 &  0.9&  0.77 & 1,254.85 &  0.25 & 1.07 & 420.76 &  0.18 & 0.98 & 153.20 &0.32 &28.78  \\
            (1,2] & 2,333.06 &  0.5& 1.07 & 1,326.71 & 0.70 & 2.13 & 549.12 & 0.82  & 1.93 & 180.71 & 1.24& 21.60  \\
            (2,4] & 2,266.35 &  1.3&  1.70 & 1,267.97 & 2.47 & 3.53 & 482.45 &  2.53 & 3.55 & 182.18 &2.66 &14.92  \\
            (4,8] & 2,229.22 &   4.6& 4.54 & 1,371.27 & 4.67 & 6.11 & 538.91 & 5.28 & 6.28 & 250.32 &5.96 & 40.18  \\
           (8,16] & 2,207.42 & 10.5 & 7.06 & 1,159.29 &10.68 & 11.68 & 487.69 & 10.86& 10.64 & 245.09 &10.28 & 95.33  \\
          (16,32] & 2,279.80 &20.7 & 11.60 & 1,211.41 & 19.21  &23.88 & 529.77 &  22.08&19.04 & 293.68 & 21.57&56.37  \\
          (32,64] & 2,301.99 & 42.6& 28.56 & 1,280.17 & 43.14 &43.61 & 632.45 & 42.64& 40.84 & 118.26 &44.49 &29.04  \\
         (64,128] & 2,241.57 & 92.2 &71.58 & 1,112.41 &94.43  &93.50 & 419.42 & 95.19& 81.83 & 177.61 & 95.10&58.47 \\
        (128,256] & 2,235.40 &  170.0&114.75 & 1,133.85 & 173.87&148.71 & 522.21 &185.83& 226.96 & 128.09 & 180.41&77.92  \\
        \bottomrule
    \end{tabular}}
    \caption{\small{Synthetic data: MAE for $\hat{f}^{\text{\tiny{(DP)}}}$, $\hat{f}^{\text{\tiny{(NIGP)}}}$ and $\hat{f}^{\text{\tiny{(PYP)}}}$, case $J=160,N=4$.}}
    \label{tab:experiment_sin_2}
\end{table}

\begin{table}[!htb]
    \centering
    \resizebox{\textwidth}{!}{
    \begin{tabular}{lrrrrrrrrrrrrr}
        \toprule
        \multicolumn{1}{l}{} & \multicolumn{3}{c}{$\mathcal{Z}_{1.3}$} & \multicolumn{3}{c}{$\mathcal{Z}_{1.6}$} & \multicolumn{3}{c}{$\mathcal{Z}_{1.9}$} & \multicolumn{3}{c}{$\mathcal{Z}_{2.2}$} \\
        \cmidrule(r){2-4} \cmidrule(r){5-7} \cmidrule(r) {8-10} \cmidrule(r){11-13}
        Bins of $x_{m+1}$  & $\hat{f}^{\text{\tiny{(CMS)}}}$ & $\hat{f}^{\text{\tiny{(CMM)}}}$ & $\hat{f}^{\text{\tiny{(BDCM)}}}$  & $\hat{f}^{\text{\tiny{(CMS)}}}$ & $\hat{f}^{\text{\tiny{(CMM)}}}$ & $\hat{f}^{\text{\tiny{(BDCM)}}}$  & $\hat{f}^{\text{\tiny{(CMS)}}}$ & $\hat{f}^{\text{\tiny{(CMM)}}}$ & $\hat{f}^{\text{\tiny{(BDCM)}}}$  & $\hat{f}^{\text{\tiny{(CMS)}}}$ & $\hat{f}^{\text{\tiny{(CMM)}}}$ & $\hat{f}^{\text{\tiny{(BDCM)}}}$  \\[0.05cm]
        \midrule
        (0,1] & 2,212.1 & 590.48 & 126.11  & 1,262.0 & 146.11 & 36.60   & 424.80 & 130.90 & 53.37   & 154.70 &  47.10 & 17.66  \\
        (1,2] & 2,339.8 & 359.57 & 158.41  & 1,332.7 &  63.21 & 72.87   & 552.00 &  65.00 & 189.88  & 182.70 &   2.01 & 18.76  \\
        (2,4] & 2,270.9 &  69.42 & 54.81  & 1,277.8 & 301.89 & 176.24  & 487.30 & 163.55 & 42.74   & 184.70 &  97.15 & 25.86  \\
        (4,8] & 2,234.6 & 339.95 & 92.71   & 1,375.7 & 579.94 & 98.10   & 545.20 & 243.08 & 26.14   & 252.50 &  62.70 & 21.57  \\
       (8,16] & 2,213.3 & 313.37 & 62.11   & 1,165.7 & 152.53 & 59.33   & 493.20 & 196.20 & 102.04  & 247.30 &  29.70 & 22.56  \\
      (16,32] & 2,283.0 &  23.30 & 111.41  & 1,217.2 &  22.94 & 84.80   & 535.50 & 154.30 & 31.80   & 295.90 & 190.92 & 23.36  \\
      (32,64] & 2,305.7 & 133.09 & 172.81  & 1,284.6 & 209.13 & 63.20   & 637.80 & 150.05 & 37.70   & 120.60 &  71.86 & 24.01  \\
     (64,128] & 2,244.5 & 102.43 & 57.11   & 1,120.2 & 118.42 & 73.93   & 425.10 & 198.60 & 36.43  & 180.30 & 113.75 & 22.57  \\
    (128,256] & 2,237.4 & 294.43 & 118.11  & 1,141.3 & 573.12 & 48.07  & 525.90 & 267.15 & 41.26   & 129.70 & 129.70 & 22.73  \\
        \bottomrule
    \end{tabular}}
       \caption{\small{Synthetic data: MAE for $\hat{f}^{\text{\tiny{(CMS)}}}$, $\hat{f}^{\text{\tiny{(CMM)}}}$ and $\hat{f}^{\text{\tiny{(BDCM)}}}$, case $J=160,N=4$.}}
           \label{tab:experiment_sin_all2}
\end{table}

\begin{table}[!htb]
    \centering
    \resizebox{0.8 \textwidth}{!}{\begin{tabular}{lrrrrrrrr}
        \toprule
        \multicolumn{1}{l}{} & \multicolumn{4}{c}{$J=12000$ and $N=2$} & \multicolumn{4}{c}{$J=8000$ and $N=4$}\\
        \cmidrule(r){2-5} \cmidrule(r){6-9}
        \multicolumn{1}{l}{} & \multicolumn{2}{c}{20 Newsgroups} & \multicolumn{2}{c}{Enron}& \multicolumn{2}{c}{20 Newsgroups} & \multicolumn{2}{c}{Enron} \\
        \cmidrule(r){2-3} \cmidrule(r){4-5}\cmidrule(r){6-7}\cmidrule(r){8-9}
         Bins of $x_{m+1}$  & $\hat{f}^{\text{\tiny{(DP)}}}$ & $\hat{f}^{\text{\tiny{(PYP)}}}$ & $\hat{f}^{\text{\tiny{(DP)}}}$ & $\hat{f}^{\text{\tiny{(PYP)}}}$ & $\hat{f}^{\text{\tiny{(DP)}}}$ & $\hat{f}^{\text{\tiny{(PYP)}}}$ & $\hat{f}^{\text{\tiny{(DP)}}}$ & $\hat{f}^{\text{\tiny{(PYP)}}}$\\[0.05cm]
        \midrule
                 (0,1] &  46.39 &   1.22&  12.20 &   0.90 & 53.39 &   0.99&  70.98 &   1.18\\
                 (1,2] &  16.60 &   1.85&  13.80 &   1.86 & 30.49 &   2.10&  47.38 &   2.05\\
                 (2,4] &  38.40 &   3.24&  61.49 &   3.60 & 32.49 &   3.66&  52.49 &   4.14\\
                 (4,8] &  59.39 &   5.04&  88.39 &   7.68 & 38.69 &   6.59&  53.08 &   6.13 \\
                (8,16] &  54.29 &  10.90&  23.40 &  12.85 & 25.29 &  13.17&  56.98 &  11.55\\
               (16,32] &  17.80 &  20.89&  55.09 &  23.97 & 24.99 &  22.69&  89.98 &  19.29\\
               (32,64] &  40.79 &  43.93& 128.48 &  48.94 & 39.69 &  46.42& 108.37 &  47.61\\
              (64,128] &  25.99 &  77.72& 131.08 &  78.51 & 22.09 &  91.15&  55.67 &  70.81\\
             (128,256] &  13.59 & 170.82&  50.68 & 165.28  &25.79 & 191.35&  80.76 & 172.07\\
        \bottomrule
    \end{tabular}}
    \caption{\small{20 Newsgroups and Enron real data: MAE for $\hat{f}^{\text{\tiny{(PYP)}}}$ and $\hat{f}^{\text{\tiny{(DP)}}}$.}}
    \label{tab:experiment_rea}
\end{table}

\begin{table}[!htb]
    \centering
    \resizebox{0.8\textwidth}{!}{    \begin{tabular}{lrrrrrr}
        \toprule
        \multicolumn{1}{l}{} & \multicolumn{3}{c}{20 Newsgroups} & \multicolumn{3}{c}{Enron} \\
        \cmidrule(r){2-4} \cmidrule(r){5-7}
        Bins of $x_{m+1}$  & $\hat{f}^{\text{\tiny{(CMS)}}}$ &  $\hat{f}^{\text{\tiny{(CMM)}}}$ & $\hat{f}^{\text{\tiny{(BDCM)}}}$ & $\hat{f}^{\text{\tiny{(CMS)}}}$ & $\hat{f}^{\text{\tiny{(CMM)}}}$ &$\hat{f}^{\text{\tiny{(BDCM)}}}$ \\
        \midrule
                 (0,1] &  46.4 &   5.41 & 18.04 &     12.2 &   0.90 & 29.20   \\
                 (1,2] &  16.6 &  2.16  & 56.00 &     13.8 &   2.00 & 28.30   \\
                 (2,4] &  38.4 &  7.91  & 23.14 &     61.5 &   9.90 & 26.00  \\
                 (4,8] &  59.4 &  35.70 & 31.42 &     88.4 &  17.32 & 24.30   \\
                (8,16] &  54.3 &  45.40 & 22.72 &    23.4 &   9.52 & 118.70  \\
               (16,32] &  17.8 &  20.99 & 27.72 &    55.1 &  21.00 & 21.60   \\
               (32,64] &  40.8 &  58.86 & 48.24 &   128.5 & 134.47 & 24.40   \\
              (64,128] &  26.0 &  91.59 & 23.20 &   131.1 & 110.27 & 27.60   \\
             (128,256] &  13.6 & 186.92 & 28.05 &   50.7 & 140.43 & 110.40  \\
        \bottomrule
    \end{tabular}}
      \caption{\small{Real data: MAE for $\hat{f}^{\text{\tiny{(CMS)}}}$, $\hat{f}^{\text{\tiny{(CMM)}}}$ and $\hat{f}^{\text{\tiny{(BDCM)}}}$, case $J=12000$ and $N=2$.}}
  \label{tab:experiment_1_full}
\end{table}

\begin{table}[!htb]
    \centering
    \resizebox{0.8\textwidth}{!}{    \begin{tabular}{lrrrrrr}
        \toprule
        \multicolumn{1}{l}{} & \multicolumn{3}{c}{20 Newsgroups} & \multicolumn{3}{c}{Enron} \\
        \cmidrule(r){2-4} \cmidrule(r){5-7}
        Bins of $x_{m+1}$  & $\hat{f}^{\text{\tiny{(CMS)}}}$ &  $\hat{f}^{\text{\tiny{(CMM)}}}$ & $\hat{f}^{\text{\tiny{(BDCM)}}}$&  $\hat{f}^{\text{\tiny{(CMS)}}}$ & $\hat{f}^{\text{\tiny{(CMM)}}}$ &$\hat{f}^{\text{\tiny{(BDCM)}}}$ \\
        \midrule
                 (0,1] & 53.4 &   4.50 & 25.90  &  71.0 &  51.00 & 31.56  \\
                 (1,2] & 30.5 &   2.00 & 20.30  &  47.4 &  27.20 & 21.26  \\
                 (2,4] & 32.5 &   4.80 & 19.00  &  52.5 &   3.90 & 22.42  \\
                 (4,8] & 38.7 &   6.23 & 22.40  &  53.1 &  10.50 & 34.88  \\
                (8,16] & 25.3 &  13.50 & 24.50  &  57.0 &  22.20 & 14.74  \\
               (16,32] & 25.0 &  21.60 & 22.20  &  90.0 &  20.60 & 27.94  \\
               (32,64] & 39.7 &  39.22 & 24.00  & 108.4 &  61.38 & 49.38  \\
              (64,128] & 22.1 &  86.32 & 19.50  &  55.7 &  66.50 & 21.12  \\
             (128,256] & 25.8 & 183.96 & 26.30  &  80.8 &  90.20 & 34.10  \\
        \bottomrule
    \end{tabular}}
     \caption{\small{Real data: MAE for $\hat{f}^{\text{\tiny{(CMS)}}}$, $\hat{f}^{\text{\tiny{(CMM)}}}$ and $\hat{f}^{\text{\tiny{(BDCM)}}}$, case $J=8000$ and $N=4$.}}
   \label{tab:experiment_2_full}
\end{table}

\begin{table}[!htb]
    \centering
    \resizebox{0.8 \textwidth}{!}{\begin{tabular}{lrrrrrrrr}
        \toprule
        \multicolumn{1}{l}{} & \multicolumn{4}{c}{WikiText-103} & \multicolumn{4}{c}{1BWLMB}\\
        \cmidrule(r){2-5} \cmidrule(r){6-9}
        \multicolumn{1}{l}{} & \multicolumn{2}{c}{$J=50000$ and $N=2$} & \multicolumn{2}{c}{$J=35000$ and $N=4$}& \multicolumn{2}{c}{$J=140000$ and $N=2$} & \multicolumn{2}{c}{$J=100000$ and $N=4$} \\
        \cmidrule(r){2-3} \cmidrule(r){4-5}\cmidrule(r){6-7}\cmidrule(r){8-9}
        Bins of $x_{m+1}$  & $\hat{f}^{\text{\tiny{(DP)}}}$ & $\hat{f}^{\text{\tiny{(PYP)}}}$ & $\hat{f}^{\text{\tiny{(DP)}}}$ & $\hat{f}^{\text{\tiny{(PYP)}}}$ & $\hat{f}^{\text{\tiny{(DP)}}}$ & $\hat{f}^{\text{\tiny{(PYP)}}}$ & $\hat{f}^{\text{\tiny{(DP)}}}$ & $\hat{f}^{\text{\tiny{(PYP)}}}$\\[0.05cm]
        \midrule
                 (0,1] & 97.30  &  43.15 & 119.59 & 40.40 & 702.70 & 41.16 & 156.50 & 34.99 \\
                 (1,2] & 61.30  &  34.07 & 145.39 & 31.30 & 104.10 & 35.18 & 138.10 & 35.89 \\
                 (2,4] & 157.70 &  34.29 & 91.79  & 31.50 & 50.10  & 37.55 & 65.00  & 35.50 \\
                 (4,8] & 192.59 &  35.45 & 120.49 & 34.00 & 552.59 & 33.92 & 49.40  & 37.50 \\
                (8,16] & 191.59 &  41.38 & 111.09 & 32.20 & 176.40 & 34.75 & 43.70  & 32.70 \\
               (16,32] & 195.19 &  33.10 & 127.09 & 46.50 & 143.40 & 38.09 & 97.40  & 35.09 \\
               (32,64] & 248.29 &  34.29 & 102.09 & 44.30 & 600.30 & 37.31 & 168.90 & 40.39 \\
              (64,128] & 632.19 &  37.71 & 208.29 & 42.40 & 143.40 & 44.31 & 89.90  & 41.50 \\
             (128,256] & 107.69 &  42.09 & 140.29 & 59.90 & 485.29 & 48.99 & 58.60  & 55.49 \\
        \bottomrule
    \end{tabular}}
    \caption{\small{WikiText-103 and 1BWLMB real data: MAE for $\hat{f}^{\text{\tiny{(PYP)}}}$ and $\hat{f}^{\text{\tiny{(DP)}}}$.}}
    \label{tab:experiment_rea_2}
\end{table}


\section{Discussion}\label{sec6}

In this paper, we contributed to the study of the CMS-DP of \citet{Cai(18)}. At the core of the CMS-DP lies the computation of the posterior distribution of a point query, given the hashed data, and then estimates of the point query are obtained as mean functionals of such a posterior distribution. While the CMS-DP has proved to improve on some aspects of CMS, it has the major drawback that the posterior distribution of a point query is obtained through a``constructive" proof that builds upon arguments tailored to the DP prior, namely arguments that are not usable for other nonparametric priors. In this paper, we presented a ``Bayesian" proof of the CMS-DP, that is we computed the (regular) conditional distribution of a point query, given the hashed data, and we showed that such a distribution coincides with the posterior distribution obtained in \citet{Cai(18)}. Besides strengthening the BNP approach of \citet{Cai(18)} through rigorous arguments, our proof improve its flexibility by avoiding the use of properties that are peculiar to the DP, thus paving the way to go beyond the use of the DP prior. This first result led to develop a novel learning-augmented CMS under power-law data streams, referred to as CMS-PYP, which relies on BNP modeling of the data stream of tokens via a PYP prior. Under this more general framework, we applied the arguments of the “Bayesian” proof of the CMS-DP, suitably adapted to the PYP prior, to compute the posterior distribution of a point query, given the hashed data. Both the CMS-DP and the CMS-PYP have been also investigated with respect to large sample asymptotic behaviours of their corresponding posterior distributions. Applications to synthetic data and real textual data revealed that the CMS-PYP outperforms the CMS and the CMS-DP in estimating low-frequency tokens, and it is competitive with respect to the CMM and the BDCM.

Our ``Bayesian" proof of the CMS-DP can be extended to deal with more general queries. Of notable interest is the problem of estimating the overall frequency of $s\geq1$ tokens in the stream, also referred to as $s$-range query, which generalizes the point query \citep[Chapter 3]{Cor(20)}. For $m\geq1$ let $x_{1:m}$ be a stream of $\mathcal{V}$-valued tokens, and for positive integers $J$ and $N$ let $h_{1},\ldots,h_{N}$, with $h_{n}:\mathcal{V}\rightarrow [J]$, be random hash functions that are i.i.d. from a pairwise independent hash family $\mathcal{H}$. Then, assuming $x_{1:m}$ to be available through the hashed data $\{(C_{n,1},\ldots,C_{n,J})\}_{n\in[N]}$, the goal is to estimate, or recover, the vector of frequencies $(f_{x_{m+1}},\ldots,f_{x_{m+s}})$ of $s$ new tokens $(x_{m+1},\ldots,x_{m+s})$ in $x_{1:m}$, with $f_{x_{m+r}}$ being defined as
\begin{displaymath}
f_{x_{m+r}}=\sum_{i=1}^{m}\mathbbm{1}_{\{x_{i}\}}(x_{m+r})
\end{displaymath}
for $r=1,\ldots, s$, and hence the $s$-range query $\bar{f}_{s}=\sum_{1\leq r\leq s}f_{x_{m+r}}$. The arguments of the ``constructive" proof of \citet{Cai(18)} exploit the unidimensional nature of point queries, and therefore they cannot be used for the vector $(f_{x_{m+1}},\ldots,f_{x_{m+s}})$ nor for $\bar{f}_{s}$. In Appendix \ref{secrange} we show how to adapt our ``Bayesian" proof to the problem of computing the posterior distribution of $(f_{x_{m+1}},\ldots,f_{x_{m+s}})$, given hashed data, and, as an illustrative example, we present the posterior distribution of $(f_{x_{m+1}},f_{x_{m+2}})$. We focus on the DP prior, thought the same arguments apply to the PYP prior. Unfortunately, the posterior distribution of $(f_{x_{m+1}},f_{x_{m+2}})$ has a rather complicated form, and for a large $m$ the computational burden for its evaluation becomes overwhelming. We defer to future work the study of a large sample behaviour of the posterior distribution, with the aim of obtaining a simple approximated version of it.

Our work paves the way to some fruitful directions for future research in the context of the BNP approach to obtain learning-augmented CMSs. Investigating large sample asymptotic properties of the CMS-DP and CMS-PYP would be of interest, especially with the aim of obtaining simple approximated versions of the posterior distributions \eqref{post_full_dp} and \eqref{eq:posterior_py_compl}. For a single hash function, i.e. $N=1$, Proposition \ref{more_cmsdp} and Proposition \ref{asymp_py}, as well as Equation \eqref{local_limit}, provide results in this direction. However, it would be of greater interest to consider consider corresponding results for an arbitrary $N$, that is for the posterior distributions \eqref{post_full_dp} and \eqref{eq:posterior_py_compl}. Our conjecture is that, under suitable assumptions, the large $m$ limiting posterior distribution of a rescaled point query reduces to a distribution that involves only the minimum of the hashed frequencies, i.e. $\min\{c_{1},\ldots,c_{N}\}$, thus making a link with the CMS. In this respect, it would be interesting to obtain some form of central limit theorem for the posterior distributions \eqref{post_full_dp} and \eqref{eq:posterior_py_compl}. For $\alpha=0$, \citet{Cai(18)} showed that the posterior mode may recover the CMS estimate of \citet{Cor(05)}, while other CMS-DP estimates may be viewed as CMS estimates with shrinkage; it is natural to ask whether there exists a similar interplay between the CMS-PYP and variations of the CMS for power-law data streams, e.g. the CMM.  Other directions of interest consist in using the CMS-DP and CMS-PYP for large-scale streaming algorithms, e.g., for large text or streaming graphs applications \citep{Cor(12)}, as well as to accommodate nonlinear update operations, such as the conservative update \citep[Chapter 3]{Cor(20)}.


\appendix

\appendixtitleon

\begin{appendices}

\section{Proof of Proposition \ref{more_cmsdp}}\label{proof_binomial}

The proof of Equation \eqref{distr_ident} is straightforward, and it follows from Equation \eqref{eq_direct} by means of the definition of Beta-Binomial distribution \citep[Chapter 6]{Joh(05)}. With regards to the proof of Equation \eqref{distr_asym}, for $t\in\mathbb{R}^{+}$ and $u\in\mathbb{N}_{0}$, let $(t)_{[u]}:=\prod_{0\leq i\leq u-1}(t-i)$ denote the falling factorial of $t$ of order $u$, with the proviso that
$(t)_{[0]}:=1$. In particular, $(t)_{(u)}=(-1)^{u}(-t)_{[u]}$. Recall that the $(u,v)$-th Stirling number of the second type, here denoted by $S(u,v)$, is defined as the $v$-th coefficient in the expansion of $t^{u}$ into falling factorials, i.e. $t^{u}=\sum_{0\leq v\leq u}S(u,v)(t)_{[v]}$; moreover, it is assumed: $S(0,0)=1$, $S(u,0)=0$ for $u>0$ and $S(u,v)=0$ for $v>u$. Then, for $r\geq1$
\begin{align*}
\E\left[\left(\frac{F_{X_{m+1}}}{c_{n}}\right)^{r}\right]&=c_{n}^{-r}\sum_{l=0}^{c_{n}}l^{r}\frac{\frac{\theta}{J}}{\frac{\theta}{J}+c_{n}}\frac{\left(c_{n}-l+1\right)_{(l)}}{\left(\frac{\theta}{J}+c_{n}-l\right)_{(l)}}\\
&=c_{n}^{-r}\sum_{l=0}^{c_{n}}\left(\sum_{k=0}^{r}S(r,k)(l)_{[k]}\right)\frac{\frac{\theta}{J}}{\frac{\theta}{J}+c_{n}}\frac{\left(c_{n}-l+1\right)_{(l)}}{\left(\frac{\theta}{J}+c_{n}-l\right)_{(l)}}\\
&=c_{n}^{-r}\sum_{k=0}^{r}S(r,k)\sum_{l=0}^{c_{n}}\frac{\frac{\theta}{J}}{\frac{\theta}{J}+c_{n}}\frac{\left(c_{n}-l+1\right)_{(l)}}{\left(\frac{\theta}{J}+c_{n}-l\right)_{(l)}}(l)_{[k]}\\
&=c_{n}^{-r}\sum_{k=0}^{r}S(r,k)\frac{\frac{\theta}{J}}{\frac{\theta}{J}+c_{n}}\frac{\Gamma\left(c_{n}+1\right)}{\Gamma\left(\frac{\theta}{J}+c_{n}\right)}k!\sum_{l=0}^{c_{n}}{c_{n}-l\choose k}\frac{\Gamma\left(\frac{\theta}{J}+l\right)}{\Gamma\left(1+l\right)}\\
&=c_{n}^{-r}\sum_{k=0}^{r}S(r,k)\frac{\frac{\theta}{J}}{\frac{\theta}{J}+c_{n}}\frac{\Gamma\left(c_{n}+1\right)}{\Gamma\left(\frac{\theta}{J}+c_{n}\right)}k!\frac{\Gamma\left(1+\frac{\theta}{J}+c_{n}\right)}{\Gamma\left(c_{n}+1-k\right)\left(\frac{\theta}{J}\right)_{(k+1)}}\\
&=c_{n}^{-r}\frac{\theta}{J}\sum_{k=0}^{r}S(r,k)\frac{\Gamma(k+1)\Gamma\left(c_{n}+1\right)}{\Gamma\left(c_{n}+1-k\right)\left(\frac{\theta}{J}\right)_{(k+1)}}.
\end{align*}
By  a direct application of Stirling formula for the ratio of Gamma functions, as $c_{n}\rightarrow+\infty$ it holds
\begin{align*}
\E\left[\left(\frac{F_{X_{m+1}}}{c_{n}}\right)^{r}\right]&\approx c_n^{-r}\frac{\theta}{J}\sum_{k=0}^{r}S(r,k)\frac{\Gamma(k+1)}{\left(\frac{\theta}{J}\right)_{(k+1)}}c_n^{k}\\
&\approx \frac{\theta}{J}\frac{\Gamma(r+1)}{\left(\frac{\theta}{J}\right)_{(r+1)}}\\
&=\frac{\Gamma\left(r+1\right)\Gamma\left(\frac{\theta}{J}+1\right)}{\Gamma\left(\frac{\theta}{J}+r+1\right)\Gamma\left(1\right)}\\
&=\E[B_{1,\frac{\theta}{J}}].
\end{align*}
for any $r\geq1$. This completes the proof of \eqref{distr_asym}, and hence the proof of Proposition \ref{more_cmsdp} is completed.

\section{Proof of Theorem \ref{teo_pyp}}\label{proof_pyp}

The proof is along lines similar to the ``Bayesian" proof of Section \ref{sec2}. To simplify the notation, we remove the subscript $n$ from $h_{n}$ and $c_{n}$. Then, we are interest in computing the posterior distribution
\begin{align}\label{eq:maintoprovepy}
&\text{Pr}[f_{X_{m+1}}=l\,|\,C_{h(X_{m+1})}=c]\\
&\notag\quad=\text{Pr}\left[f_{X_{m+1}}=l\,|\,\sum_{i=1}^{m}\mathbbm{1}_{\{h(X_{i})\}}(h(X_{m+1}))=c\right]\\
&\notag\quad=\frac{\text{Pr}\left[f_{X_{m+1}}=l,\sum_{i=1}^{m}\mathbbm{1}_{\{h(X_{i})\}}(h(X_{m+1}))=c\right]}{\text{Pr}\left[\sum_{i=1}^{m}\mathbbm{1}_{\{h(X_{i})\}}(h(X_{m+1}))=c\right]}
\end{align}
for $l=0,1,\ldots,m$. The independence between $h_{n}$ and $X_{1:m}$ allows us to invoke the ``freezing lemma'' \citep[Lemma 4.1]{Bal(17)}, according to which we can treat $h_{n}$ as it was fixed, i.e. non-random. We start with the denominator of \eqref{eq:maintoprovepy}. Uniformity of the hash function $h$ implies that $h$ induces a (fixed) $J$-partition $\{B_{1},\ldots,B_{J}\}$ of $\mathcal{V}$ such that $B_{j}=\{v\in\mathcal{V}\text{ : }h(v)=j\}$ and $\nu(B_{j})=J^{-1}$ for $j=1,\ldots,J$. Accordingly, we can write the denominator of \eqref{eq:maintoprovepy} as
\begin{align}\label{eq:denom1py}
&\text{Pr}\left[\sum_{i=1}^{m}\mathbbm{1}_{\{h(X_{i})\}}(h(X_{m+1}))=c\right]\\
&\notag\quad=J{m\choose c}\E[(P(B_{j}))^{c+1}(1-P(B_{j}))^{m-c}]\\
&\notag\quad=J{m\choose c}\E[(P(B_{j}))^{c+1}P(\bar{B}_{j})^{m-c}]\\
&\notag\quad=J{m\choose c}\sum_{i=0}^{c+1}\sum_{j=0}^{m-c}\frac{\left(\frac{\theta}{\alpha}\right)_{(i+j)}}{(\theta)_{(m+1)}}\left(\frac{1}{J}\right)^{i}\left(1-\frac{1}{J}\right)^{j}\mathscr{C}(c+1,i;\alpha)\mathscr{C}(m-c,j;\alpha),
\end{align}
where the last equality follows from \citet[Equation 3.3]{San(06)}. This completes the study of the denominator of \eqref{eq:maintoprovepy}. Now, we consider the numerator of \eqref{eq:maintoprovepy}. Let us define the event $B(m,l)=\{X_{1}=\cdots=X_{l}=X_{m+1},\{X_{l+1},\ldots,X_{m}\}\cap\{X_{m+1}\}=\emptyset\}$. In particular, we write
\begin{align}\label{eq:num1_1py}
&\text{Pr}\left[f_{X_{m+1}}=l,\sum_{i=1}^{m}\mathbbm{1}_{\{h(X_{i})\}}(h(X_{m+1}))=c\right]\\
&\notag={m\choose l}\text{Pr}\Bigg[B(m,l),\,\sum_{i=1}^{m}\mathbbm{1}_{\{h(X_{i})\}}(h(X_{m+1}))=c\Bigg]\\
&\notag={m\choose l}\text{Pr}\Bigg[B(m,l),\,\sum_{i=l+1}^{m}\mathbbm{1}_{\{h(X_{i})\}}(h(X_{m+1}))=c-l\Bigg].
\end{align}
That is, the distribution of $(f_{X_{m+1}},C_{j})$ is completely determined by the knowledge of the distribution of $(X_{1},\ldots,X_{m+1})$. Let $\Pi(s,k)$ denote the set of all possible partitions of the set $\{1,\ldots,s\}$ into $k$ disjoints subsets $\pi_{1},\ldots,\pi_{k}$ such that $n_{i}$ is the cardinality of $\pi_{i}$. In particular, from \citet[Equation 3.5]{San(06)}, for any measurable $A_{1},\ldots,A_{m+1}$ we have that
\begin{align*}
\text{Pr}[X_{1}\in A_{1},\ldots,X_{m+1}\in A_{m+1}]&=\sum_{k=1}^{m+1}\frac{\prod_{i=0}^{k-1}(\theta+i\alpha)}{(\theta)_{(m+1)}}\\
&\quad\times\sum_{(\pi_{1},\ldots,\pi_{k})\in\Pi(n+1,k)}\prod_{i=1}^{k}(1-\alpha)_{(n_{i}-1)}\nu(\cap_{m\in\pi_{i}}A_{m})
\end{align*}
for $m\geq1$. Let $\mathscr{V}$ be the Borel $\sigma$-algebra of $\mathcal{V}$. Let $\nu_{\pi_{1},\ldots,\pi_{k}}$ be a probability measure on $(\mathcal{V}^{m+1},\mathscr{V}^{m+1})$ defined as
\begin{displaymath}
\nu_{\pi_{1},\ldots,\pi_{k}}(A_{1}\times\cdots\times A_{m+1})=\prod_{1\leq i\leq k}\nu(\cap_{m\in\pi_{i}}A_{m}),
\end{displaymath}
and attaching to $B(m,l)$ a value that is either $0$ or $1$. In particular, $\nu_{\pi_{1},\ldots,\pi_{k}}(B(m,l))=1$ if and only if one of the $\pi_{i}$'s is equal to the set $\{1,\ldots,l,m+1\}$. Hence, based on the measure $\nu_{\pi_{1},\ldots,\pi_{k}}$, we write
\begin{align*}
&\text{Pr}\left[B(m,l),\sum_{i=l+1}^{m}\mathbbm{1}_{\{h(X_{i})\}}(h(X_{m+1}))=c-l\right]\\
&\quad=\sum_{k=2}^{m-l+1}\frac{\prod_{i=0}^{k-1}(\theta+i\alpha)}{(\theta)_{(m+1)}}\\
&\quad\quad\times\sum_{(\pi_{1},\ldots,\pi_{k-1})\in\Pi(m-l,k-1)}(1-\alpha)_{(l)}\prod_{i=1}^{k-1}(1-\alpha)_{(n_{i}-1)}\nu_{\pi_{1},\ldots,\pi_{k}}\left(\sum_{i=l+1}^{m}\mathbbm{1}_{\{h(X_{i})\}}(h(X_{m+1}))=c-l\right)\\
&\quad=\theta\frac{(\theta+\alpha)_{(m-l)}}{(\theta)_{(m+1)}}(1-\alpha)_{(l)}\\
&\quad\quad\times\sum_{r=1}^{m-l}\frac{\prod_{i=0}^{r-1}(\theta+\alpha+i\alpha)}{(\theta+\alpha)_{(m-l)}}\sum_{(\pi_{1},\ldots,\pi_{r})\in\Pi(m-l,r)}\prod_{i=1}^{r}(1-\alpha)_{(n_{i}-1)}\nu_{\pi_{1},\ldots,\pi_{r}}\left(\sum_{i=1}^{m-l}\mathbbm{1}_{\{h(X_{i})\}}(h(X_{m+1}))=c-l\right).
\end{align*}
Now, 
\begin{displaymath}
\sum_{r=1}^{m-l}\frac{\prod_{i=0}^{r-1}(\theta+\alpha+i\alpha)}{(\theta+\alpha)_{(m-l)}}\sum_{(\pi_{1},\ldots,\pi_{r})\in\Pi(m-l,r)}\prod_{i=1}^{r}(1-\alpha)_{(n_{i}-1)}\nu_{\pi_{1},\ldots,\pi_{r}}\left(\cdot\right)
\end{displaymath}
is the distribution of a random sample $(X_{1},\ldots,X_{m-l})$ from $P\sim\text{PYP}(\alpha,\theta+\alpha;\nu)$. Again, the distribution of $(X_{1},\ldots,X_{m-l})$ is given in \citet[Equation 3.5]{San(06)}. In particular, we write
\begin{align*}
&\text{Pr}\left[B(m,l),\sum_{i=l+1}^{m}\mathbbm{1}_{\{h(X_{i})\}}(h(X_{m+1}))=c-l\right]\\
&\quad=\theta\frac{(\theta+\alpha)_{(m-l)}}{(\theta)_{(m+1)}}(1-\alpha)_{(l)}\\
&\quad\quad\times\sum_{r=1}^{m-l}\frac{\prod_{i=0}^{r-1}(\theta+\alpha+i\alpha)}{(\theta+\alpha)_{(m-l)}}\sum_{(\pi_{1},\ldots,\pi_{r})\in\Pi(m-l,r)}\prod_{i=1}^{r}(1-\alpha)_{(n_{i}-1)}\nu_{\pi_{1},\ldots,\pi_{r}}\left(\sum_{i=1}^{m-l}\mathbbm{1}_{\{h(X_{i})\}}(h(X_{m+1}))=c-l\right)\\
&\quad=\theta\frac{(\theta+\alpha)_{(m-l)}}{(\theta)_{(m+1)}}(1-\alpha)_{(l)}{m-l\choose c-l}\E[(P(B_{j}))^{c-l}(1-P(B_{j}))^{m-c}]\\
&\quad=\theta\frac{(\theta+\alpha)_{(m-l)}}{(\theta)_{(m+1)}}(1-\alpha)_{(l)}{m-l\choose c-l}\E[(P(B_{j}))^{c-l}P(\bar{B}_{j})^{m-c}]\\
&\quad=\theta\frac{(\theta+\alpha)_{(m-l)}}{(\theta)_{(m+1)}}(1-\alpha)_{(l)}{m-l\choose c-l}\sum_{i=0}^{c-l}\sum_{j=0}^{m-c}\frac{\left(\frac{\theta+\alpha}{\alpha}\right)_{(i+j)}}{(\theta+\alpha)_{(m-l)}}\left(\frac{1}{J}\right)^{i}\left(1-\frac{1}{J}\right)^{j}\mathscr{C}(c-l,i;\alpha)\mathscr{C}(m-c,j;\alpha),
\end{align*}
where the second identity and the last identity follow from an application of \citet[Proposition 3.1]{San(06)} and \citet[Equation 3.3]{San(06)}, respectively, under the PYP prior; see also the formule displayed at page 469 of \citet{San(06)}). Accordingly, from  \eqref{eq:num1_1py} we can write that
\begin{align}\label{eq:num1_2py}
&\text{Pr}\left[f_{X_{m+1}}=l,\sum_{i=1}^{m}\mathbbm{1}_{\{h(X_{i})\}}(h(X_{m+1}))=c\right]\\
&\notag\quad={m\choose l}\theta\frac{(\theta+\alpha)_{(m-l)}}{(\theta)_{(m+1)}}(1-\alpha)_{(l)}{m-l\choose c-l}\\
&\notag\quad\quad\times\sum_{i=0}^{c-l}\sum_{j=0}^{m-c}\frac{\left(\frac{\theta+\alpha}{\alpha}\right)_{(i+j)}}{(\theta+\alpha)_{(m-l)}}\left(\frac{1}{J}\right)^{i}\left(1-\frac{1}{J}\right)^{j}\mathscr{C}(c-l,i;\alpha)\mathscr{C}(m-c,j;\alpha).
\end{align}
This completes the study of the numerator of \eqref{eq:maintoprovepy}. By combining \eqref{eq:maintoprovepy} with \eqref{eq:denom1py} and \eqref{eq:num1_2py} we obtain
\begin{align}\label{eq:final_1py}
&\text{Pr}\left[f_{X_{m+1}}=l\,|\,\sum_{i=1}^{m}\mathbbm{1}_{\{h(X_{i})\}}(h(X_{m+1}))=c\right]\\
&\notag\quad=\frac{\theta}{J}{c\choose l}\frac{(\theta+\alpha)_{(m-l)}}{(\theta)_{(m+1)}}(1-\alpha)_{(l)}\\
&\notag\quad\quad\times\frac{\sum_{i=0}^{c-l}\sum_{j=0}^{m-c}\frac{\left(\frac{\theta+\alpha}{\alpha}\right)_{(i+j)}}{(\theta+\alpha)_{(m-l)}}\left(\frac{1}{J}\right)^{i}\left(1-\frac{1}{J}\right)^{j}\mathscr{C}(c-l,i;\alpha)\mathscr{C}(m-c,j;\alpha)}{\sum_{i=0}^{c+1}\sum_{j=0}^{m-c}\frac{\left(\frac{\theta}{\alpha}\right)_{(i+j)}}{(\theta)_{(m+1)}}\left(\frac{1}{J}\right)^{i}\left(1-\frac{1}{J}\right)^{j}\mathscr{C}(c+1,i;\alpha)\mathscr{C}(m-c,j;\alpha)}\\
&\notag\quad=\frac{\theta}{J}{c\choose l}(1-\alpha)_{(l)}\frac{\sum_{i=0}^{c-l}\sum_{j=0}^{m-c}\left(\frac{\theta+\alpha}{\alpha}\right)_{(i+j)}\left(\frac{1}{J}\right)^{i}\left(1-\frac{1}{J}\right)^{j}\mathscr{C}(c-l,i;\alpha)\mathscr{C}(m-c,j;\alpha)}{\sum_{i=0}^{c+1}\sum_{j=0}^{m-c}\left(\frac{\theta}{\alpha}\right)_{(i+j)}\left(\frac{1}{J}\right)^{i}\left(1-\frac{1}{J}\right)^{j}\mathscr{C}(c+1,i;\alpha)\mathscr{C}(m-c,j;\alpha)}.
\end{align}
for $l=0,1,\ldots, c$. By an application of \citet[Equation 2.56 and Equation 2.60]{Cha(05)} it is easy to show that \eqref{eq:final_1py} is a proper distribution on $\{0,1,\ldots,c\}$. The proof is completed.

\section{Theorem \ref{teo_direct} from Theorem \ref{teo_pyp} with $\alpha=0$}\label{sec4app} 

We show how Theorem \ref{teo_pyp} reduces to Theorem \ref{teo_direct} by setting $\alpha=0$. First, we recall two identities involving the  generalized factorial coefficient $\mathscr{C}(m,k;\alpha)$ and the signless Stirling number of the first type. See \citet[Chapter 2]{Cha(05)} for details. In particular, it holds
\begin{equation}\label{eq_prop_ss}
\sum_{k=0}^{m}a^{k}|s(m,k)|=(a)_{(m)}
\end{equation} 
for $a>0$, and
\begin{equation}\label{eq_prop_gfc}
\lim_{\alpha\rightarrow0}\frac{\mathscr{C}(m,k;\alpha)}{\alpha^{k}}=|s(m,k)|.
\end{equation} 
Hereafter, we apply the identities \eqref{eq_prop_ss} and \eqref{eq_prop_gfc} in order to show that Theorem \ref{teo_pyp} reduces to Theorem \ref{teo_direct} by setting $\alpha=0$. In this respect, we rewrite the posterior distribution \eqref{eq:posterior_py} as follows
\begin{align*}
&\text{Pr}[f_{X_{m+1}} = l\,|\, C_{n, h_{n}(X_{m+1})}=c_{n}]\\
&\quad=\frac{\theta}{J}{c_{n}\choose l}(1-\alpha)_{(l)}\frac{\sum_{i=0}^{c_{n}-l}\sum_{j=0}^{m-c_{n}}\left(\frac{\theta+\alpha}{\alpha}\right)_{(i+j)}\left(\frac{1}{J}\right)^{i}\left(1-\frac{1}{J}\right)^{j}\mathscr{C}(c_{n}-l,i;\alpha)\mathscr{C}(m-c_{n},j;\alpha)}{\sum_{i=0}^{c_{n}+1}\sum_{j=0}^{m-c_{n}}\left(\frac{\theta}{\alpha}\right)_{(i+j)}\left(\frac{1}{J}\right)^{i}\left(1-\frac{1}{J}\right)^{j}\mathscr{C}(c_{n}+1,i;\alpha)\mathscr{C}(m-c_{n},j;\alpha)}.
\end{align*}
Then,
\begin{align*}
&\lim_{\alpha\rightarrow0}\text{Pr}[f_{X_{m+1}} = l\,|\, C_{n, h_{n}(X_{m+1})}=c_{n}]\\
&\quad=\lim_{\alpha\rightarrow0}\frac{\theta}{J}{c_{n}\choose l}(1-\alpha)_{(l)}\\
&\quad\quad\times\frac{\sum_{i=0}^{c_{n}-l}\sum_{j=0}^{m-c_{n}}\left(\frac{\theta+\alpha}{\alpha}\right)_{(i+j)}\alpha^{i+j}\left(\frac{1}{J}\right)^{i}\left(1-\frac{1}{J}\right)^{j}\frac{\mathscr{C}(c_{n}-l,i;\alpha)}{\alpha^{i}}\frac{\mathscr{C}(m-c_{n},j;\alpha)}{\alpha^{j}}}{\sum_{i=0}^{c_{n}+1}\sum_{j=0}^{m-c_{n}}\left(\frac{\theta}{\alpha}\right)_{(i+j)}\alpha^{i+j}\left(\frac{1}{J}\right)^{i}\left(1-\frac{1}{J}\right)^{j}\frac{\mathscr{C}(c_{n}+1,i;\alpha)}{\alpha^{i}}\frac{\mathscr{C}(m-c_{n},j;\alpha)}{\alpha^{j}}}\\
&\text{[by the identity \eqref{eq_prop_gfc}]}\\
&\quad=\frac{\theta}{J}{c_{n}\choose l}l!\\
&\quad\quad\times\frac{\sum_{i=0}^{c_{n}-l}\left(\frac{\theta}{J}\right)^{i}|s(c_{n}-l,i)|\sum_{j=0}^{m-c_{n}}\left(\theta\left(1-\frac{1}{J}\right)\right)^{j}|s(m-c_{n},j)|}{\sum_{i=0}^{c_{n}+1}\left(\frac{\theta}{J}\right)^{i}|s(c_{n}+1,i)|\sum_{j=0}^{m-c_{n}}\left(\theta\left(1-\frac{1}{J}\right)\right))^{j}|s(m-c_{n},j)|}\\
&\text{[by the identity \eqref{eq_prop_ss}]}\\
&\quad=\frac{\theta}{J}{c_{n}\choose l}l!\frac{\left(\frac{\theta}{J}\right)_{(c_{n}-l)}\left(\theta\left(1-\frac{1}{J}\right)\right)_{(m-c_{n})}}{\left(\frac{\theta}{J}\right)_{(c_{n}-l)}\left(\theta\left(1-\frac{1}{J}\right)\right)_{(m-c_{n})}}\\
&\quad=\frac{\theta}{J}\frac{\Gamma(c_{n}+1)\Gamma(c_{n}-l+\frac{\theta}{J})}{\Gamma(c_{n}-l+1)\Gamma(\frac{\theta}{J}+c_{n}+1)}\\
&\quad=\frac{\frac{\theta}{J}}{\frac{\theta}{J}+c_{n}}\frac{(c_{n}-l+1)_{(l)}}{(\frac{\theta}{J}+c_{n}-l)_{(l)}},
\end{align*}
which is the expression for the posterior distribution stated in Theorem \ref{teo_direct}. The proof is completed.

\section{Proof of Equation \eqref{eq:posterior_py_alternative}}\label{sec3app}

Let $X_{1:m}$ be a random sample from $P\sim\text{PYP}(\alpha,\theta;\nu)$, with $\alpha\in[0,1)$ and $\theta>-\alpha$, and let $K_{m}$ be the number of distinct types in $X_{1:m}$. We recall from \eqref{eq:distpy} that for $k=1,\ldots,m$ it holds
\begin{displaymath}
\text{Pr}[K_{m}=k]=\frac{\left(\frac{\theta}{\alpha}\right)_{(k)}}{(\theta)_{(m)}}\mathscr{C}(m,k;\alpha).
\end{displaymath}
Now, assuming $c_{n}>0$ and $m-c_{n}>0$, we rewrite the posterior distribution of Theorem \ref{teo_pyp} in terms of the distribution of $K_{m}$. In particular, for any $l=0,1,\ldots,c_{n}-1$ we can write that
\begin{align*}
&\text{Pr}[f_{X_{m+1}} = l\,|\, C_{n, h_{n}(X_{m+1})}=c_{n}]\\
&\quad=\frac{\theta}{J}{c_{n}\choose l}(1-\alpha)_{(l)}\frac{\sum_{i=0}^{c_{n}-l}\sum_{j=0}^{m-c_{n}}\left(\frac{\theta+\alpha}{\alpha}\right)_{(i+j)}\left(\frac{1}{J}\right)^{i}\left(1-\frac{1}{J}\right)^{j}\mathscr{C}(c_{n}-l,i;\alpha)\mathscr{C}(m-c_{n},j;\alpha)}{\sum_{i=0}^{c_{n}+1}\sum_{j=0}^{m-c_{n}}\left(\frac{\theta}{\alpha}\right)_{(i+j)}\left(\frac{1}{J}\right)^{i}\left(1-\frac{1}{J}\right)^{j}\mathscr{C}(c_{n}+1,i;\alpha)\mathscr{C}(m-c_{n},j;\alpha)}\\
&\quad=\frac{\theta}{J}{c_{n}\choose l}(1-\alpha)_{(l)}\frac{(\theta)_{(c_{n}-l)}(\theta)_{(m-c_{n})}}{(\theta)_{(c_{n}+1)}(\theta)_{(m-c_{n})}}\\
&\quad\quad\times\frac{\sum_{i=0}^{c_{n}-l}\sum_{j=0}^{m-c_{n}}\left(\frac{\theta+\alpha}{\alpha}\right)_{(i+j)}\frac{\left(\frac{1}{J}\right)^{i}\left(1-\frac{1}{J}\right)^{j}}{\left(\frac{\theta}{\alpha}\right)_{(i)}\left(\frac{\theta}{\alpha}\right)_{(j)}}\frac{\left(\frac{\theta}{\alpha}\right)_{(i)}}{(\theta)_{(c_{n}-l)}}\mathscr{C}(c_{n}-l,i;\alpha)\frac{\left(\frac{\theta}{\alpha}\right)_{(j)}}{(\theta)_{(m-c_{n})}}\mathscr{C}(m-c_{n},j;\alpha)}{\sum_{i=0}^{c_{n}+1}\sum_{j=0}^{m-c_{n}}\left(\frac{\theta}{\alpha}\right)_{(i+j)}\frac{\left(\frac{1}{J}\right)^{i}\left(1-\frac{1}{J}\right)^{j}}{\left(\frac{\theta}{\alpha}\right)_{(i)}\left(\frac{\theta}{\alpha}\right)_{(j)}}\frac{\left(\frac{\theta}{\alpha}\right)_{(i)}}{(\theta)_{(c_{n}+1)}}\mathscr{C}(c_{n}+1,i;\alpha)\frac{\left(\frac{\theta}{\alpha}\right)_{(j)}}{(\theta)_{(m-c_{n})}}\mathscr{C}(m-c_{n},j;\alpha)}\\
&\quad=\frac{\theta}{J}{c_{n}\choose l}(1-\alpha)_{(l)}\frac{(\theta)_{(c_{n}-l)}}{(\theta)_{(c_{n}+1)}}\\
&\quad\quad\times\frac{\sum_{i=0}^{c_{n}-l}\sum_{j=0}^{m-c_{n}}\left(\frac{\theta+\alpha}{\alpha}\right)_{(i+j)}\frac{\left(\frac{1}{J}\right)^{i}\left(1-\frac{1}{J}\right)^{j}}{\left(\frac{\theta}{\alpha}\right)_{(i)}\left(\frac{\theta}{\alpha}\right)_{(j)}}\text{Pr}[K_{c_{n}-l}=i]\text{Pr}[K_{m-c_{n}}=j]}{\sum_{i=0}^{c_{n}+1}\sum_{j=0}^{m-c_{n}}\left(\frac{\theta}{\alpha}\right)_{(i+j)}\frac{\left(\frac{1}{J}\right)^{i}\left(1-\frac{1}{J}\right)^{j}}{\left(\frac{\theta}{\alpha}\right)_{(i)}\left(\frac{\theta}{\alpha}\right)_{(j)}}\text{Pr}[K_{c_{n}+1}=i]\text{Pr}[K_{m-c_{n}}=j]}\\
&\quad=\frac{\theta}{J}{c_{n}\choose l}(1-\alpha)_{(l)}\frac{(\theta)_{(c_{n}-l)}\E\left[\left(\frac{\theta+\alpha}{\alpha}\right)_{(K_{c_{n}-l}+K_{m-c_{n}})}\frac{\left(\frac{1}{J}\right)^{K_{c_{n}-l}}\left(1-\frac{1}{J}\right)^{K_{m-c_{n}}}}{\left(\frac{\theta}{\alpha}\right)_{(K_{c_{n}-l})}\left(\frac{\theta}{\alpha}\right)_{(K_{m-c_{n}})}}\right]}{(\theta)_{(c_{n}+1)}\E\left[\left(\frac{\theta}{\alpha}\right)_{(K_{c_{n}+1}+K_{m-c_{n}})}\frac{\left(\frac{1}{J}\right)^{K_{c_{n}+1}}\left(1-\frac{1}{J}\right)^{K_{m-c_{n}}}}{\left(\frac{\theta}{\alpha}\right)_{(K_{c_{n}+1})}\left(\frac{\theta}{\alpha}\right)_{(K_{m-c_{n}})}}\right]}\\
&\quad=\frac{\frac{\alpha}{J}{c_{n}\choose l}(1-\alpha)_{(l)}}{(\theta+c_{n}-l)_{(l+1)}}\frac{\E\left[\frac{\Gamma\left(\frac{\theta+\alpha}{\alpha}+K_{c_{n}-l}+K_{m-c_{n}}\right)}{\Gamma\left(\frac{\theta}{\alpha}+K_{c_{n}-l}\right)\Gamma\left(\frac{\theta}{\alpha}+K_{m-c_{n}}\right)}\left(\frac{1}{J}\right)^{K_{c_{n}-l}}\left(1-\frac{1}{J}\right)^{K_{m-c_{n}}}\right]}{\E\left[\frac{\Gamma\left(\frac{\theta}{\alpha}+K_{c_{n}+1}+K_{m-c_{n}}\right)}{\Gamma\left(\frac{\theta}{\alpha}+K_{c_{n}+1}\right)\Gamma\left(\frac{\theta}{\alpha}+K_{m-c_{n}}\right)}\left(\frac{1}{J}\right)^{K_{c_{n}+1}}\left(1-\frac{1}{J}\right)^{K_{m-c_{n}}}\right]},
\end{align*}
where $K_{c_{n}-l}$ and $K_{m-c_{n}}$ in the numerator are independent random variables for any $l=0,1,\ldots,c_{n}-1$, and $K_{c_{n}+1}$ and $K_{m-c_{n}}$ in the denominator are independent random variables. For $l=c_{n}$
\begin{align*}
&\text{Pr}[f_{X_{m+1}} = c_{n}\,|\, C_{n, h_{n}(X_{m+1})}=c_{n}]\\
&\quad=\frac{\theta}{J}(1-\alpha)_{(c_{n})}\frac{\sum_{j=0}^{m-c_{n}}\left(\frac{\theta+\alpha}{\alpha}\right)_{(j)}\left(1-\frac{1}{J}\right)^{j}\mathscr{C}(m-c_{n},j;\alpha)}{\sum_{i=0}^{c_{n}+1}\sum_{j=0}^{m-c_{n}}\left(\frac{\theta}{\alpha}\right)_{(i+j)}\left(\frac{1}{J}\right)^{i}\left(1-\frac{1}{J}\right)^{j}\mathscr{C}(c_{n}+1,i;\alpha)\mathscr{C}(m-c_{n},j;\alpha)}\\
&\quad=\frac{\theta}{J}(1-\alpha)_{(c_{n})}\frac{(\theta)_{(m-c_{n})}}{(\theta)_{(c_{n}+1)}(\theta)_{(m-c_{n})}}\\
&\quad\quad\times\frac{\sum_{j=0}^{m-c_{n}}\left(\frac{\theta+\alpha}{\alpha}\right)_{(j)}\frac{\left(1-\frac{1}{J}\right)^{j}}{\left(\frac{\theta}{\alpha}\right)_{(j)}}\frac{\left(\frac{\theta}{\alpha}\right)_{(j)}}{(\theta)_{(m-c_{n})}}\mathscr{C}(m-c_{n},j;\alpha)}{\sum_{i=0}^{c_{n}+1}\sum_{j=0}^{m-c_{n}}\left(\frac{\theta}{\alpha}\right)_{(i+j)}\frac{\left(\frac{1}{J}\right)^{i}\left(1-\frac{1}{J}\right)^{j}}{\left(\frac{\theta}{\alpha}\right)_{(i)}\left(\frac{\theta}{\alpha}\right)_{(j)}}\frac{\left(\frac{\theta}{\alpha}\right)_{(i)}}{(\theta)_{(c_{n}+1)}}\mathscr{C}(c_{n}+1,i;\alpha)\frac{\left(\frac{\theta}{\alpha}\right)_{(j)}}{(\theta)_{(m-c_{n})}}\mathscr{C}(m-c_{n},j;\alpha)}\\
&\quad=\frac{\theta}{J}(1-\alpha)_{(c_{n})}\frac{1}{(\theta)_{(c_{n}+1)}}\\
&\quad\quad\times\frac{\sum_{j=0}^{m-c_{n}}\left(\frac{\theta+\alpha}{\alpha}\right)_{(j)}\frac{\left(1-\frac{1}{J}\right)^{j}}{\left(\frac{\theta}{\alpha}\right)_{(j)}}\text{Pr}[K_{m-c_{n}}=j]}{\sum_{i=0}^{c_{n}+1}\sum_{j=0}^{m-c_{n}}\left(\frac{\theta}{\alpha}\right)_{(i+j)}\frac{\left(\frac{1}{J}\right)^{i}\left(1-\frac{1}{J}\right)^{j}}{\left(\frac{\theta}{\alpha}\right)_{(i)}\left(\frac{\theta}{\alpha}\right)_{(j)}}\text{Pr}[K_{c_{n}+1}=i]\text{Pr}[K_{m-c_{n}}=j]}\\
&\quad=\frac{\theta}{J}(1-\alpha)_{(c_{n})}\frac{\E\left[\left(\frac{\theta+\alpha}{\alpha}\right)_{(K_{m-c_{n}})}\frac{\left(1-\frac{1}{J}\right)^{K_{m-c_{n}}}}{\left(\frac{\theta}{\alpha}\right)_{(K_{m-c_{n}})}}\right]}{(\theta)_{(c_{n}+1)}\E\left[\left(\frac{\theta}{\alpha}\right)_{(K_{c_{n}+1}+K_{m-c_{n}})}\frac{\left(\frac{1}{J}\right)^{K_{c_{n}+1}}\left(1-\frac{1}{J}\right)^{K_{m-c_{n}}}}{\left(\frac{\theta}{\alpha}\right)_{(K_{c_{n}+1})}\left(\frac{\theta}{\alpha}\right)_{(K_{m-c_{n}})}}\right]}\\
&\quad=\frac{\frac{\alpha}{J}(1-\alpha)_{(c_{n})}}{\Gamma(\theta/\alpha)(\theta)_{(c_{n}+1)}}\frac{\E\left[\frac{\Gamma\left(\frac{\theta+\alpha}{\alpha}+K_{m-c_{n}}\right)}{\Gamma\left(\frac{\theta}{\alpha}+K_{m-c_{n}}\right)}\left(1-\frac{1}{J}\right)^{K_{m-c_{n}}}\right]}{\E\left[\frac{\Gamma\left(\frac{\theta}{\alpha}+K_{c_{n}+1}+K_{m-c_{n}}\right)}{\Gamma\left(\frac{\theta}{\alpha}+K_{c_{n}+1}\right)\Gamma\left(\frac{\theta}{\alpha}+K_{m-c_{n}}\right)}\left(\frac{1}{J}\right)^{K_{c_{n}+1}}\left(1-\frac{1}{J}\right)^{K_{m-c_{n}}}\right]},
\end{align*}
where $K_{c_{n}+1}$ and $K_{m-c_{n}}$ are independent random variables. This completes the proof of Equation \eqref{eq:posterior_py_alternative}.

\section{An alternative expression for Equation \eqref{eq:posterior_py}}\label{sec3app1}

For any $\alpha\in(0,1)$, an alternative expression for \eqref{eq:posterior_py} may be given in terms of the distribution of exponentially tilted $\alpha$-stable random variables \citep{Zol(86)}. In particular, if $g_{\alpha}$ denotes the density function of a positive $\alpha$-stable distribution, then for any $c>0$ an exponentially tilted $\alpha$-stable random variable is defined as the random variable $T_{\alpha,c}$ whose distribution has density function $f_{S_{\alpha,c}}(x)\propto \text{exp}\{-c^{1/\alpha}x\}g_{\alpha}(x)\mathbbm{1}_{\mathbb{R}^{+}}(x)$. If $c_{n}>0$, then for $l=0,1,\ldots,c_{n}$
\begin{align*}
&\text{Pr}[f_{X_{m+1}} = l\,|\, C_{n, h_{n}(X_{m+1})}=c_{n}]\\
&\quad=\frac{\theta}{J}{c_{n}\choose l}(1-\alpha)_{(l)}\frac{\sum_{i=0}^{c_{n}-l}\sum_{j=0}^{m-c_{n}}\left(\frac{\theta+\alpha}{\alpha}\right)_{(i+j)}\left(\frac{1}{J}\right)^{i}\left(1-\frac{1}{J}\right)^{j}\mathscr{C}(c_{n}-l,i;\alpha)\mathscr{C}(m-c_{n},j;\alpha)}{\sum_{i=0}^{c_{n}+1}\sum_{j=0}^{m-c_{n}}\left(\frac{\theta}{\alpha}\right)_{(i+j)}\left(\frac{1}{J}\right)^{i}\left(1-\frac{1}{J}\right)^{j}\mathscr{C}(c_{n}+1,i;\alpha)\mathscr{C}(m-c_{n},j;\alpha)}\\
&\quad=\frac{\theta}{J}{c_{n}\choose l}(1-\alpha)_{(l)}\\
&\quad\quad\times\frac{\frac{1}{\Gamma\left(\frac{\theta+\alpha}{\alpha}\right)}\int_{0}^{+\infty}x^{\frac{\theta+\alpha}{\alpha}-1}\text{e}^{-x}\left(\sum_{i=0}^{c_{n}-l}\sum_{j=0}^{m-c_{n}}\left(\frac{x}{J}\right)^{i}\left(x\left(1-\frac{1}{J}\right)\right)^{j}\mathscr{C}(c_{n}-l,i;\alpha)\mathscr{C}(m-c_{n},j;\alpha)\right)\ddr x}{\frac{1}{\Gamma\left(\frac{\theta}{\alpha}\right)}\int_{0}^{+\infty}x^{\frac{\theta}{\alpha}-1}\text{e}^{-x}\left(\sum_{i=0}^{c_{n}+1}\sum_{j=0}^{m-c_{n}}\left(\frac{x}{J}\right)^{i}\left(x\left(1-\frac{1}{J}\right)\right)^{j}\mathscr{C}(c_{n}+1,i;\alpha)\mathscr{C}(m-c_{n},j;\alpha)\right)\ddr x}
\end{align*}
By means of \citet[Equation 13]{Fav(15)} we can write the numerator and the denominator of the previous expression in terms of the distribution of $T_{\alpha,c}$, for suitable choices of $c$. That is, 
\begin{align}\label{eq:posterior_py_alternative1}
&\notag\frac{\theta}{J}{c_{n}\choose l}(1-\alpha)_{(l)}\\
&\notag\quad\times\frac{\frac{1}{\Gamma\left(\frac{\theta+\alpha}{\alpha}\right)}\int_{0}^{+\infty}x^{\frac{\theta+\alpha}{\alpha}-1}\text{e}^{-x}\left(\sum_{i=0}^{c_{n}-l}\sum_{j=0}^{m-c_{n}}\left(\frac{x}{J}\right)^{i}\left(x\left(1-\frac{1}{J}\right)\right)^{j}\mathscr{C}(c_{n}-l,i;\alpha)\mathscr{C}(m-c_{n},j;\alpha)\right)\ddr x}{\frac{1}{\Gamma\left(\frac{\theta}{\alpha}\right)}\int_{0}^{+\infty}x^{\frac{\theta}{\alpha}-1}\text{e}^{-x}\left(\sum_{i=0}^{c_{n}+1}\sum_{j=0}^{m-c_{n}}\left(\frac{x}{J}\right)^{i}\left(x\left(1-\frac{1}{J}\right)\right)^{j}\mathscr{C}(c_{n}+1,i;\alpha)\mathscr{C}(m-c_{n},j;\alpha)\right)\ddr x}\\
&\notag\quad=\frac{\theta}{J}{c_{n}\choose l}(1-\alpha)_{(l)}\\
&\notag\quad\quad\times\frac{\frac{1}{\Gamma\left(\frac{\theta+\alpha}{\alpha}\right)}\int_{0}^{+\infty}x^{\frac{\theta+\alpha}{\alpha}-1}\text{e}^{-x}\left(\left(\frac{x}{J}\right)^{\frac{c_{n}-l}{\alpha}}\E\left[T^{c_{n}-l}_{\alpha,\frac{x}{J}}\right]\left(x\left(1-\frac{1}{J}\right)\right)^{\frac{m-c_{n}}{\alpha}}\E\left[T^{m-c_{n}}_{\alpha,x\left(1-\frac{1}{J}\right)}\right]\right)\ddr x}{\frac{1}{\Gamma\left(\frac{\theta}{\alpha}\right)}\int_{0}^{+\infty}x^{\frac{\theta}{\alpha}-1}\text{e}^{-x}\left(\left(\frac{x}{J}\right)^{\frac{c_{n}+1}{\alpha}}\E\left[T^{c_{n}+1}_{\alpha,\frac{x}{J}}\right]\left(x\left(1-\frac{1}{J}\right)\right)^{\frac{m-c_{n}}{\alpha}}\E\left[T^{m-c_{n}}_{\alpha,x\left(1-\frac{1}{J}\right)}\right]\right)\ddr x}\\
&\notag\quad=\frac{\theta}{J}{c_{n}\choose l}(1-\alpha)_{(l)}\\
&\notag\quad\quad\times\frac{\frac{\left(\frac{1}{J}\right)^{\frac{c_{n}-l}{\alpha}}\left(1-\frac{1}{J}\right)^{\frac{m-c_{n}}{\alpha}}}{\Gamma\left(\frac{\theta+\alpha}{\alpha}\right)}\int_{0}^{+\infty}x^{\frac{\theta+\alpha+m-l}{\alpha}-1}\text{e}^{-x}\left(\E\left[T^{c_{n}-l}_{\alpha,\frac{x}{J}}\right]\E\left[T^{m-c_{n}}_{\alpha,x\left(1-\frac{1}{J}\right)}\right]\right)\ddr x}{\frac{\left(\frac{1}{J}\right)^{\frac{c_{n}+1}{\alpha}}\left(1-\frac{1}{J}\right)^{\frac{m-c_{n}}{\alpha}}}{\Gamma\left(\frac{\theta}{\alpha}\right)}\int_{0}^{+\infty}x^{\frac{\theta+m+1}{\alpha}-1}\text{e}^{-x}\left(\E\left[T^{c_{n}+1}_{\alpha,\frac{x}{J}}\right]\E\left[T^{m-c_{n}}_{\alpha,x\left(1-\frac{1}{J}\right)}\right]\right)\ddr x}\\
&\quad=\frac{\theta}{J}{c_{n}\choose l}(1-\alpha)_{(l)}\\
&\notag\quad\quad\times\frac{\left(\frac{1}{J}\right)^{\frac{c_{n}-l}{\alpha}}\left(\frac{\theta+\alpha}{\alpha}\right)_{(m-l)}\int_{0}^{+\infty}\E\left[T^{c_{n}-l}_{\alpha,\frac{x}{J}}\right]\E\left[T^{m-c_{n}}_{\alpha,x\left(1-\frac{1}{J}\right)}\right]f_{G_{\frac{\theta+\alpha+m-l}{\alpha},1}}(x)\ddr x}{\left(\frac{1}{J}\right)^{\frac{c_{n}+1}{\alpha}}\left(\frac{\theta}{\alpha}\right)_{(m+1)}\int_{0}^{+\infty}\E\left[T^{c_{n}+1}_{\alpha,\frac{x}{J}}\right]\E\left[T^{m-c_{n}}_{\alpha,x\left(1-\frac{1}{J}\right)}\right]f_{G_{\frac{\theta+m+1}{\alpha},1}}(x)\ddr x},
\end{align}
where $f_{G_{a,b}}$ is the density function of a Gamma distribution with parameter $(a,b)$. Equation \eqref{eq:posterior_py_alternative1} allows for an MC evaluation of \eqref{eq:posterior_py}, which requires to sample from a Gamma distribution and to sample $T_{\alpha,c}$, for suitable choices of $c$. See \citet{Dev(09)} and references therein.

\section{Proof Equation \eqref{local_limit} and Equation \eqref{local_bino}}\label{app_local}

Under the setting of Theorem \ref{teo_pyp}, we consider $m\rightarrow+\infty$, while $c_{n}$ is fixed. For any $l=0,1,\ldots,c_{n}$
\begin{align}\label{local_prima}
&\notag\text{Pr}[f_{X_{m+1}} = l\,|\, C_{n, h_{n}(X_{m+1})}=c_{n}]\\
&\notag\quad=\frac{\theta}{J}{c_{n}\choose l}(1-\alpha)_{(l)}\frac{\sum_{i=0}^{c_{n}-l}\sum_{j=0}^{m-c_{n}}\left(\frac{\theta+\alpha}{\alpha}\right)_{(i+j)}\left(\frac{1}{J}\right)^{i}\mathscr{C}(c_{n}-l,i;\alpha)\frac{\left(1-\frac{1}{J}\right)^{j}\mathscr{C}(m-c_{n},j;\alpha)}{\sum_{j=0}^{m-c_{n}}\left(1-\frac{1}{J}\right)^{j}\mathscr{C}(m-c_{n},j;\alpha)}}{\sum_{i=0}^{c_{n}+1}\sum_{j=0}^{m-c_{n}}\left(\frac{\theta}{\alpha}\right)_{(i+j)}\left(\frac{1}{J}\right)^{i}\mathscr{C}(c_{n}+1,i;\alpha)\frac{\left(1-\frac{1}{J}\right)^{j}\mathscr{C}(m-c_{n},j;\alpha)}{\sum_{j=0}^{m-c_{n}}\left(1-\frac{1}{J}\right)^{j}\mathscr{C}(m-c_{n},j;\alpha)}}\\
&\quad=\frac{\theta}{J}{c_{n}\choose l}(1-\alpha)_{(l)}\frac{\sum_{i=0}^{c_{n}-l}\left(\frac{\theta+\alpha}{\alpha}\right)_{(i)}\left(\frac{1}{J}\right)^{i}\mathscr{C}(c_{n}-l,i;\alpha)\sum_{j=0}^{m-c_{n}}\left(\frac{\theta+\alpha}{\alpha}+i\right)_{(j)}\frac{\left(1-\frac{1}{J}\right)^{j}\mathscr{C}(m-c_{n},j;\alpha)}{\sum_{j=0}^{m-c_{n}}\left(1-\frac{1}{J}\right)^{j}\mathscr{C}(m-c_{n},j;\alpha)}}{\sum_{i=0}^{c_{n}+1}\left(\frac{\theta}{\alpha}\right)_{(i)}\left(\frac{1}{J}\right)^{i}\mathscr{C}(c_{n}+1,i;\alpha)\sum_{j=0}^{m-c_{n}}\left(\frac{\theta}{\alpha}+i\right)_{(j)}\frac{\left(1-\frac{1}{J}\right)^{j}\mathscr{C}(m-c_{n},j;\alpha)}{\sum_{j=0}^{m-c_{n}}\left(1-\frac{1}{J}\right)^{j}\mathscr{C}(m-c_{n},j;\alpha)}}.
\end{align}
Now, consider the numerator of the last member in \eqref{local_prima}. From \citet[Lemma 2]{DF(20a)}, as $m\rightarrow+\infty$
\begin{align}\label{local_seconda}
&\lim_{m\rightarrow+\infty}\sum_{j=0}^{m-c_{n}}\left(\frac{\theta+\alpha}{\alpha}+i\right)_{(j)}\frac{\left(1-\frac{1}{J}\right)^{j}\mathscr{C}(m-c_{n},j;\alpha)}{\sum_{j\geq 1}\left(1-\frac{1}{J}\right)^{j}\mathscr{C}(m-c_{n},j;\alpha)}\\
&\notag\quad=\sum_{j\geq 1}\left(\frac{\theta+\alpha}{\alpha}+i\right)_{(j)}\text{e}^{-\left(1-\frac{1}{J}\right)}\frac{\left(1-\frac{1}{J}\right)^{j-1}}{(j-1)!}\\
&\notag\quad=\text{e}^{-\left(1-\frac{1}{J}\right)}\frac{\left(\frac{\theta+\alpha}{\alpha}+i\right)}{\left(\frac{1}{J}\right)^{\frac{\theta+\alpha}{\alpha}+i+1}}.
\end{align}
Now, consider the denominator of the last member in \eqref{local_prima}. Again, from \citet[Lemma 2]{DF(20a)}, as $m\rightarrow+\infty$
\begin{align}\label{local_terza}
&\lim_{m\rightarrow+\infty}\sum_{j=0}^{m-c_{n}}\left(\frac{\theta}{\alpha}+i\right)_{(j)}\frac{\left(1-\frac{1}{J}\right)^{j}\mathscr{C}(m-c_{n},j;\alpha)}{\sum_{j=0}^{m-c_{n}}\left(1-\frac{1}{J}\right)^{j}\mathscr{C}(m-c_{n},j;\alpha)}\\
&\notag\quad=\sum_{j\geq 1}\left(\frac{\theta}{\alpha}+i\right)_{(j)}\text{e}^{-\left(1-\frac{1}{J}\right)}\frac{\left(1-\frac{1}{J}\right)^{j-1}}{(j-1)!}\\
&\notag\quad=\text{e}^{-\left(1-\frac{1}{J}\right)}\frac{\left(\frac{\theta}{\alpha}+i\right)}{\left(\frac{1}{J}\right)^{\frac{\theta}{\alpha}+i+1}}.
\end{align}
Then, by combining \eqref{local_prima} with \eqref{local_seconda} and \eqref{local_terza}, for any $l=0,1,\ldots,c_{n}$, as $m\rightarrow+\infty$ we can write
\begin{align*}
&\lim_{m\rightarrow+\infty}\text{Pr}[f_{X_{m+1}} = l\,|\, C_{n, h_{n}(X_{m+1})}=c_{n}]\\
&\quad=\frac{\theta}{J}{c_{n}\choose l}(1-\alpha)_{(l)}\frac{\sum_{i=0}^{c_{n}-l}\left(\frac{\theta+\alpha}{\alpha}\right)_{(i)}\left(\frac{1}{J}\right)^{i}\mathscr{C}(c_{n}-l,i;\alpha)\text{e}^{-\left(1-\frac{1}{J}\right)}\frac{\left(\frac{\theta+\alpha}{\alpha}+i\right)}{\left(\frac{1}{J}\right)^{\frac{\theta+\alpha}{\alpha}+i+1}}}{\sum_{i=0}^{c_{n}+1}\left(\frac{\theta}{\alpha}\right)_{(i)}\left(\frac{1}{J}\right)^{i}\mathscr{C}(c_{n}+1,i;\alpha)\text{e}^{-\left(1-\frac{1}{J}\right)}\frac{\left(\frac{\theta}{\alpha}+i\right)}{\left(\frac{1}{J}\right)^{\frac{\theta}{\alpha}+i+1}}}\\
&\quad=\frac{\theta}{J}{c_{n}\choose l}(1-\alpha)_{(l)}\frac{\left(\frac{1}{J}\right)^{-\frac{\theta+\alpha}{\alpha}-1}}{\left(\frac{1}{J}\right)^{-\frac{\theta}{\alpha}-1}}\frac{\left(\theta+\alpha\right)_{(c_{n}-l)}}{\left(\theta\right)_{(c_{n}+1)}}\frac{\sum_{i=0}^{c_{n}-l}\frac{\left(\frac{\theta+\alpha}{\alpha}\right)_{(i)}}{\left(\theta+\alpha\right)_{(c_{n}-l)}}\mathscr{C}(c_{n}-l,i;\alpha)\left(\frac{\theta+\alpha}{\alpha}+i\right)}{\sum_{i=0}^{c_{n}+1}\frac{\left(\frac{\theta}{\alpha}\right)_{(i)}}{(\theta)_{c_{n}+1}}\mathscr{C}(c_{n}+1,i;\alpha)\left(\frac{\theta}{\alpha}+i\right)}\\
&\quad=\frac{\theta}{J}{c_{n}\choose l}(1-\alpha)_{(l)}\frac{\left(\frac{1}{J}\right)^{-\frac{\theta+\alpha}{\alpha}-1}}{\left(\frac{1}{J}\right)^{-\frac{\theta}{\alpha}-1}}\frac{\left(\theta+\alpha\right)_{(c_{n}-l)}}{\left(\theta\right)_{(c_{n}+1)}}\frac{\sum_{i=0}^{c_{n}-l}\frac{\left(\frac{\theta+\alpha}{\alpha}\right)_{(i)}}{\left(\theta+\alpha\right)_{(c_{n}-l)}}\mathscr{C}(c_{n}-l,i;\alpha)\left(\frac{\theta+\alpha}{\alpha}+i\right)}{\sum_{i=0}^{c_{n}+1}\frac{\left(\frac{\theta}{\alpha}\right)_{(i)}}{(\theta)_{c_{n}+1}}\mathscr{C}(c_{n}+1,i;\alpha)\left(\frac{\theta}{\alpha}+i\right)}\\
&\text{[by \citet[Equation 3.13]{Pit(06)}]}\\
&\quad=\frac{\theta}{J}{c_{n}\choose l}(1-\alpha)_{(l)}\frac{\left(\frac{1}{J}\right)^{-\frac{\theta+\alpha}{\alpha}-1}}{\left(\frac{1}{J}\right)^{-\frac{\theta}{\alpha}-1}}\frac{\left(\theta+\alpha\right)_{(c_{n}-l)}}{\left(\theta\right)_{(c_{n}+1)}}\frac{\frac{\theta+\alpha}{\alpha}+\frac{(\theta+2\alpha)_{(c_{n-l})}}{\alpha(\theta+\alpha+1)_{(c_{n}-l-1)}}-\frac{\theta+\alpha}{\alpha}}{\frac{\theta}{\alpha}+\frac{(\theta+\alpha)_{(c_{n}+1)}}{\alpha(\theta+1)_{(c_{n}+1-1)}}-\frac{\theta}{\alpha}}\\
&\quad=\frac{\theta}{J}{c_{n}\choose l}(1-\alpha)_{(l)}\left(\frac{1}{J}\right)^{-1}\frac{\left(\theta+\alpha\right)_{(c_{n}-l)}}{\left(\theta\right)_{(c_{n}+1)}}\frac{\frac{(\theta+2\alpha)_{(c_{n-l})}}{(\theta+\alpha+1)_{(c_{n}-l-1)}}}{\frac{(\theta+\alpha)_{(c_{n}+1)}}{(\theta+1)_{(c_{n}+1-1)}}}\\
&\quad={c_{n}\choose l}(1-\alpha)_{(l)}\frac{(\theta+2\alpha)_{(c_{n}-l)}}{(\theta+\alpha+1)_{(c_{n})}}.
\end{align*}
This completes the proof of Equation \eqref{local_limit}. Equation \eqref{local_bino} follows by a direct calculation from \eqref{local_limit}.

\section{Proof of Proposition \ref{asymp_py}}\label{proof_asymp_py}

Let $B_{a,b}$ be a Beta random variable with parameter $(a,b)$, and denote by $f_{B_{a,b}}$ the density function of the distribution of $B_{a,b}$. We start by some  considerations on the distribution of  $B_{a,b}$:
\begin{itemize}
\item[i)]
\begin{displaymath}
\frac{\Gamma(\theta+\alpha+m-l)\Gamma(1-\alpha+l)}{\Gamma(\theta+m+1)} = \int_0^1 t^{\theta+\alpha+m-l-1} (1-t)^{l-\alpha} \ddr t;
\end{displaymath}
\item[ii)]
\begin{displaymath}
\frac{\Gamma(\theta+\alpha+m-l)\Gamma(1-\alpha+l)}{\Gamma(\theta+m+1)} = \frac{\Gamma(\theta+\alpha)\Gamma(1-\alpha)}{\Gamma(\theta+1)} \E[B_{\theta+\alpha,1-\alpha}^{m-l}(1-B_{\theta+\alpha,1-\alpha}^l)].
\end{displaymath}
\end{itemize}
Moreover, we observe that we can rewrite the numerator and the denominator of \eqref{eq:posterior_py_alternative} as follows
\begin{align*}
&\int_{0}^{+\infty}\int_{0}^{+\infty} g_{\alpha}(h)g_{\alpha}(x) x^{-\theta-\alpha}
\frac{\left(\frac{h}{x}(J-1)^{\frac{1}{\alpha}}\right)^{m-c}}{\left(\frac{h}{x}(J-1)^{\frac{1}{\alpha}}+1\right)^{\theta+m-l+\alpha}}\ddr x\ddr h \\
&\quad= \frac{\Gamma(2 + \theta/\alpha)}{\Gamma(1+\theta+\alpha)} \int_{0}^{+\infty}\int_{0}^{+\infty} f_{S_{\alpha,0}}(h) f_{S_{\alpha,\theta+\alpha}}(x)
\frac{\left(\frac{h}{x}(J-1)^{\frac{1}{\alpha}}\right)^{m-c}}{\left(\frac{h}{x}(J-1)^{\frac{1}{\alpha}}+1\right)^{\theta+m-l+\alpha}}\ddr x\ddr h\\
&\quad=\frac{\Gamma(2 + \theta/\alpha)}{\Gamma(1+\theta+\alpha)} \int_{0}^{+\infty} \left[\frac{x^{m-c}}{(x+1)^{\theta+m-l+\alpha}}\right] f_{Z_{\alpha,\theta+\alpha}}(x)\ddr x
\end{align*}
and
\begin{align*}
& \int_{0}^{+\infty}\int_{0}^{+\infty}g_{\alpha}(h)g_{\alpha}(x)x^{-\theta} 
\frac{\left(\frac{h}{x}(J-1)^{\frac{1}{\alpha}}\right)^{m-c}}{\left(\frac{h}{x}(J-1)^{\frac{1}{\alpha}}+1\right)^{\theta+m+1}}\ddr x\ddr h \\
&\quad= \frac{\Gamma(1 + \theta/\alpha)}{\Gamma(1+\theta)} \int_{0}^{+\infty}\int_{0}^{+\infty}f_{S_{\alpha,0}}(h)f_{S_{\alpha,\theta}}(x)
\frac{\left(\frac{h}{x}(J-1)^{\frac{1}{\alpha}}\right)^{m-c}}{\left(\frac{h}{x}(J-1)^{\frac{1}{\alpha}}+1\right)^{\theta+m+1}}\ddr x\ddr h\\
&\quad=\frac{\Gamma(1 + \theta/\alpha)}{\Gamma(1+\theta)} \int_{0}^{+\infty} \left[\frac{x^{m-c}}{(x+1)^{\theta+m+1}}\right] f_{W_{\alpha,\theta}}(x)\ddr x,
\end{align*}
respectively. First, we prove that the distribution $\text{Pr}[f_{X_{m+1}}\in\cdot\,|\, C_{n, h_{n}(X_{m+1})}=c_{n}]$ admits a representation in terms of a suitable mixture of Binomial distribution. In particular, we write
\begin{align}\label{parz_binmix}
&\notag\text{Pr}[f_{X_{m+1}} = l\,|\, C_{n, h_{n}(X_{m+1})}=c_{n}]\\
&\notag\quad=\frac{\theta}{J}{c_{n}\choose l}(1-\alpha)_{(l)}\frac{\sum_{i=0}^{c_{n}-l}\sum_{j=0}^{m-c_{n}}\left(\frac{\theta+\alpha}{\alpha}\right)_{(i+j)}\left(\frac{1}{J}\right)^{i}\left(1-\frac{1}{J}\right)^{j}\mathscr{C}(c_{n}-l,i;\alpha)\mathscr{C}(m-c_{n},j;\alpha)}{\sum_{i=0}^{c_{n}+1}\sum_{j=0}^{m-c_{n}}\left(\frac{\theta}{\alpha}\right)_{(i+j)}\left(\frac{1}{J}\right)^{i}\left(1-\frac{1}{J}\right)^{j}\mathscr{C}(c_{n}+1,i;\alpha)\mathscr{C}(m-c_{n},j;\alpha)}\\
&\notag\quad=\frac{\theta}{J}{c_{n}\choose l}\frac{(\theta+\alpha)_{(m-l)}}{(\theta)_{(m+1)}}(1-\alpha)_{(l)}\\
&\notag\quad\quad\times\frac{\sum_{i=0}^{m-c_{n}}{m-c\choose i}(-1)^{m-c_{n}-i}\sum_{k=0}^{m-l-i}\frac{\left(\frac{\theta+\alpha}{\alpha}\right)_{(k)}}{(\theta+\alpha)_{(m-l-i)}}\frac{1}{J^{k}}\mathscr{C}(m-l-i,k;\alpha)}{\sum_{i=0}^{m-c_{n}}{m-c_{n}\choose i}(-1)^{m-c_{n}-i}\sum_{k=0}^{m-i+1}\frac{\left(\frac{\theta}{\alpha}\right)_{(k)}}{(\theta)_{(m-i+1)}}\frac{1}{J^{k}}\mathscr{C}(m-i+1,k;\alpha)}\\
&\notag\quad=\alpha{c_{n}\choose l}\frac{\Gamma(\theta+\alpha+m-l)}{\Gamma(\theta+m+1)}(1-\alpha)_{(l)}\frac{\int_{(0,+\infty)^{2}}g_{\alpha}(h)g_{\alpha}(x)x^{-\theta-\alpha}\frac{\left(\frac{h}{x}(J-1)^{\frac{1}{\alpha}}\right)^{m-c_{n}}}{\left(\frac{h}{x}(J-1)^{\frac{1}{\alpha}}+1\right)^{\theta+m-l+\alpha}}\ddr x\ddr h}{\int_{(0,+\infty)^{2}}g_{\alpha}(h)g_{\alpha}(x)x^{-\theta}\frac{\left(\frac{h}{x}(J-1)^{\frac{1}{\alpha}}\right)^{m-c_{n}}}{\left(\frac{h}{x}(J-1)^{\frac{1}{\alpha}}+1\right)^{\theta+m+1}}\ddr x\ddr h}\\
&\notag\quad={c_{n}\choose l} \int_0^1 t^{m-l} (1-t)^l \frac{\int_{0}^{+\infty} \left[\frac{x^{m-c_{n}}}{(x+1)^{\theta+m-l+\alpha}}\right] f_{Z_{\alpha,\theta+\alpha}}(x)\ddr x}{\int_{0}^{+\infty} \left[\frac{x^{m-c_{n}}}{(x+1)^{\theta+m+1}}\right] f_{W_{\alpha,\theta}}(x)\ddr x}f_{B_{\theta+\alpha,1-\alpha}}(t)\ddr t\\
&\notag\quad=\frac{1}{D(m,c_{n}; \alpha,\theta,J)} \int_0^1 \int_{0}^{+\infty} \left[ {c_{n}\choose l} (1-t)^l \left(\frac{t}{x+1}\right)^{c_{n}-l} \right] \frac{\left(\frac{tx}{x+1}\right)^{m-c_{n}}}{(x+1)^{\theta+\alpha}} f_{B_{\theta+\alpha,1-\alpha}}(t) f_{Z_{\alpha,\theta+\alpha}}(x) \ddr t\ddr x\\
&\notag\quad= \frac{1}{D(m,c_{n}; \alpha,\theta,J)} \int_0^1 \int_{0}^{+\infty} \left[ {c_{n}\choose l} (1-t)^l \left(\frac{t}{x+1}\right)^{c_{n}-l} \right] \frac{\left(\frac{tx}{x+1}\right)^{m-c_{n}}}{(x+1)^{\theta+\alpha}} f_{B_{\theta+\alpha,1-\alpha}}(t) f_{Z_{\alpha,\theta+\alpha}}(x) \ddr t\ddr x\\
&\notag\quad= \frac{\frac{\Gamma(\theta+1)\Gamma(m-l+\theta+\alpha)\Gamma(l+1-\alpha)}{\Gamma(\theta+\alpha)\Gamma(1-\alpha)\Gamma(m+\theta+1)}}{D(m,c_{n}; \alpha,\theta,J)} \int_{0}^{+\infty}  {c\choose l} (x+1)^{l-c_{n}} \frac{\left(\frac{x}{x+1}\right)^{m-c_{n}}}{(x+1)^{\theta+\alpha}} f_{Z_{\alpha,\theta+\alpha}}(x) \ddr x\\
&\quad= \frac{\frac{\Gamma(\theta+1)\Gamma(m-l+\theta+\alpha)\Gamma(l+1-\alpha)}{\Gamma(\theta+\alpha)\Gamma(1-\alpha)\Gamma(m+\theta+1)}}{D(m,c_{n}; \alpha,\theta,J)}{c_{n}\choose l} \int_{0}^{+\infty}x^{m-c_{n}}(x+1)^{l-m-\theta-\alpha} f_{Z_{\alpha,\theta+\alpha}}(x)\ddr x,
\end{align}
where
\begin{displaymath}
D(m,c_{n}; \alpha,\theta,J)=\int_{0}^{+\infty} \left[\frac{x^{m-c_{n}}}{(x+1)^{\theta+m+1}}\right] f_{W_{\alpha,\theta}}(x)\ddr x.
\end{displaymath}
It is easy to show that \eqref{parz_binmix} is mixture of Binomial distributions. In particular, from \eqref{parz_binmix} we write
\begin{align}\label{parz_binmix_2}
&\text{Pr}[f_{X_{m+1}} = l\,|\, C_{n, h_{n}(X_{m+1})}=c_{n}]\\
&\notag\quad=\frac{1}{D(m,c_{n}; \alpha,\theta,J)}\\
&\notag\quad\quad\times \int_0^1 \int_{0}^{+\infty} \left[{c_{n}\choose l}\left(\frac{1-t}{\frac{x+1-xt}{x+1}}\right)^{l}\left(\frac{\frac{t}{x+1}}{\frac{x+1-xt}{x+1}}\right)^{c_{n}-l}\right]\frac{\left(\frac{x+1-xt}{x+1}\right)^{c_{n}} \left(\frac{tx}{x+1}\right)^{m-c_{n}}}{(x+1)^{\theta+\alpha}} f_{B_{\theta+\alpha,1-\alpha}}(t) f_{Z_{\alpha,\theta+\alpha}}(x) \ddr t\ddr x,
\end{align}
Now, $F_{X_{m+1}}$ be a random variable with distribution \eqref{parz_binmix_2} and compute the moment of order $r$ of $F_{X_{m+1}}$. From the representation of the distribution of $F_{X_{m+1}}$ as a mixture of Binomial distribution,  
\begin{align*}
&\E[(F_{X_{m+1}})^{r}]\\
&\quad=\E[f^{r}_{X_{m+1}}\,|\, C_{n, h_{n}(X_{m+1})}=c_{n}]\\
&\quad=\frac{1}{D(m,c_{n}; \alpha,\theta,J)}\\
&\quad\quad\times \int_0^1 \int_{0}^{+\infty} \left[\sum_{l=0}^{c_{n}} {c_{n}\choose l} l^r (1-t)^l \left(\frac{t}{x+1}\right)^{c_{n}-l} \right] 
\frac{\left(\frac{tx}{x+1}\right)^{m-c_{n}}}{(x+1)^{\theta+\alpha}} f_{B_{\theta+\alpha,1-\alpha}}(t) f_{Z_{\alpha,\theta+\alpha}}(x) \ddr t\ddr x.
\end{align*}
Now, in the previous expression, we consider the summation within brackets. Recall that the $(u,v)$-th Stirling number of the second type, here denoted by $S(u,v)$, is defined as the $v$-th coefficient in the expansion of $t^{u}$ into falling factorials, i.e. $t^{u}=\sum_{0\leq v\leq u}S(u,v)(t)_{[v]}$; moreover, it is assumed: $S(0,0)=1$, $S(u,0)=0$ for $u>0$ and $S(u,v)=0$ for $v>u$.  Then, we write
\begin{align*}
&\sum_{l=0}^{c_{n}} {c_{n}\choose l} l^r (1-t)^l \left(\frac{t}{x+1}\right)^{c_{n}-l}\\
&\quad= \sum_{l=0}^{c_{n}} \left(\sum_{k=0}^r S(r,k) k! {l\choose k}\right) {c_{n}\choose l} (1-t)^l \left(\frac{t}{x+1}\right)^{c_{n}-l} \\
&\quad= \sum_{k=0}^r S(r,k) \frac{c_{n}!}{(c_{n}-k)!} (1-t)^k \sum_{l=k}^{c_{n}} \binom{c_{n}-k}{l-k}  (1-t)^{l-k} \left(\frac{t}{x+1}\right)^{c_{n}-l} \\
&\quad= \sum_{k=0}^r S(r,k) \frac{c_{n}!}{(c_{n}-k)!} (1-t)^k \sum_{j=0}^{c_{n}-k} \binom{c_{n}-k}{j}  (1-t)^j \left(\frac{t}{x+1}\right)^{c_{n}-k-j} \\
&\quad= \sum_{k=0}^r S(r,k) \frac{c_{n}!}{(c_{n}-k)!} (1-t)^k \left( 1-t + \frac{t}{x+1}\right)^{c_{n}-k} \\
&\quad= c_{n}^r (1-t)^r \left( 1-t + \frac{t}{x+1}\right)^{c_{n}-r} + O(c_{n}^{r-1}),
\end{align*}
where $O(c_{n}^{r-1})$ in the last identity is intended as $c_{n}\rightarrow+\infty$. Accordingly, we can write the following
\begin{align}\label{Mom_r}
&\E\left[\left(\frac{F_{X_{m+1}}}{c_{n}}\right)^{r}\right]\\
&\notag\quad=\E\left[\left(\frac{f_{X_{m+1}}}{c_{n}}\right)^{r}\,|\, C_{n, h_{n}(X_{m+1})}=c_{n}\right]\\
&\notag\quad= \frac{1}{D(m,c_{n}; \alpha,\theta,J)}\\
&\notag\quad\quad\times\int_0^1 \int_{0}^{+\infty} (1-t)^r \left( 1-t + \frac{t}{x+1}\right)^{c_{n}-r} \frac{\left(\frac{tx}{x+1}\right)^{m-c_{n}}}{(x+1)^{\theta+\alpha}} f_{B_{\theta+\alpha,1-\alpha}}(t) f_{Z_{\alpha,\theta+\alpha}}(x) \ddr t\ddr x + O\left(\frac{1}{c_{n}}\right).
\end{align}
Now, the double integral on the right-hand side of \eqref{Mom_r} can be rewritten by means of the following change of variable: $y =(1+x)/((1+x)(1-t) + t) \in (1, \frac{1}{1-t})$. In particular, we can write
\begin{align}\label{brackets}
&\int_0^1 \int_{0}^{+\infty} (1-t)^r \left( 1-t + \frac{t}{x+1}\right)^{c_{n}-r} \frac{\left(\frac{tx}{x+1}\right)^{m-c_{n}}}{(x+1)^{\theta+\alpha}} f_{B_{\theta+\alpha,1-\alpha}}(t) f_{Z_{\alpha,\theta+\alpha}}(x) \ddr t\ddr x\\
&\notag\quad= \int_0^1 (1-t)^r\\
&\notag\quad\quad\times \left[\int_{1}^{\frac{1}{1-t}} y^{r-c_{n}} \left(\frac{y-1}{y}\right)^{m-c_{n}} \left(\frac{1 - y(1-t)}{yt}\right)^{\theta+\alpha}
 \frac{tf_{Z_{\alpha,\theta+\alpha}}\left(\frac{y-1}{1 - y(1-t)}\right)}{[1 - y(1-t)]^2} \ddr y \right]f_{B_{\theta+\alpha,1-\alpha}}(t) \ddr t.
\end{align}
We develop a large $m$ asymptotic analysis of \eqref{brackets}, as well as of $D(m,c_{n}; \alpha,\theta,J)$, under the large $m$ asymptotic regime $c_{n}=\lambda m$. We start from the term $D(m,c_{n}; \alpha,\theta,J)$, which we rewrite as
\begin{equation} \label{Den}
D(m,c_{n}; \alpha,\theta,J) = \int_{0}^{+\infty} \left[\frac{x^{\beta}}{x+1}\right]^m \varphi(x)\ddr x
\end{equation}
where $\beta := 1-\lambda$ and  $\varphi(x) := f_{W_{\alpha,\theta}}(x)/(1+x)^{\theta+1}$. The function $\psi : x \mapsto x^{\beta}/(x+1)$ has a unique maximum point $\overline{x} := \frac{\beta}{1-\beta} =\frac{1-\lambda}{\lambda}$. Moreover, straightforward computations show that
\begin{displaymath}
\psi'(x)= \frac{\beta x^{\beta-1} - (1-\beta)x^{\beta}}{(1+x)^2} \\
\end{displaymath}
and
\begin{displaymath}
\psi''(x)= \frac{-\beta(1-\beta)x^{\beta-2} -2\beta(2-\beta)x^{\beta-1}+(1-\beta)(2-\beta)x^{\beta}}{(1+x)^3}\ . 
\end{displaymath}
Then, $\psi''(\overline{x}) = - \overline{x}^{\beta-1}/(1+\overline{x})^3$ and the Laplace method leads to the following large $m$ behaviour
\begin{equation} \label{Den_Laplace}
D(m,c_{n}; \alpha,\theta,J) \sim \frac{1}{\sqrt m} \varphi(\overline{x}) \left[ \frac{\overline{x}^{\beta}}{1+\overline{x}} \right]^{m + \frac 12} \sqrt{\frac{2\pi(1+\overline{x})^3}{\overline{x}^{\beta-1}}}\ .
\end{equation}
We consider \eqref{brackets}, i.e. the integral within brackets on the right-hand side of \eqref{brackets}, which we rewrite as
\begin{align*}
& \int_{1}^{\frac{1}{1-t}} y^{r} \left[\frac{(y-1)^{\beta}}{y}\right]^{m} \left(\frac{1 - y(1-t)}{yt}\right)^{\theta+\alpha}  \frac{tf_{Z_{\alpha,\theta+\alpha}}\left(\frac{y-1}{1 - y(1-t)}\right)}{[1 - y(1-t)]^2} \ddr y \\
&\quad = \int_{0}^{\frac{t}{1-t}} (1+x)^r \left[\frac{x^{\beta}}{1+x}\right]^m \varphi_t(x) \ddr x,
\end{align*}
where
\begin{displaymath}
\varphi_t(x) := \left(\frac{1 - (x+1)(1-t)}{(x+1)t}\right)^{\theta+\alpha}  \frac{tf_{Z_{\alpha,\theta+\alpha}}\left(\frac{x}{1 - (x+1)(1-t)}\right)}{[1 - (x+1)(1-t)]^2}\ .
\end{displaymath}
To apply the Laplace method, note that $(0,1) \ni t \mapsto t/(1-t) \in (0,+\infty)$ is a strictly monotonically increasing function. Thus, $\beta < t$ entails $\overline{x} := \frac{\beta}{1-\beta} < \frac{t}{1-t}$ and, for such $t$, it holds
\begin{equation} \label{main_Laplace}
\int_{0}^{\frac{t}{1-t}} (1+x)^r \left[\frac{x^{\beta}}{1+x}\right]^m \varphi_t(x) \ddr x \sim \frac{(1+\overline{x})^r}{\sqrt m} \varphi_t(\overline{x}) 
\left[ \frac{\overline{x}^{\beta}}{1+\overline{x}} \right]^{m + \frac 12} \sqrt{\frac{2\pi(1+\overline{x})^3}{\overline{x}^{\beta-1}}}
\end{equation}
for large $m$. On the other hand, $\beta > t$ entails $\overline{x} := \frac{\beta}{1-\beta} > \frac{t}{1-t}$ and, for such $t$, there holds a similar large $m$ asymptotic expansion. Now, by exploiting the fact that $\psi : x \mapsto x^{\beta}/(x+1)$ is a strictly monotonically increasing function for $x \in (0,t/(1-t))$, then we can write the following
\begin{displaymath}
\int_{0}^{\frac{t}{1-t}} (1+x)^r \left[\frac{x^{\beta}}{1+x}\right]^m \varphi_t(x) \ddr x \sim \frac{1}{m^{\theta+2\alpha}} \left[\frac{\left(\frac{t}{1-t}\right)^{\beta}}{1+\frac{t}{1-t}}\right]^m \rho(t) =
\frac{1}{m^{\theta+2\alpha}} [t^{\beta}(1-t)^{1-\beta}]^m \rho(t)
\end{displaymath}
for large $m$, where $\rho$ is a suitable function independent of $m$. Accordingly, we can write that
\begin{equation} \label{negligible_part}
\frac{1}{m^{\theta+2\alpha}} \int_0^{\beta} (1-t)^r [t^{\beta}(1-t)^{1-\beta}]^m \rho(t) f_{B_{\theta+\alpha,1-\alpha}}(t) \ddr t \sim \frac{C(\beta)}{m^{1+\theta+2\alpha}} \left[ \frac{\overline{x}^{\beta}}{1+\overline{x}} \right]^m
\end{equation}
as $m \to +\infty$. Then, starting from Equation \eqref{Mom_r} and then gathering \eqref{main_Laplace} and \eqref{negligible_part} we can write
\begin{align}\label{almost_final}
&\E\left[\left(\frac{F_{X_{m+1}}}{c_{n}}\right)^{r}\right]\\
&\notag\quad\sim \frac{1}{D(m,c_{n}; \alpha,\theta,J)} \Bigg\{\frac{1}{m^{\theta+2\alpha}} \int_0^{\beta} (1-t)^r [t^{\beta}(1-t)^{1-\beta}]^m \rho(t) f_T(t) \ddr t \\
&\notag\quad\quad+ \frac{(1+\overline{x})^r}{\sqrt m} \sqrt{\frac{2\pi(1+\overline{x})^3}{\overline{x}^{\beta-1}}} \left[\frac{\overline{x}^{\beta}}{1+\overline{x}} \right]^{m + \frac 12} \int_{\beta}^1 (1-t)^r
\varphi_t(\overline{x}) f_T(t) \ddr t \Bigg\}.
\end{align}
as $m \to +\infty$. According to \eqref{Den_Laplace} the first term in the right-hand side of \eqref{almost_final} is  negligible, and hence
\begin{align*}
&\E\left[\left(\frac{F_{X_{m+1}}}{c_{n}}\right)^{r}\right]\\
&\notag\quad\sim \frac{(1+\overline{x})^r}{\varphi(\overline{x})} \int_{\beta}^1(1-t)^r \varphi_t(\overline{x}) f_{B_{\theta+\alpha,1-\alpha}}(t) \ddr t\\
&\notag\quad = \left(\frac{m}{c_{n}}\right)^r \frac{1}{\varphi(\overline{x})}
\int_0^{\lambda} \tau^r \varphi_{1-\tau}(\overline{x}) f_{B_{\theta+\alpha,1-\alpha}}(1-\tau) \ddr \tau 
\end{align*}
as $m \to +\infty$, which, because of the large $m$ asymptotic regime $c_{n}=\lambda m$, completes the proof.

\section{Proof of Equation \eqref{eq:posterior_py_compl}}\label{sec5app}

Because of the independence assumption of $\mathcal{H}$, i.e. the hash functions $h_{n}$'s are i.i.d. according to the strong universal family $\mathcal{H}$, and by an application of Bayes theorem, we can write
\begin{align}\label{numer_many}
&\notag\text{Pr}[f_{X_{m+1}} = l\,|\, \{C_{n, h_{n}(X_{m+1})}\}_{n\in[N]}= \{c_{n}\}_{n\in[N]}] \\
&\notag\quad=\frac{1}{\text{Pr}[\{C_{n, h_{n}(X_{n+1})}\}_{n\in[N]}=\{c_{n}\}_{n\in[N]}]}\text{Pr}[f_{X_{m+1}}=l]\prod_{n=1}^{N}\text{Pr}[C_{n,h_{n}(X_{m+1})}=c_{n}\,|\,f_{X_{m+1}}=l]\\
&\notag\quad=\frac{1}{\text{Pr}[\{C_{n, h_{n}(X_{m+1})}\}_{n\in[N]}=\{c_{n}\}_{n\in[N]}]}\text{Pr}[f_{X_{m+1}}=l]\prod_{n=1}^{N}\frac{\text{Pr}[C_{n,h_{n}(X_{m+1})}=c_{n},\,f_{X_{m+1}}=l]}{\text{Pr}[f_{X_{m+1}}=l]}\\
&\notag\quad=\frac{1}{\text{Pr}[\{C_{n, h_{n}(X_{m+1})}\}_{n\in[N]}=\{c_{n}\}_{n\in[N]}]}(\text{Pr}[f_{X_{m+1}}=l])^{1-N}\\
&\notag\quad\quad\times\prod_{n=1}^{N}\text{Pr}[C_{n,h_{n}(X_{m+1})}=c_{n}]\text{Pr}[f_{X_{m+1}}=l\,|\,C_{n,h_{n}(X_{m+1})}=c_{n}]\\
&\quad=(\text{Pr}[f_{X_{m+1}}=l])^{1-N}\prod_{n=1}^{N}\text{Pr}[f_{X_{m+1}}=l\,|\,C_{n,h_{n}(X_{m+1})}=c_{n}]
\end{align}
for $l=0,1,\ldots,\min\{c_{1},\ldots,c_{N}\}$, where $\text{Pr}[f_{X_{m+1}}=l\,|\,C_{n,h_{n}(X_{m+1})}=c_{n}]$ is precisely the posterior distribution computed in Theorem \ref{teo_pyp} with respect to the hash function $h_{n}$, whereas
\begin{align*}
\text{Pr}[f_{X_{m+1}}=l]&=\sum_{\mathbf{m}\in\mathcal{M}_{m,k}}\text{Pr}[\mathbf{M}_{m}=\mathbf{m}]\text{Pr}[X_{m+1} \in \mathbf{v}_{l}\,|\, X_{1:m}]\\
&=\sum_{\mathbf{m}\in\mathcal{M}_{m,k}}\text{Pr}[\mathbf{M}_{m}=\mathbf{m}]\text{Pr}[X_{m+1} \in \mathbf{v}_{l}\,|\, X_{1:m}],
\end{align*}
where $\text{Pr}[\mathbf{M}_{m}=\mathbf{m}]$ is in Equation \eqref{eq_ewe_py} and $\text{Pr}[X_{m+1} \in \mathbf{v}_{l}\,|\, X_{1:m}]$ is in Equation \eqref{eq:pred_seen_py}. That is,
\begin{align*}
\text{Pr}[f_{X_{m+1}}=l]=\sum_{\mathbf{m}\in\mathcal{M}_{m,k}}m!\frac{\left(\frac{\theta}{\alpha}\right)_{(k)}}{(\theta)_{(m)}}\prod_{i=1}^{m}\left(\frac{\alpha(1-\alpha)_{(i-1)}}{i!}\right)^{m_{i}}\frac{1}{m_{i}!}
\begin{cases} 
 \frac{\theta+k\alpha}{\theta+m}&\mbox{ if } l=0\\[0.4cm] 
\frac{m_{l}(l-\alpha)}{\theta+m}&\mbox{ if } l\geq1.
\end{cases}.
\end{align*}
For $l=0$
\begin{align*}
\text{Pr}[f_{X_{m+1}}=0]&=\sum_{\mathbf{m}\in\mathcal{M}_{m,k}}m!\frac{\left(\frac{\theta}{\alpha}\right)_{(k)}}{(\theta)_{(m)}}\prod_{i=1}^{m}\left(\frac{\alpha(1-\alpha)_{(i-1)}}{i!}\right)^{m_{i}}\frac{1}{m_{i}!}\frac{\theta+k\alpha}{\theta+m}\\
&=\frac{\theta}{\theta+m}+\frac{\alpha}{\theta+m}\E[K_{m}]\\
&=\frac{\theta}{\theta+m}+\frac{\alpha}{\theta+m}\left(\frac{(\theta+\alpha)_{(m)}}{\alpha(\theta+1)_{(m-1)}}-\frac{\theta}{\alpha}\right),
\end{align*}
where the last equality follows from \citet[Equation 3.13]{Pit(06)}. Accordingly, we can write that
\begin{displaymath}
\text{Pr}[f_{X_{m+1}}=0]=\frac{(\theta+\alpha)_{(m)}}{(\theta+1)_{(m)}}
\end{displaymath}
For $l\geq1$
\begin{align*}
\text{Pr}[f_{X_{m+1}}=l]&=\sum_{\mathbf{m}\in\mathcal{M}_{m,k}}m!\frac{\left(\frac{\theta}{\alpha}\right)_{(k)}}{(\theta)_{(m)}}\prod_{i=1}^{m}\left(\frac{\alpha(1-\alpha)_{(i-1)}}{i!}\right)^{m_{i}}\frac{1}{m_{i}!}\frac{m_{l}(l-\alpha)}{\theta+m}\\
&=\frac{l-\alpha}{\theta+m}\E[M_{l,m}]\\
&=\frac{l-\alpha}{\theta+m}\frac{(1-\alpha)_{(l-1)}}{l!}(m)_{[l]}\frac{(\theta+\alpha)_{(m-l)}}{(\theta+1)_{(m-1)}}
\end{align*}
where the last equality follows from \citet[Proposition 1]{Fav(13)}. Accordingly, for $l=1,\ldots,m$,
\begin{equation}\label{denomin_many}
\text{Pr}[f_{X_{m+1}}=l]=\frac{(1-\alpha)_{(l)}}{l!}(m)_{[l]}\frac{(\theta+\alpha)_{(m-l)}}{(\theta+1)_{(m)}}.
\end{equation}
Equation \eqref{eq:posterior_py_compl} follows by combining the distribution \eqref{numer_many} with \eqref{denomin_many}. This completes the proof.

\section{CMS for range queries under DP priors}\label{secrange}

We assume that the stream $x_{1:m}$ is modeled as a random sample $X_{1:m}$ from an unknown discrete distribution $P$, which is endowed with a DP prior, i.e. $P \sim\,\text{DP}(\theta;\nu)$. Let $h_{1},\ldots,h_{N}$ be a collection of random hash functions that are i.i.d. from the strong universal family $\mathcal{H}$, and assume that $h_{1},\ldots,h_{N}$ are independent of $X_{1:m}$ for any $m\geq1$; in particular, by de Finetti's representation theorem, $h_{1},\ldots,h_{N}$ are independent of $P \sim\,\text{DP}(\theta;\nu)$. Under this BNP framework, a $s$-range query induces the posterior distribution of the frequencies $(f_{x_{m+1}},\ldots,f_{x_{m+s}})$ given the hashed frequencies $\{(C_{n,h_n(v_{1})},\ldots,C_{n,h_n(v_{s})})\}_{n\in[N]}$, for arbitrary $\{x_{m+1},\ldots,x_{m+s}\}\in\mathcal{V}$. This posterior distribution, in turn, induces the posterior distribution of the $s$-range query $\bar{f}_{s}$ given $\{(C_{n,h_n(v_{1})},\ldots,C_{n,h_n(v_{s})})\}_{n\in[N]}$. CMS-DP estimates of $\bar{f}_{s}$ are obtained as functionals of the posterior distribution of $\bar{f}_{s}$ given $\{(C_{n,h_n(v_{1})},\ldots,C_{n,h_n(v_{s})})\}_{n\in[N]}$. To compute the posterior distribution of $(f_{x_{m+1}},\ldots,f_{x_{m+s}})$ given $\{(C_{n,h_n(v_{1})},\ldots,C_{n,h_n(v_{s})})\}_{n\in[N]}$, it is natural to consider $s$ additional random samples $(X_{m+1},\ldots,X_{m+s})$. In particular, for any  $r=1,\ldots,s$ let $f_{X_{m+r}}$ be the frequency of $X_{m+r}$ in $X_{1:m}$, i.e.,
\begin{displaymath}
f_{X_{m+r}}=\sum_{i=1}^{m}\mathbbm{1}_{\{X_{i}\}}(X_{m+r})
\end{displaymath}
and let $C_{n,h_{n}(X_{m+r})}$ be the hashed frequency of all $X_{i}$'s, for $i=1,\ldots,m$, such that $h_{n}(X_{i})=h_{n}(X_{m+r})$, i.e.,
\begin{displaymath}
C_{n,h_{n}(X_{m+r})}=\sum_{i=1}^{m}\mathbbm{1}_{h_{n}(X_{i})}(h(X_{m+r})).
\end{displaymath}
Now, let $\mathbf{X}_{s}=(X_{m+1},\ldots,X_{m+s})$ and for $n\in[N]$ let $\mathbf{f}_{\mathbf{X}_{s}}=(f_{X_{m+1}},\ldots,f_{X_{m+s}})$. For $n\in[N]$ let $\mathbf{C}_{n,h_{n}(\mathbf{X}_{s})}=(C_{n, h_{n}(X_{m+1})},\ldots,C_{n, h_{n}(X_{m+s})}$. For each $h_{n}$ we are interested in the posterior distribution
\begin{align}\label{eq:maintoprove2}
&\text{Pr}\left[\mathbf{f}_{\mathbf{X}_{s}}=\mathbf{l}_{s}\,|\,\mathbf{C}_{n,h_{n}(\mathbf{X}_{s})}=\mathbf{c}_{n}\right]=\frac{\text{Pr}[\mathbf{f}_{\mathbf{X}_{s}}=\mathbf{l}_{s},\mathbf{C}_{n,h_{n}(\mathbf{X}_{s})}=\mathbf{c}_{n}]}{\text{Pr}[\mathbf{C}_{n,h_{n}(\mathbf{X}_{s})}=\mathbf{c}_{n}]}.
\end{align}
for $\mathbf{l}_{s}\in\{0,1,\ldots,m\}^{s}$. For the collection of hash functions $h_{1},\ldots,h_{N}$, the posterior distribution of $\mathbf{f}_{\mathbf{X}_{s}}$ given $\{\mathbf{C}_{n,h_{n}(\mathbf{X}_{s})}\}_{n\in[N]}$ follows from the posterior distribution \eqref{eq:maintoprove2} by the assumption that the $h_{n}$'s are i.i.d. according to the strong universal family $\mathcal{H}$, and Bayes theorem. 

Hereafter we show that the ``Bayesian" proof of Section \ref{sec2} can be readily extended to the computation of the posterior distribution \eqref{eq:maintoprove2}.  We outline this extension for any range $s\geq1$, and then we present an explicit example for $s=2$. To simplify the notation, we remove the subscript $n$ from $h_{n}$ and $\mathbf{c}_n$. Then, we are interested in computing the posterior distribution
\begin{align}\label{eq:maintoprove3}
&\text{Pr}\left[\mathbf{f}_{\mathbf{X}_{s}}=\mathbf{l}_{s}\,|\,\mathbf{C}_{h(\mathbf{X}_{s})}=\mathbf{c}\right]=\frac{\text{Pr}[\mathbf{f}_{\mathbf{X}_{s}}=\mathbf{l}_{s},\mathbf{C}_{h(\mathbf{X}_{s})}=\mathbf{c}]}{\text{Pr}[\mathbf{C}_{h(\mathbf{X}_{s})}=\mathbf{c}]}.
\end{align}
For $s=1$ the posterior distribution \eqref{eq:maintoprove3} reduces to \eqref{eq:maintoprove}. The independence between $h_{n}$ and $X_{1:m}$ allows us to invoke the ``freezing lemma'' \citep[Lemma 4.1]{Bal(17)}, according to which we can treat $h_{n}$ as it was fixed, i.e. non-random. We analyze the posterior distribution \eqref{eq:maintoprove3} starting from its denominator. In particular, the denominator of \eqref{eq:maintoprove3} can be written as follows
\begin{align*}
\text{Pr}[\mathbf{C}_{h(\mathbf{X}_{s})}=\mathbf{c}] &= \sum_{(j_1, \dots, j_s) \in [J]^s} \text{Pr}[\mathbf{C}_{h(\mathbf{X}_{s})}=\mathbf{c}, 
h(X_{m+1})=j_1, \dots, h(X_{m+s})=j_s] \\
& = \sum_{(j_1, \dots, j_s) \in [J]^s} \text{Pr}\Bigg[\sum_{i=1}^m \mathbbm{1}_{h(X_{i})}(j_1) = c_{1}, \dots, \sum_{i=1}^m \mathbbm{1}_{h(X_{i})}(j_s) = c_{s}, \\
& \qquad\qquad\qquad\qquad\qquad\qquad\qquad\qquad\quad\quad h(X_{m+1})=j_1, \dots, h(X_{m+s})=j_s \Bigg].
\end{align*}
To evaluate
\begin{equation}\label{eq:prob_range1}
\text{Pr}\Bigg[\sum_{i=1}^m \mathbbm{1}_{h(X_{i})}(j_1) = c_{1}, \dots, \sum_{i=1}^m \mathbbm{1}_{h(X_{i})}(j_s) = c_{s}, h(X_{m+1})=j_1, \dots, h(X_{m+s})=j_s \Bigg],
\end{equation}
we split the sum over $[J]^{s}$ and we organize the summands as follows. First, we introduce a variable $k$ which counts how many distinct object there are in each vector $(j_1, \dots, j_s)$, so that $k \in \{1, 2, \dots, \min\{s,J\}\}$. Second, we consider the vector $(r_1, \dots, r_k)$ of frequencies of the distinct $k$ objects. Third, we consider the vector $(j_1^{\ast}, \dots, j_k^{\ast})$ of distinct objects with $\{j_1^{\ast}, \dots, j_k^{\ast}\} \subseteq \{1,\dots, J\}$. Then, we evaluate the probability \eqref{eq:prob_range1} in the distinguishing case that
\begin{displaymath}
\left\{\begin{array}{l}
j_1 = \dots = j_{r_1} =: j_1^{\ast}\\
j_{r_1+1} = \dots = j_{r_1+r_2} =: j_2^{\ast} \\
\dots \\ 
j_{r_1+\dots+r_{k-1}+1} = \dots = j_{r_1+\dots+r_k} =: j_k^{\ast}
\end{array}\right.
\end{displaymath}
such that the probability \eqref{eq:prob_range1} of interest is different from zero if and only if the following holds true
\begin{displaymath}
\left\{\begin{array}{l}
c_{1} = \dots = c_{r_1} =: c_{1}^{\ast}\\
c_{r_1+1} = \dots = c_{r_1+r_2} =: c_{2}^{\ast} \\
\dots \\ 
c_{r_1+\dots+r_{k-1}+1} = \dots = c_{r_1+\dots+r_k} =: c_{k}^{\ast}.
\end{array}\right.
\end{displaymath}
That is,
\begin{align*}
\text{Pr}&\Bigg[\sum_{i=1}^m \mathbbm{1}_{h(X_{i})}(j_1^{\ast}) = c_{1}^{\ast}, \dots, \sum_{i=1}^m \mathbbm{1}_{h(X_{i})}(j_k^{\ast}) = c_{k}^{\ast}, h(X_{m+1})=\dots=h(X_{m+r_1})=j_1^{\ast},\ldots \\
&\quad\quad\quad\quad\quad\quad\quad\quad\quad\quad\quad\quad\quad\quad \dots, h(X_{m+r_1+\dots+r_{k-1}+1})=\dots=h(X_{m+r_1+\dots+r_k})=j_k^{\ast}\Bigg].
\end{align*}
Now, we set $B_r^{\ast} := \{x \in \mathcal V\text{ : } h(x) = j_r^{\ast}\}$ for any $r \in \{1, \dots, k\}$ and we set $B_{k+1}^{\ast} = \left(\cup_{r=1}^k B_r^{\ast} \right)^C$. Thus, $\{B_1^{\ast}, \dots, B_{k+1}^{\ast}\}$ is a finite partition of $\mathcal V$. If $k = J$, then $B_{k+1}^{\ast} = \emptyset$ and in such case we intend that $\{B_1^{\ast}, \dots, B_{k+1}^{\ast}\}$ is replaced by $\{B_1^{\ast}, \dots, B_k^{\ast}\}$. Accordingly, we can write the identity
\begin{align*}
\text{Pr}&\Bigg[\sum_{i=1}^m \mathbbm{1}_{h(X_{i})}(j_1^{\ast}) = c_{1}^{\ast}, \dots, \sum_{i=1}^m \mathbbm{1}_{h(X_{i})}(j_k^{\ast}) = c_{k}^{\ast}, h(X_{m+1})=\dots=h(X_{m+r_1})=j_1^{\ast},\ldots \\
&\quad\quad\quad\quad\quad\quad\quad\quad\quad\quad\quad\quad\quad\quad \dots, h(X_{m+r_1+\dots+r_{k-1}+1})=\dots=h(X_{m+r_1+\dots+r_k})=j_k^{\ast}\Bigg] \\
&\notag = \binom{m}{c_{1}^{\ast}, \dots, c_{k}^{\ast}} \int_{\Delta_k} \left( \prod_{i=1}^k p_i^{c_{i}^{\ast}+r_i} \right)(1 - p_1 - \dots - p_k)^{m - \sum_{i=1}^k c_{i}^{\ast}} 
\mu_{B_1^{\ast}, \dots, B_{k+1}^{\ast}}(\ddr p_1 \dots \ddr p_k) 
\end{align*}
where $\mu_{B_1^{\ast}, \dots, B_{k+1}^{\ast}}$ is the distribution of $(P(B_1^{\ast}), \dots, P(B_k^{\ast}))$ which, by the finite-dimensional projective property of the DP, is a Dirichlet distribution with parameter $(\theta/J,\ldots,\theta/J)$ on $\Delta_k$. If $k < J$
\begin{align*}
\text{Pr}&\Bigg[\sum_{i=1}^m \mathbbm{1}_{h(X_{i})}(j_1^{\ast}) = c_{1}^{\ast}, \dots, \sum_{i=1}^m \mathbbm{1}_{h(X_{i})}(j_k^{\ast}) = c_{k}^{\ast}, h(X_{m+1})=\dots=h(X_{m+r_1})=j_1^{\ast},\ldots \\
&\quad\quad\quad\quad\quad\quad\quad\quad\quad\quad\quad\quad\quad\quad \dots, h(X_{m+r_1+\dots+r_{k-1}+1})=\dots=h(X_{m+r_1+\dots+r_k})=j_k^{\ast}\Bigg] \\
&\quad=\frac{\Gamma(\theta)}{[\Gamma(\frac{\theta}{J})]^k \Gamma((J-k)\frac{\theta}{J})} \frac{\left[\prod_{i=1}^k \Gamma(\frac{\theta}{J}+ c_{i}^{\ast}+r_i)\right] \Gamma((J-k)\frac{\theta}{J} + m - \sum_{i=1}^k c_{i}^{\ast})}{\Gamma(\theta+m+s)},
\end{align*}
and if $k = J$
\begin{align*}
\text{Pr}&\Bigg[\sum_{i=1}^m \mathbbm{1}_{h(X_{i})}(j_1^{\ast}) = c_{1}^{\ast}, \dots, \sum_{i=1}^m \mathbbm{1}_{h(X_{i})}(j_k^{\ast}) = c_{k}^{\ast}, h(X_{m+1})=\dots=h(X_{m+r_1})=j_1^{\ast},\ldots \\
&\quad\quad\quad\quad\quad\quad\quad\quad\quad\quad\quad\quad\quad\quad \dots, h(X_{m+r_1+\dots+r_{k-1}+1})=\dots=h(X_{m+r_1+\dots+r_k})=j_k^{\ast}\Bigg] \\
&\quad=\frac{\Gamma(\theta)}{(\Gamma(\frac{\theta}{J}))^k} \frac{\prod_{i=1}^k \Gamma(\frac{\theta}{J} + c_{i}^{\ast}+r_i)}{\Gamma(\theta+m+s)}.
\end{align*}
Upon denoting by $I_k(c_{n,1}^{\ast}, \dots, c_{n,k}^{\ast}; r_1, \dots, r_k)$ the right expression of the integral, we conclude that
\begin{align}\label{eq:prob_range2}
&\text{Pr}[\mathbf{C}_{h(\mathbf{X}_{s})}=\mathbf{c}] \\
&\notag= \sum_{k=1}^{ \min\{s,J\}} \frac{J!}{(J-k)!}\\
&\notag\quad\times \sum_{(\pi_1, \dots, \pi_k) \in \Pi(s,k)} \Delta(\pi_1, \dots, \pi_k; c_{1}, \dots, c_{s}) 
\binom{m}{c_{1}^{\ast}, \dots, c_{k}^{\ast}}I_k(c_{1}^{\ast}, \dots, c_{k}^{\ast}; |\pi_1|, \dots, |\pi_k|),
\end{align}
where: i) $\Pi(s,k)$ denotes the set of all possible partitions of the set $\{1, \dots, s\}$ into $k$ disjoint subsets $\pi_1, \dots, \pi_k$; $|\pi_i|$ stands for the cardinality of the subset $\pi_i$; ii) $\Delta(\pi_1, \dots, \pi_k; c_{1}, \dots, c_{s})$ is either $0$ or $1$ with the proviso that it equals 1 if and only if, for all $z \in \{1, \dots, k\}$ for which $|\pi_z| \geq 2$, all the integers $c_{i}$ with $i \in \pi_z$ are equal; for any $i \in \{1, \dots, k\}$, $c_{i}$ represents the common integer associated to $\pi_i$. Formula \eqref{eq:prob_range2} simplifies remarkably for small values of $s$. For instance, 
\begin{itemize}
\item[i)] for $s=1$
\begin{displaymath}
\text{Pr}[C_{h}(X_{m+1})=c_{1}]=J \binom{m}{c_{1}} I_1(c_1; 1);
\end{displaymath}
\item[ii)] for $s=2$
\begin{align}\label{Den2}
&\text{Pr}[C_{h}(X_{m+1})=c_{1},C_{h}(X_{m+2})=c_{2}]\\
&\notag\quad=J \mathbbm{1}\{c_{1} = c_{2}\} \binom{m}{c_{1}} I_1(c_1; 2) + J(J-1) \binom{m}{c_{1}, c_{2}} I_2(c_{1}, c_{2}; 1,1).
\end{align}
\end{itemize}
We conclude by studying the numerator in \eqref{eq:maintoprove3}. This expression is determined by the complete knowledge of the joint distribution of $(X_1, \dots, X_{n+s})$. As above, we can start by writing
\begin{align*}
& \text{Pr}[\mathbf{f}_{\mathbf{X}_{s}}=\mathbf{l}_{s},\mathbf{C}_{h(\mathbf{X}_{s})}=\mathbf{c}]\\
&= \sum_{k=1}^s  \sum_{(\pi_1, \dots, \pi_k) \in \Pi(s,k)} \Delta(\pi_1, \dots, \pi_k; l_1, \dots, l_s)\binom{n}{l_1^{\ast}, \dots, l_k^{\ast}}\\
&\quad \times \text{Pr}\left[ B(m; l_1^{\ast}, \dots, l_k^{\ast}; \pi_1, \dots, \pi_k) \cap \left\{\sum_{i=1}^m \mathbbm{1}_{h(X_{i})}(j_1) = c_{1}, \dots, \sum_{i=1}^m \mathbbm{1}_{h(X_{i})}(j_s) = c_{s}\right\} \right]
\end{align*}
where the event $B(m; l_1^{\ast}, \dots, l_k^{\ast})$ is characterized by the relations among random variables $X_{m+r}$'s
\begin{align*}
X_1 = \dots = X_{l_1^{\ast}} = X_{m+r} &\quad \text{for\ all}\ r \in \pi_1 \\ 
X_{l_1^{\ast}+1} = \dots = X_{l_1^{\ast}+ l_2^{\ast}} = X_{m+r} &\quad \text{for\ all}\ r \in \pi_2 \\
\dots & \dots  \\
X_{l_1^{\ast}+\dots+l_{k-1}^{\ast}+1} = \dots = X_{l_1^{\ast}+ \dots+l_{k}^{\ast}} = X_{m+r} &\quad \text{for\ all}\ r \in \pi_k \\
X_{n+r_1} \neq X_{n+r_2}  &\quad \text{for\ all}\ r_1 \in \pi_a, r_2 \in \pi_b\quad \text{for\ all}\ a \neq b \\
\{X_{l_1^{\ast}+ \dots+l_{k}^{\ast}+1}, \dots X_m\} \cap \{X_{m+1}, \dots, X_{m+s}\} = \emptyset & \ . 
\end{align*}
The numerator of \eqref{eq:maintoprove3} can be treated as the denominator of \eqref{eq:maintoprove3}, namely by exploiting the double partition structure induced by the above relations on the random variables $X_i$'s and $h(X_i)$'s. We observe that the combination of this two partition structures proves particularly cumbersome to be written for general $s\geq1$. For this reason, further manipulations of the posterior distribution \eqref{eq:maintoprove3} will be deferred to the proof the next theorem, where we assume $s=2$.

\begin{theorem}\label{teo_dir_ran}
For $m\geq1$, let $x_{1:m}$ be a stream of tokens that are modeled as a random sample $X_{1:m}$ from $P\sim\text{DP}(\theta;\nu)$, and let $(X_{m+1},X_{m+2})$ be a pair of additional random samples from $P$. Moreover, let $h_{n}$ be a random hash function distributed as the strong universal family $\mathcal{H}$, and let $h_{n}$ be independent of $X_{1:m}$ for any $m\geq1$, that is $h_{n}$ is independent of $P$. Then
\begin{align*}
&\text{Pr}[f_{X_{m+1}} = l_{1},\,f_{X_{m+2}} = l_{2}\,|\, C_{n, h_{n}(X_{m+1})}=c_{n,1},C_{n,h_{n}(X_{m+2})}=c_{n,2}]\\
&\notag\quad= \frac{\mathrm{Num}(l_1, l_2, c_{n,1}, c_{n,2})}{\mathrm{Den}(c_{n,1}, c_{n,2})}\quad l_{1},l_{2}\geq0
\end{align*}
with
\begin{itemize}
\item[i)]
\begin{align*}
\mathrm{Den}(c_{n,1}, c_{n,2})&=J\mathbbm{1}\{c_{n,1}=c_{n,2}=c\}\frac{(\frac{\theta}{J})_{(c+2)}(\theta-\frac{\theta}{J})_{(m-c)}}{c!(m-c)!}\\
&\quad+J(J-1)\frac{(\frac{\theta}{J})_{(c_{n,1}+1)}(\frac{\theta}{J})_{(c_{n,2}+1)}(\theta-\frac{2\theta}{J})_{(m-c_{n,1}-c_{n,2})}}{c_{n,1}!c_{n,2}!(m-c_{n,1}-c_{n,2})!};
\end{align*}
\item[ii)]
\begin{align*} 
\mathrm{Num}(l_1, l_2, c_{n,1}, c_{n,2})&= \mathbbm{1}\{l_1 = l_2 =: l, c_{n,1}=c_{n,2}=c\} \frac{\theta(l+1)(\frac{\theta}{J})_{(c-l)}(\theta-\frac{\theta}{J})_{(m-c)}}{(c-l)! (m-c)!}\\
&\quad+ \mathbbm{1}\{c_{n,1}=c_{n,2}=c\}\frac{\theta^2(\frac{\theta}{J})_{(c-l_{1}-l_{2})}(\theta-\frac{\theta}{J})_{(m-c)}}{J (c-l_1-l_2)! (m-c)!}\\
&\quad\quad+ \left(\frac{J-1}{J}\right) \frac{\theta^2(\frac{\theta}{J})_{(c_{n,1}-l_{1})}(\frac{\theta}{J})_{(c_{n,2}-l_{2})}(\theta-\frac{2\theta}{J})_{(m-c_{n,1}-c_{n,2})}}{(c_{n,1}-l_1)! (c_{n,2}-l_2)! (m-c_{n,1}-c_{n,2})!}.
\end{align*}
\end{itemize}
\end{theorem}

\begin{proof}
Following the ``Bayesian" proof for $s\geq1$, we start by expressing the posterior distribution of $(f_{X_{m+1}},f_{X_{m+2}})$ given $C_{n,h_{n}(X_{m+1})}$ and $C_{n,h_{n}(X_{m+2})}$ as a ratio of two probabilities, and then we deal with the numerator and denominator. That is, we write the following expression 
\begin{align}\label{object_range}
&\text{Pr}[f_{X_{m+1}} = l_{1},\,f_{X_{m+1}} = l_{2}\,|\, C_{n, h_{n}(X_{m+1})}=c_{n,1},C_{n,h_{n}(X_{m+2})}=c_{n,2}]\\
&\notag\quad=\frac{\text{Pr}\left[ f_{X_{m+1}} = l_1, f_{X_{m+2}} = l_2, \sum_{i=1}^m \mathbbm{1}_{h_{n}(X_{i})}(X_{m+1}) = c_{n,1}, \sum_{i=1}^m \mathbbm{1}_{h_{n}(X_{i})}(X_{m+2}) = c_{n,2} \right]}{\text{Pr}\left[C_{n, h_{n}(X_{m+1})}=c_{n,1},C_{n,h_{n}(X_{m+2})}=c_{n,2}\right]}
\end{align}
Observe that the denominator of the posterior distribution \eqref{object_range} reduces to \eqref{Den2}. Then, by using the finite-dimensional projective property of the DP, we can write the following expressions
\begin{align*}
&J\mathbbm{1}\{c_{n,1}=c_{n,2}=c\}{m\choose c}I_{1}(c;2)\\
&\quad=J\mathbbm{1}\{c_{n,1}=c_{n,2}=c\}{m\choose c}\\
&\quad\quad\times\int_{0}^{1}p^{c+2}(1-p)^{m-c}\frac{\Gamma(\theta)}{\Gamma(\theta/J)\Gamma(\theta(1-1/J))}p^{\theta/J-1}(1-p)^{\theta(1-1/J)-1}\text{d}p\\
&\quad=J\mathbbm{1}\{c_{n,1}=c_{n,2}=c\}{m\choose c}\frac{\Gamma(\theta)}{\Gamma(\theta/J)\Gamma(\theta(1-1/J))}\frac{\Gamma(\theta/J+c+2)\Gamma(\theta(1-1/J)+m-c)}{\Gamma(\theta+m+2)}
\end{align*}
and
\begin{align*}
&J(J-1){m\choose c_{n,1},c_{n,2}}I_{2}(c_{n,2},c_{n,2};1,1)\\
&\quad=J(J-1){m\choose c_{n,1},c_{n,2}}\\
&\quad\quad\times\int_{\Delta_{2}}p_{1}^{c_{n,1}+1}p_{2}^{c_{n,2}+1}(1-p_{1}-p_{2})^{m-c_{n,1}-c_{n,2}}\\
&\quad\quad\quad\times\frac{\Gamma(\theta)}{\Gamma(\theta/J)\Gamma(\theta/J)\Gamma(\theta(1-2/J))}p_{1}^{\theta/J-1}p_{2}^{\theta/J-1}(1-p_{1}-p_{2})^{\theta(1-2/J)-1}\text{d}p_{1}\text{d}p_{2}\\
&\quad=J(J-1){m\choose c_{n,1},c_{n,2}}\frac{\Gamma(\theta)}{(\Gamma(\theta/J))^{2}\Gamma(\theta(1-2/J))}\\
&\quad\quad\times\frac{\Gamma(\theta/J+c_{n,1}+1)\Gamma(\theta/J+c_{n,2}+1)\Gamma(\theta(1-2/J)+m-c_{n,1}-c_{n,2})}{\Gamma(\theta+m+2)}.
\end{align*}
Then,
\begin{align}\label{object_range_denominator}
&\text{Pr}\left[C_{n, h_{n}(X_{m+1})}=c_{n,1},C_{n,h_{n}(X_{m+2})}=c_{n,2}\right]\\
&\notag\quad=J\mathbbm{1}\{c_{n,1}=c_{n,2}=c\}{m\choose c_{n,1}}\frac{\Gamma(\theta)}{\Gamma(\theta/J)\Gamma(\theta(1-1/J))}\\
&\notag\quad\quad\quad\times\frac{\Gamma(\theta/J+c+1)\Gamma(\theta(1-1/J)+m-c)}{\Gamma(\theta+m+2)}\\
&\notag\quad\quad+J(J-1){m\choose c_{n,1},c_{n,2}}\frac{\Gamma(\theta)}{(\Gamma(\theta/J))^{2}\Gamma(\theta(1-2/J))}\\
&\notag\quad\quad\quad\times\frac{\Gamma(\theta/J+c_{n,1}+1)\Gamma(\theta/J+c_{n,2}+1)\Gamma(\theta(1-2/J)+m-c_{n,1}-c_{n,2})}{\Gamma(\theta+m+2)}.
\end{align}
Now, we focus on the numerator of the posterior distribution \eqref{object_range}, which is rewritten as follows
\begin{align} \label{split_Num2_Dirichlet}
&\text{Pr}\left[ f_{X_{m+1}} = l_1, f_{X_{m+2}} = l_2, \sum_{i=1}^m \mathbbm{1}_{h_{n}(X_{i})}(X_{m+1}) = c_{n,1}, \sum_{i=1}^m \mathbbm{1}_{h_{n}(X_{i})}(X_{m+2}) = c_{n,2} \right]\\
&\quad= \text{Pr}\left[ f_{X_{n+1}} = l_1, f_{X_{m+2}} = l_2, \sum_{i=1}^m \mathbbm{1}_{h_{n}(X_{i})}(X_{m+1}) = c_{n,1}, \sum_{i=1}^m \mathbbm{1}_{h_{n}(X_{i})}(X_{m+2}) = c_{n,2}, X_{m+1} = X_{m+2} \right] \nonumber \\
&\quad\quad+ \text{Pr}\left[ f_{X_{n+1}} = l_1, f_{X_{m+2}} = l_2, \sum_{i=1}^m \mathbbm{1}_{h_{n}(X_{i})}(X_{m+1}) = c_{n,1}, \sum_{i=1}^m \mathbbm{1}_{h_{n}(X_{i})}(X_{m+2}) = c_{n,2}, X_{m+1} \neq X_{m+2} \right].\nonumber 
\end{align}
First, we consider the first term on the right-hand side of the probability \eqref{split_Num2_Dirichlet}. In particular, we write
\begin{align*}
&\text{Pr}\left[ f_{X_{m+1}} = l_1, f_{X_{m+2}} = l_2, \sum_{i=1}^m \mathbbm{1}_{h_{n}(X_{i})}(X_{m+1}) = c_{n,1}, \sum_{i=1}^m \mathbbm{1}_{h_{n}(X_{i})}(X_{m+2}) = c_{n,2}, X_{m+1} = X_{m+2} \right] \\
&\quad= \mathbbm{1}\{l_1 = l_2 =: l, c_{n,1}=c_{n,2}=c\} \binom{m}{l}\\
&\quad\quad\times \text{Pr}\left[ X_1 = \dots, X_l = X_{m+1} = X_{m+2}, \{X_{l+1}, \dots, X_m\} \cap \{X_{m+1}\} = \emptyset,  \sum_{i=1}^m \mathbbm{1}_{h_{n}(X_{i})}(X_{m+1}) = c\right]\\
&\quad= \mathbbm{1}\{l_1 = l_2 =: l, c_{n,1}=c_{n,2}=c\} \binom{m}{l}\\
&\quad\quad\times\text{Pr}\left[X_1 = \dots, X_l = X_{m+1}= X_{m+2}, \{X_{l+1}, \dots, X_m\} \cap \{X_{m+1}\} = \emptyset, \sum_{i=l+1}^m \mathbbm{1}_{h_{n}(X_{i})}(X_{m+1}) = c-l \right]
\end{align*}
which is determined by the distribution of $(X_1, \dots, X_{m+2})$. In view of \citet[Equation 3.5]{San(06)}
\begin{displaymath}
\text{Pr}[X_1 \in C_1, \dots, X_{m+2} \in C_{m+2}] = \sum_{k=1}^{m+2} \frac{\theta^k}{(\theta)_{(m+2)}} \sum_{(\pi_1, \dots, \pi_k) \in \Pi(m+2,k)} \prod_{i=1}^k (|\pi_i| -1)! \
\nu(\cap_{r\in \pi_i} C_r)\ .
\end{displaymath}
We set $D(m,l) := \{X_1 = \dots, X_l = X_{m+1}= X_{m+2}, \{X_{l+1}, \dots, X_m\} \cap \{X_{m+1}\} = \emptyset\}$, and we define $\mu_{(\pi_1, \dots, \pi_k)}$ as the probability measure on $(\mathcal V^{m+2}, \mathscr V^{n+2})$ generated by the following identity
\begin{displaymath}
\nu_{\pi_1, \dots, \pi_k}(C_1 \times \dots \times C_{m+2}) := \prod_{i=1}^k \nu(\cap_{r\in \pi_i} C_r)\ , 
\end{displaymath}
It is clear that such measures attach to $D(m,l)$ a probability value that is either 0 or 1. In particular, $\nu_{\pi_1, \dots, \pi_k}(D(m,l)) = 1$ if and only if one of the $\pi$'s (e.g. $\pi_k$, being these partitions given up to the order) is exactly equal to the set $\{1, \dots, l, m+1, m+2\}$. Accordingly, we write
\begin{align*}
& \text{Pr}\left[D(m,l), \sum_{i=l+1}^m \mathbbm{1}_{h_{n}(X_{i})}(X_{m+1}) = c-l \right] \\
&\quad= \sum_{k=2}^{m-l+1} \frac{\theta^k}{(\theta)_{(m+2) }} \sum_{(\pi_1, \dots, \pi_{k-1}) \in \Pi(m-l,k-1)} (l+1)! \prod_{i=1}^{k-1} (|\pi_i| -1)! 
\nu_{\pi_1, \dots, \pi_k}\left( \sum_{i=l+1}^m \mathbbm{1}_{h_{n}(X_{i})}(X_{m+1})= c-l \right) \\
&\quad= \frac{\theta (\theta)_{(m-l)}}{(\theta)_{(m+2) }} (l+1)! 
\sum_{r=1}^{m-l} \frac{\theta^r}{(\theta)_{(m-l) }} \sum_{(\pi_1, \dots, \pi_r) \in \Pi(m-l,r)} \prod_{i=1}^r (|\pi_i| -1)! \times \\
&\quad\quad\times \left\{\sum_{j=1}^J \nu(\{j\}) 
\nu_{\pi_1, \dots, \pi_r}\left( \sum_{i=l+1}^m \mathbbm{1}_{h_{n}(X_{i})}(j)= c-l\right) \right\} \\
&\quad= \frac{\theta (\theta)_{(m-l) }}{J (\theta)_{(m+2) }} (l+1)! \sum_{r=1}^{m-l} \frac{\theta^r}{(\theta)_{(m-l) }} \sum_{(\pi_1, \dots, \pi_r) \in \Pi(m-l,r)} \prod_{i=1}^r (|\pi_i| -1)! \times \\
&\quad\quad\times \left\{\sum_{j=1}^J  \nu_{\pi_1, \dots, \pi_r}\left( \sum_{i=l+1}^m \mathbbm{1}_{h_{n}(X_{i})}(j)= c-l\right)\right\}  \ .
\end{align*}
Hence,
\begin{align} \label{Num2_Dirichlet_A} 
&\text{Pr}\left[ f_{X_{m+1}} = l_1, f_{X_{m+2}} = l_2, \sum_{i=1}^m \mathbbm{1}_{h_{n}(X_{i})}(X_{m+1}) = c_{n,1}, \sum_{i=1}^m \mathbbm{1}_{h_{n}(X_{i})}(X_{m+2}) = c_{n,2}, X_{m+1} = X_{m+2} \right]\\
&\quad=  \mathbbm{1}\{l_1 = l_2 =: l, c_{n,1}=c_{n,2}=c\}
\frac{m!}{(c-l)! (m-c)!} \frac{\theta(l+1)}{\Gamma(\theta+m+2)} \times  \nonumber \\
&\quad\quad \times \frac{\Gamma(\theta)}{\Gamma(\theta/J) \Gamma(\theta(1 - 1/J))} \Gamma(\theta/J + c - l) \Gamma(\theta(1 - 1/J) + m - c)\nonumber.  
\end{align}
Now, we consider the second term on the right-hand side of the probability \eqref{split_Num2_Dirichlet}. In particular, we write
\begin{align*}
&\text{Pr}\left[ f_{X_{m+1}} = l_1, f_{X_{m+2}} = l_2, \sum_{i=1}^m \mathbbm{1}_{h_{n}(X_{i})}(X_{m+1}) = c_{n,1}, \sum_{i=1}^m \mathbbm{1}_{h_{n}(X_{i})}(X_{m+2}) = c_{n,2}, X_{m+1} \neq X_{m+2}\right] \\
&\quad= \binom{m}{l_1, l_2} \text{Pr}\Bigg[X_1 = \dots, X_{l_1} = X_{m+1}, X_{l_1+1} = \dots, X_{l_1+l_2} = X_{m+2}, X_{m+1} \neq X_{m+2}, \\
&\quad\quad\quad\quad\quad\quad \{X_{l_1+l_2+1}, \dots, X_m\} \cap \{X_{m+1}, X_{m+2}\} = \emptyset, \sum_{i=1}^m \mathbbm{1}_{h_{n}(X_{i})}(X_{m+1}) = c_{n,1}, \sum_{i=1}^m \mathbbm{1}_{h_{n}(X_{i})}(X_{m+2}) = c_{n,2}\Bigg]\\
&\quad=\binom{m}{l_1, l_2}\text{Pr}\Bigg[X_1 = \dots, X_{l_1} = X_{m+1}, X_{l_1+1} = \dots, X_{l_1+l_2} = X_{m+2}, X_{m+1} \neq X_{m+2}, \\
&\quad\quad\quad\quad\quad\quad \{X_{l_1+l_2+1}, \dots, X_m\} \cap \{X_{m+1}, X_{m+2}\} = \emptyset, \\ 
&\quad\quad\quad\quad\quad\quad l_2 \mathbbm{1}_{h_{n}(X_{l_1+1})}(X_{m+1}) + \sum_{i=l_1+l_2+1}^m \mathbbm{1}_{h_{n}(X_{i})}(X_{m+1}) = c_{n,1}-l_1, \\
&\quad\quad\quad\quad\quad\quad l_1 \mathbbm{1}_{h_{n}(X_{1})}(X_{m+2})+ \sum_{i=l_1+l_2+1}^m \mathbbm{1}_{h_{n}(X_{i})}(X_{m+2}) = c_{n,2}-l_2\Bigg]\ .
\end{align*}
Setting 
\begin{align*}
E(n,l_1,l_2) &:= \Bigg\{X_1 = \dots, X_{l_1} = X_{m+1}, X_{l_1+1} = \dots, X_{l_1+l_2} = X_{m+2}, X_{m+1} \neq X_{m+2}, \\
&\quad\quad\quad\quad\quad\quad\quad\quad\quad\quad\quad\quad\quad\quad\quad\quad \{X_{l_1+l_2+1}, \dots, X_m\} \cap \{X_{m+1}, X_{m+2}\} = \emptyset\Bigg\},
\end{align*}
we have that $\nu_{\pi_1, \dots, \pi_k}(E(n,l_1,l_2)) = 1$ if and only if two of the $\pi$'s (e.g. $\pi_{k-1}$ and $\pi_k$, being these partitions given up to the order) are exactly equal to the sets $\{1, \dots, l_1, m+1\}$ and $\{l_1+1, \dots, l_1+l_2, m+2\}$, respectively. Therefore, from above, we write the following probability
\begin{align*}
& \text{Pr}\Bigg[E(n,l_1,l_2), l_2 \mathbbm{1}_{h_{n}(X_{l_1+1})}(X_{m+1}) + \sum_{i=l_1+l_2+1}^m \mathbbm{1}_{h_{n}(X_{i})}(X_{m+1}) = c_{n,1}-l_1, \\
&\quad\quad\quad\quad\quad\quad\quad\quad\quad\quad\quad\quad\quad\quad\quad l_1 \mathbbm{1}_{h_{n}(X_{1})}(X_{m+2}) + \sum_{i=l_1+l_2+1}^m \mathbbm{1}_{h_{n}(X_{i})}(X_{m+2}) = c_{n,2}-l_2\Bigg] \\
&\quad= \sum_{k=3}^{n-l_1-l_2+2} \frac{\theta^k}{(\theta)_{(m+2) }} \sum_{(\pi_1, \dots, \pi_{k-2}) \in \Pi(m-l_1-l_2,k-2)} l_1!  l_2! \prod_{i=1}^{k-2} (|\pi_i| -1)! \times \\
&\quad\quad \times \nu_{\pi_1, \dots, \pi_k}\Bigg( l_2 \mathbbm{1}_{h_{n}(X_{l_1+1})}(X_{m+1}) + \sum_{i=l_1+l_2+1}^m \mathbbm{1}_{h_{n}(X_{i})}(X_{m+1}) = c_{n,1}-l_1,\\
&\quad\quad\quad\quad\quad\quad\quad\quad\quad\quad\quad\quad\quad\quad\quad l_1 \mathbbm{1}_{h_{n}(X_{1})}(X_{m+2}) + \sum_{i=l_1+l_2+1}^m \mathbbm{1}_{h_{n}(X_{i})}(X_{m+2}) = c_{n,2}-l_2\Bigg) \\
&\quad= \frac{\theta^2 (\theta)_{(m-l_1-l_2)}}{(\theta)_{(m+2) }} l_1! l_2! 
\sum_{r=1}^{m-l_1-l_2} \frac{\theta^r}{(\theta)_{(m-l_1-l_2) }} \sum_{(\pi_1, \dots, \pi_r) \in \Pi(m-l_1-l_2,r)} \prod_{i=1}^r (|\pi_i| -1)!\\
&\quad\quad \times \Bigg[ \sum_{(j_1,j_2) \in [J]^2} \nu(\{j_1\}) \nu(\{j_2\})\\
&\quad\quad\quad\times \nu_{\pi_1, \dots, \pi_r}\Bigg( 
\sum_{i=l_1+l_2+1}^m \mathbbm{1}_{h_{n}(X_{i})}(j_1) = c_{n,1}-l_1- l_2 \mathds{1}\{j_1 = j_2\}, \\
&\quad\quad\quad\quad\quad\quad\quad\quad\quad\quad\quad\quad\quad\quad\quad\quad\quad\quad \sum_{i=l_1+l_2+1}^m \mathbbm{1}_{h_{n}(X_{i})}(j_2) = c_{n,2}-l_2-l_1 \mathds{1}\{j_1 = j_2\} \Bigg) \Bigg] \ . 
\end{align*}
We observe that the expression within the brackets in the last term, as a sum over $[J]^{2}$,
can be split into the sum of two terms, according on whether $j_1=j_2$ or not. Therefore, we write
\begin{align*}
&\sum_{r=1}^{m-l_1-l_2} \frac{\theta^r}{(\theta)_{(m-l_1-l_2) }} \sum_{(\pi_1, \dots, \pi_r) \in \Pi(m-l_1-l_2,r)} \prod_{i=1}^r (|\pi_i| -1)! \\
&\quad\quad\times \Bigg[ \sum_{j_1=j_2 \in [J]} \nu(\{j_1\}) \nu(\{j_2\}) \nu_{\pi_1, \dots, \pi_r}\Bigg( 
\sum_{i=l_1+l_2+1}^m \mathbbm{1}_{h_{n}(X_{i})}(j_1) = c_{n,1}-l_1- l_2, \\
&\quad\quad\quad\quad\quad\quad\quad\quad\quad\quad\quad\quad\quad\quad\quad\quad\quad\quad\quad\quad\quad\quad \sum_{i=l_1+l_2+1}^m \mathbbm{1}_{h_{n}(X_{i})}(j_2) = c_{n,2}-l_2-l_1 \Bigg) \Bigg] \\
&\quad= \frac{1}{J} \mathbbm{1}\{c_{n,1} = c_{n,2} =: c\}
\binom{m-l_1-l_2}{c-l_1-l_2}\frac{\Gamma(\theta)}{\Gamma(\theta/J) \Gamma(\theta(1 - 1/J))}\\
&\quad\quad \times \frac{\Gamma(\theta/J + c - l_1-l_2) \Gamma(\theta(1 - 1/J) + m - c)}{\Gamma(\theta+m-l_1-l_2)}\ .  
\end{align*}
On the other hand, assuming $J\geq 3$
\begin{align*}
&\sum_{r=1}^{m-l_1-l_2} \frac{\theta^r}{(\theta)_{(m-l_1-l_2) }} \sum_{(\pi_1, \dots, \pi_r) \in \Pi(m-l_1-l_2,r)} \prod_{i=1}^r (|\pi_i| -1)!\\
&\quad\quad \times \Bigg[ \sum_{\substack{(j_1,j_2) \in [J]^2 \\ j_1\neq j_2}} 
\nu(\{j_1\}) \nu(\{j_2\}) \nu_{\pi_1, \dots, \pi_r}\Bigg( \sum_{i=l_1+l_2+1}^m \mathbbm{1}_{h_{n}(X_{i})}(j_1) = c_{n,1}-l_1, \\
& \quad\quad\quad\quad\quad\quad\quad\quad\quad\quad\quad\quad\quad\quad\quad\quad\quad\quad\quad\quad\quad\quad\quad\sum_{i=l_1+l_2+1}^m \mathbbm{1}_{h_{n}(X_{i})}(j_2) = c_{n,2}-l_2 \Bigg) \Bigg] \\
&\quad= \frac{J-1}{J} \binom{m-l_1-l_2}{c_{n,1}-l_1, c_{n,2}-l_2} \frac{\Gamma(\theta)}{[\Gamma(\theta/J)]^2 \Gamma(\theta(1 - 2/J))} \\
&\quad\quad \times\frac{\Gamma(\theta/J + c_{n,1} - l_1) \Gamma(\theta/J + c_{n,2} - l_2)\Gamma(\theta(1 - 2/J) + m - c_{n,1}-c_{n,2})}{\Gamma(\theta+m-l_1-l_2)}.  
\end{align*}
Then,
\begin{align} \label{Num2_Dirichlet_B}
&\text{Pr}\left[ f_{X_{m+1}} = l_1, f_{X_{m+2}} = l_2, \sum_{i=1}^m \mathbbm{1}_{h_{n}(X_{i})}(X_{m+1}) = c_{n,1}, \sum_{i=1}^m \mathbbm{1}_{h_{n}(X_{i})}(X_{m+2}) = c_{n,2}, X_{m+1} \neq X_{m+2}\right] \\
&\quad= \binom{m}{l_1, l_2} \frac{\theta^2 (\theta)_{(m-l_1-l_2) }}{(\theta)_{(m+2) }} l_1! l_2! \nonumber \\
&\quad\quad\times \Bigg[ \frac{1}{J} \mathbbm{1}\{c_{n,1}=c_{n,2}=c\} 
\binom{m-l_1-l_2}{c-l_1-l_2}\frac{\Gamma(\theta)}{\Gamma(\theta/J) \Gamma(\theta(1 - 1/J))} 
\frac{\Gamma(\theta/J + c - l_1-l_2) \Gamma(\theta(1 - 1/J) + m - c)}{\Gamma(\theta+m-l_1-l_2)} \nonumber \\
&\quad+ \frac{J-1}{J} \binom{m-l_1-l_2}{c_{n,1}-l_1, c_{n,2}-l_2} \frac{\Gamma(\theta)}{[\Gamma(\theta/J)]^2 \Gamma(\theta(1 - 2/J))} \nonumber \\
&\quad\quad\times\frac{\Gamma(\theta/J + c_{n,1} - l_1) \Gamma(\theta/J + c_{n,2} - l_2)\Gamma(\theta(1 - 2/J) + m - c_{n,1}-c_{n,2})}{\Gamma(\theta+m-l_1-l_2)} \Bigg].\nonumber
\end{align}
Then, by combining the probability \eqref{Num2_Dirichlet_A} and the probability \eqref{Num2_Dirichlet_B} we write the following expression 
\begin{align}
&\text{Pr}\left[ f_{X_{m+1}} = l_1, f_{X_{m+2}} = l_2, \sum_{i=1}^m \mathbbm{1}_{h_{n}(X_{i})}(X_{m+1}) = c_{n,1}, \sum_{i=1}^m \mathbbm{1}_{h_{n}(X_{i})}(X_{m+2}) = c_{n,2} \right] \nonumber \\
&= \frac{m!}{\Gamma(\theta+m+2)} \Big\{ \mathbbm{1}\{l_1 = l_2 =: l, c_{n,1}=c_{n,2}=c\} \frac{\theta(l+1)}{(c-l)! (m-c)!} \beta_1(\theta, J) \times\nonumber \\
&\qquad \times \Gamma(\theta/J + c - l) \Gamma(\theta(1 - 1/J) + m - c)\nonumber \\
&+ \mathbbm{1}\{c_{n,1}=c_{n,2}=c\}\frac{\theta^2}{J (c-l_1-l_2)! (m-c)!} \beta_1(\theta, J) \Gamma(\theta/J + c - l_1-l_2) \Gamma(\theta(1 - 1/J) + m - c)\nonumber \\
&+ \left(\frac{J-1}{J}\right) \frac{\theta^2}{(c_{n,1}-l_1)! (c_{n,2}-l_2)! (m-c_{n,1}-c_{n,2})!} \beta_2(\theta, J)  \times\nonumber \\
&\qquad \times \Gamma(\theta/J + c_{n,1} - l_1) \Gamma(\theta/J + c_{n,2} - l_2)\Gamma(\theta(1 - 2/J) + m - c_{n,1} - c_{n,2}) \Big\} \label{Num2_Dirichlet}
\end{align}
with
$$
\beta_1(\theta, J) :=\frac{\Gamma(\theta)}{\Gamma(\theta/J) \Gamma(\theta(1 - 1/J))}
$$
and
$$
\beta_2(\theta, J) := \frac{\Gamma(\theta)}{[\Gamma(\theta/J)]^2 \Gamma(\theta(1 - 2/J))}.
$$
The proof is completed by combing the posterior distribution \eqref{object_range} with probabilities \eqref{object_range_denominator} and \eqref{Num2_Dirichlet}.
\end{proof}

Theorem \ref{teo_dir_ran} extends Theorem \ref{teo_direct} to the more general problem of estimating $2$-range queries. In particular, for the collection of hash functions $h_{1},\ldots,h_{N}$, the posterior distribution of $(f_{X_{m+1}},f_{X_{m+2}})$ given $\{(C_{n, h_{n}(X_{m+1})},C_{n, h_{n}(X_{m+2})})\}_{n\in[N]}$ follows from Theorem \ref{teo_dir_ran}  by the assumption that the $h_{n}$'s are i.i.d. according to the strong universal family $\mathcal{H}$, and Bayes theorem. CMS-DP estimates of the $2$-range query $\bar{f}_{2}=f_{x_{m+1}}+f_{x_{m+2}}$ are then obtained as functionals of the posterior distribution of $\bar{f}_{2}$, e.g. posterior mode, posterior mean and posterior median. To conclude, it remains to estimate the prior's parameter $\theta>0$ based on hashed frequencies; this is obtained following the empirical Bayes procedure described in Section \ref{sec2}.

\end{appendices}


\section*{Acknowledgement}

The authors are grateful to the Editor and four anonymous Referees for their comments and corrections that allow to improve remarkably the paper. Stefano Favaro wishes to thank Graham Cormode, Matteo Sesia and Luca Trevisan for stimulating discussions on sketches and generalizations thereof. Emanuele Dolera and Stefano Favaro received funding from the European Research Council (ERC) under the European Union's Horizon 2020 research and innovation programme under grant agreement No 817257. Emanuele Dolera and Stefano Favaro are thankful for the financial support of Italian Ministry of Education, University and Research (MIUR), ``Dipartimenti di Eccellenza" grant 2018-2022. Stefano Favaro is also affiliated to IMATI-CNR ``Enrico Magenes" (Milan, Italy). 



\begin{thebibliography}{9}

\bibitem[Aamand et al.(2019)]{Aam(19)}
\textsc{Aamand, A., Indyk, P. and Vakilian, A.} (2019). Frequency estimation algorithms under Zipfian distribution. \textit{Preprint arXiv:1908.05198}.

\bibitem[Aggarwal and Yu(2010)]{Agg(10)}
\textsc{Aggarwal, C. and Yu, P.} (2010). On classification of high-cardinality data streams. In \textit{Proceedings of the 2010 SIAM International Conference on Data Mining}.

\bibitem[Bacallado et al.(2017)]{Bac(17)}
\textsc{Bacallado, S., Battiston, M., Favaro, S. and Trippa, L.} (2017). Sufficientness postulates for Gibbs-type priors and hierarchical generalizations. \textit{Statistical Science} \textbf{32}, 487--500.

\bibitem[Baldi(2017)]{Bal(17)}
\textsc{Baldi, P.} (2017) \textit{Stochastic calculus.} Springer.

\bibitem[Barab\'asi(2005)]{Bar(05)}
\textsc{Barab\'asi, A.L.} (2005) The origin of bursts and heavy tails in human dynamics. \textit{Nature} \textbf{435}, 227.

\bibitem[Bernton et al.(2019)]{Ber(19)}
\textsc{Bernton, E., Jacob, P.E., Gerber, M. and Robert, C.P.} (2019). On parameter estimation with the Wasserstein distance. \textit{Information and Inference} \textbf{8}, 657--676.

\bibitem[Cai et al.(2018)]{Cai(18)}
\textsc{Cai, D., Mitzenmacher, M. and Adams, R.P.} (2018). A Bayesian nonparametric view on count--min sketch. In \textit{Advances in Neural Information Processing Systems}.

\bibitem[Cancho and Sol\'e(2020)]{Can(03)}
\textsc{Cancho, R.F. and Sol\'e, R.V.} (2003). Least effort and the origins of scaling in human language. \textit{Proceeding of the National Academy of Sciences of USA} \textbf{100}, 788--791.

\bibitem[Charalambides(2005)]{Cha(05)}
\textsc{Charalambides, C.} (2005) \textit{Combinatorial methods in discrete distributions.} Wiley.

\bibitem[Clauset et al.(2009)]{Cla(09)}
\textsc{Clauset, A., Shalizi, C.R. and Newman, M.E.J.} (2009). Power-law distributions in empirical data. \textit{SIAM Review} \textbf{51}, 661--703.

\bibitem[Cormode et al.(2012)]{Cor(12)}
\textsc{Cormode, G., Garofalakis, M. and Haas, P.J.} (2012). \textit{Synopses for massive data: samples, histograms, wavelets, sketches}. Foundations and Trends in Databases.

\bibitem[Cormode and Muthukrishnan(2005)]{Cor(05)}
\textsc{Cormode, G. and Muthukrishnan, S.} (2005). An improved data stream summary: the count-min sketch and its applications. \textit{Journal of Algorithms} \textbf{55}, 58--75.

\bibitem[Cormode and Yi(2020)]{Cor(20)}
\textsc{Cormode, G. and Yi, K.} (2020). \textit{Small summaries for big data}. Cambridge University Press.

\bibitem[De Blasi et al.(2015)]{Deb(15)}  
\textsc{De Blasi, P., Favaro, S., Lijoi, A., Mena, R.H., Pr\"unster, I. and Ruggiero, M.} (2015). Are Gibbs-type priors the most natural generalization of the Dirichlet process? \textit{IEEE Transactions on Pattern Analysis and Machine Intelligence} \textbf{37}, 212--229.

\bibitem[Devroye(2009)]{Dev(09)}
\textsc{Devroye, L.} (2009). Random variate generation for exponentially and polynomially tilted stable distributions. \textit{ACM Transactions on Modeling and Computer Simulation} \textbf{19}, 4.

\bibitem[Dolera(2013)]{Do(13)}
\textsc{Dolera, E.} (2013). Estimates of the approximation of weighted sums of conditionally independent random variables by the normal law. \textit{J. Inequal. Appl.} \textbf{2013}, 320.


\bibitem[Dolera and Favaro(2020a)]{DF(20a)}
\textsc{Dolera, E. and Favaro, S.} (2020). A Berry--Esseen theorem for Pitman's $\alpha$--diversity. \textit{The Annals of Applied Probability} \textbf{30}, 847--869.

\bibitem[Dolera and Favaro(2020b)]{DF(20b)}
\textsc{Dolera, E. and Favaro, S.} (2020). Rates of convergence in de Finetti’s representation theorem, and Hausdorff moment problem. \textit{Bernoulli} \textbf{26}, 1294--1322.


\bibitem[Dolera et al.(2021)]{Dol(21)}
\textsc{Dolera, E., Favaro, S. and Peluchetti, S.} (2021). A Bayesian nonparametric approach to count-min sketch under power-law data stream. In \textit{International Conference on Artificial Intelligence and Statistics}.

\bibitem[Dwork et al.(2010)]{Dwo(10)}
\textsc{Dwork, C. and Naor, M. and Pitassi, T. and Rothblum, G. and Yekhanin, S.} (2010). Pan-private streaming algorithms. In \textit{Proceedings of the Symposium on Innovations in Computer Science}.

\bibitem[Favaro et al.(2009)]{Fav(09)} 
\textsc{Favaro, S., Lijoi, A., R.H., Mena and Pr\"unster, I.} (2009). Bayesian nonparametric inference for species variety with a two parameter Poisson-Dirichlet process prior. \textit{Journal of the Royal Statistical Society Series B} \textbf{71}, 992--1008.

\bibitem[Favaro et al.(2013)]{Fav(13)} 
\textsc{Favaro, S., Lijoi, A. and Pr\"unster, I.} (2013). Conditional formulae for Gibbs-type exchangeable random partitions. \textit{The Annals of Applied Probability} \textbf{23}, 1721--1754.

\bibitem[Favaro et al.(2015)]{Fav(15)}
\textsc{Favaro, S. and Nipoti, B. and Teh, Y.W.} (2015). Random variate generation for Laguerre-type exponentially tilted alpha-stable distributions. \textit{Electronic Journal of Statistics} \textbf{9}, 1230--1242.

\bibitem[Ferguson(1973)]{Fer(73)}
\textsc{Ferguson, T.S.} (1973). A Bayesian analysis of some nonparametric problems. \textit{The Annals of Statistics} \textbf{1}, 209--230.

\bibitem[Ghosal and van der Vaart(2017)]{Gho(17)}
\textsc{Ghosal, S. and van der Vaart, A.} (2017) \textit{Fundamentals of Nonparametric Bayesian Inference.} Cambridge University Press. 

\bibitem[Gnedin et al.(2007)]{Gne(07)}
\textsc{Gnedin, A., Hansen, B. and Pitman, J.} (2007). Notes on the occupancy problems with infinitely many boxes: general asymptotics and power law. \textit{Probability Surveys} \textbf{4}, 146--171.

\bibitem[Goyal et al.(2012)]{Goy(12)}
\textsc{Goyal, A., Daum\'e, H. and Cormode, G.} (2012). Sketch algorithms for estimating point queries in NLP. In \textit{Proceedings of the Joint Conference on Empirical Methods in Natural Language Processing and Computational Natural Language Learning}.

\bibitem[Goya et al.(2009)]{Goy(09)}
\textsc{Goyal, A., Daum\'e, H. and Venkatasubramanian, S.} (2009). Streaming for large scale NLP: language modeling. In \textit{Proceedings of the Conference of the North American Chapter of the Association for Computational Linguistics}.

\bibitem[Harald(2001)]{Har(01)}
\textsc{Harald, B.R.} (2001). \textit{Word Frequency Distributions}. Springer

\bibitem[Harrison(2010)]{Har(10)}
\textsc{Harrison, B.A.} (2010) \textit{Move prediction in the game of Go.} Ph.D Thesis, Harvard University.

\bibitem[Hsu et al.(2019)]{Hsu(19)}
\textsc{Hsu, C., Indyk, P., Katabi, D. and Vakilian, A.} (2019) Learning-based frequency estimation algorithms. In \textit{Proceedings of the International Conference on Learning Representations}.

\bibitem[Huberman and  Adamic(1999)]{Hub(99)}
\textsc{Huberman, B.A. and  Adamic, L.A.} (1999) Internet: growth dynamics of the World-Wide Web. \textit{Nature} \textbf{401}, 131

\bibitem[James(2002)]{Jam(02)}
\textsc{James, L.F.} (2002). Poisson process partition calculus with applications to exchangeable models and Bayesian nonparametrics. \textit{Preprint  arXiv:math/0205093}.

\bibitem[James at al.(2009)]{Jam(09)}
\textsc{James, L.F., Pr\"unster, I., Lijoi, A.} (2009). Posterior analysis for normalized random measures with independent increments. \textit{Scandinavian Journal of Statistics} \textbf{36}, 76--97.

\bibitem[Johnson et al.(2005)]{Joh(05)} 
\textsc{Johnson, N.L., Kemp, A.W. and Kotz, S.} (2005) \textit{Univariate discrete distributions}, Wiley Series in Probability and Statistics.

\bibitem[Kingman(1993)]{Kin(93)}
\textsc{Kingman, J.F.C.} (1993). \textit{Poisson processes.} Wiley Online Library.

\bibitem[Leo Elworth et al.(2020)]{Leo(20)}
\textsc{Leo Elworth, L.A., Wang, Q., Kota, P.K., Barberan, C.J., Coleman, B., Balaji, A., Gupta, G., Baraniuk, R.G., Shrivastava, A. and Treangen, T.J.} (2020).  To petabytes and beyond: recent advances in probabilistic and signal processing algorithms and their application to metagenomics. \textit{Nucleic Acids Research} \textbf{48} 5217--5234.

\bibitem[Letham at al.(2019)]{Let(19)}
\textsc{Letham, B., Karrer, B., Ottoni, G. and Bakshy, E.} (2019). Constrained Bayesian optimization with noisy experiments. \textit{Bayesian Analysis} \textbf{14}, 495--519.

\bibitem[Lijoi et al.(2005)]{Lij(05)}
\textsc{Lijoi, A., Mena, R. H., and Pr\"unster, I.} (2005). Hierarchical mixture modeling with normalized inverse-Gaussian
priors. \textit{Journal of the American Statistical Association} \textbf{100}, 1278--1291.

\bibitem[Monechi et al.(2017)]{Mon(17)}
\textsc{Monechi, B., Ruiz-Serrano, A., Tria, F., and Loreto, V.} (2017). Waves of novelties in the expansion into the adjacent possible. \textit{PloS ONE} \textbf{12}

\bibitem[Muchnik et al.(2013)]{Muc(13)}
\textsc{Muchnik, L., Pei, S., Parra, L.C., Reis, S.D.S, Andrade, J.S., Havlin, S. and Makse, H.A.} (2013). Origins of power-law degree distribution
in the heterogeneity of human activity in social networks. \textit{Nature Scientific Reports} \textbf{3}, 1783

\bibitem[Perman et al.(1992)]{Per(92)}
\textsc{Perman, M., Pitman, J. and Yor, M.} (1992). Size-biased sampling of Poisson point processes and excursions. \textit{Probability Theory and Related Fields} \textbf{92}, 21--39.

\bibitem[Pitel and Fouquier(2015)]{Pit(15)}
\textsc{Pitel, G. and Fouquier, G.} (2015). Count-min-log sketch: approximately counting with approximate counters. In \textit{Proceedings of the International Symposium on Web Algorithm}.

\bibitem[Pitman(1995)]{Pit(95)}
\textsc{Pitman, J.} (1995). Exchangeable and partially exchangeable random partitions. \textit{Probability Theory and Related Fields} \textbf{102}, 145--158.

\bibitem[Pitman(2003)]{Pit(03)}
\textsc{Pitman, J.} (2003). Poisson-Kingman partitions. In \textit{Science and Statistics: A Festschrift for Terry Speed}, Goldstein, D.R. Eds. Institute of Mathematical Statistics.

\bibitem[Pitman(2006)]{Pit(06)}
\textsc{Pitman, J.} (2006). \textit{Combinatorial stochastic processes}. Lecture Notes in Mathematics, Springer Verlag.

\bibitem[Pitman and Yor(1997)]{Pit(97)}
\textsc{Pitman, J. and Yor, M.} (1997). The two parameter Poisson-Dirichlet distribution derived from a stable subordinator. \textit{The Annals of Probability} \textbf{25}, 855--900.

\bibitem[Pr\"unster(2002)]{Pru(02)}   
\textsc{Pr\"unster, I.} (2002). \textit{Random probability measures derived from increasing additive processes and their application to Bayesian statistics}. Ph.d thesis, University of Pavia.

\bibitem[Regazzini(1978)]{Reg(78)}
\textsc{Regazzini, E.} (1978). \textit{Intorno ad alcune questioni relative alla definizione del premio secondo la teoria della credibili\`a.} \textit{Giornale dell'Istituto Italiano degli Attuari} \textbf{41}, 77--89.

\bibitem[Regazzini(2001)]{Reg(01)}
\textsc{Regazzini, E.} (2001). \textit{Foundations of Bayesian statistics and some theory of Bayesian nonparametric methods.} Lecture Notes, Stanford University.

\bibitem[Regazzini et al.(2003)]{Reg(03)} 
\textsc{Regazzini, E., Lijoi, A. and Pr\"unster, I.} (2003). Distributional results for means of normalized random measures with independent increments. \textit{The Annals of Statistics} \textbf{31}, 560--585.

\bibitem[Rybski(2016)]{Ryb(09)}
\textsc{Rybski, D., Buldyrev, S.V., Havlin, S., Liljeros, F. and Makse, H A. } (2016). Scaling laws of human interaction activity. \textit{Proceeding of the National Academy of Sciences of USA} \textbf{106}, 12640.

\bibitem[Sangalli(2006)]{San(06)}
\textsc{Sangalli, M.L.} (2006). Some developments of the normalized random measures with independent increments. \textit{Sankhya A} \textbf{68}, 461--487.

\bibitem[Sethuraman(1994)]{Set(94)}
\textsc{Sethuraman, J.} (1994). A constructive definition of Dirichlet priors. \textit{Statistica Sinica} \textbf{4}, 639--650.

\bibitem[Song et al.(2009)]{Son(09)}
\textsc{Song, H.H., Cho, T.W., Dave, V., Zhang, Y. and Qiu, L.} (2009). Scalable proximity estimation and link prediction in online social networks. In \textit{Proceedings of the ACM SIGCOMM Conference on Internet measurement}.

\bibitem[Ting(2018)]{Tin(18)}
\textsc{Ting, D.} (2018). Count-min: optimal estimation and tight error bounds using empirical error distributions. In \textit{International Conference on Knowledge Discovery and Data Mining}.

\bibitem[Tria et al.(2014)]{Tri(14)}
\textsc{Tria, F., Loreto, V., Servedio, V.D.P and Strogatz, S.H.} (2014). The dynamics of correlated novelties. \textit{Nature Scientific Reports} \textbf{4}, 5890.

\bibitem[Zabell(1997)]{Zab(97)}
\textsc{Zabell, S.L.}  (1997). The continuum of inductive methods revisited. In \textit{The cosmos of science: essays in exploration}, Earman, J. and Norton, J.D. Eds. Universty of Pittsburgh Press.

\bibitem[Zhang et al.(2014)]{Zha(14)}
\textsc{Zhang, Q., Pell, J., Canino-Koning, R., Howe, A.C. and Brown, C.T.} (2014). These are not the k-mers you are looking for: efficient online $k$-mer counting using a probabilistic data structure. \textit{PloS one} \textbf{9}.

\bibitem[Zipf(1949)]{Zip(49)}
\textsc{Zipf, G.K.} (1949). \textit{Human behaviour and the principle of least effort: an introduction to human ecology}. Addison-Wesley.


\bibitem[Zolotarev(1986)]{Zol(86)}
\textsc{Zolotarev, V.M.} (1986). \textit{One dimensional stable distributions.} American Mathematical Society.


\end{thebibliography}
\end{document}